\documentclass[twoside, 11pt]{article}

\usepackage{jmlr2e}
\usepackage[algo2e,ruled]{algorithm2e}
\usepackage{graphicx} % more modern
\usepackage{subfigure}
\usepackage{hyperref,multirow,float}
\usepackage{xspace}
\usepackage{subfigure}
\usepackage{amsmath}
\usepackage{steinmetz}
\usepackage{float}

\jmlrheading{16}{2015}{1-32}{01/15}{0x/15}{Dhruv Mahajan, Nikunj Agrawal, S. Sathiya Keerthi and S. Sundararajan}

\ShortHeadings{An efficient distributed learning algorithm}{Mahajan, Agrawal, Keerthi and Sundararajan}
\firstpageno{1}

\begin{document}

\title{An efficient distributed learning algorithm based on effective local functional approximations}

\author{\name Dhruv Mahajan \email dhrumaha@microsoft.com \\
       \addr Cloud \& Information Services Lab\\
       Microsoft Corporation\\
       Mountain View, CA 94043, USA
       \AND
       \name Nikunj Agrawal \email nikunj157@gmail.com \\
       \addr Indian Institute of Technology \\
       Dept. of Computer Science \& Engineering \\
       Kanpur, India
       \AND
       \name S. Sathiya Keerthi \email keerthi@microsoft.com \\
       \addr Cloud \& Information Services Lab\\
       Microsoft Corporation\\
       Mountain View, CA 94043, USA
       \AND
       \name Sundararajan Sellamanickam \email ssrajan@microsoft.com \\
       \addr Microsoft Research \\
       Bangalore, India
       \AND 
       \name L\'{e}on Bottou \email leonbo@microsoft.com \\
      \addr Microsoft Research \\
      New York, USA
       }

\editor{xxx}

\maketitle
\begin{abstract}%   <- trailing '%' for backward compatibility of .sty file
Scalable machine learning over big data is an important problem that is receiving a lot of attention in recent years. On popular distributed environments such as Hadoop running on a cluster of commodity machines, communication costs are substantial and algorithms need to be designed suitably considering those costs.
In this paper we give a novel approach to the distributed training of linear classifiers (involving smooth losses and $L_2$ regularization) that is designed to reduce the total communication costs. At each iteration, the nodes minimize locally formed approximate objective functions; then the resulting minimizers are combined to form a descent direction to move. Our approach gives a lot of freedom in the formation of the approximate objective function as well as in the choice of methods to solve them. The method is shown to have $O(\log(1/\epsilon))$ time convergence. The method can be viewed as an iterative parameter mixing method. A special instantiation yields a parallel stochastic gradient descent method with strong convergence. When communication times between nodes are large, our method is much faster than the Terascale method~\citep{agarwal2011}, which is a state of the art distributed solver based on the statistical query model~\citep{chu2006} that computes function and gradient values in a distributed fashion. We also evaluate against other recent distributed methods and demonstrate superior performance of our method.
\end{abstract}


\def\grad{\nabla}
\def\wtilde{\tilde{w}}
\def\Cone{{\cal{C}}^1}
\def\kappap{\kappa^\prime}
\def\Lhat{\hat{L}}
\def\fhat{\hat{f}}
\def\what{\hat{w}}
\def\dhat{\hat{d}}
\def\mysgn{\operatorname{sgn}}
\def\ftilde{\tilde{f}}
\def\khat{\hat{k}}
\def\defs{\stackrel{\text{def}}{=}}
\def\vhat{\hat{v}}

\def\ttilde{\tilde{t}}
\def\that{\hat{t}}
\def\tstar{t^\star}
\newcommand{\yrcite}[1]{\citeyearpar{#1}}
\long\def\comment#1{}

\def\K{{\bf ToDo:~}}

\section{Introduction}
\label{intro}

In recent years, machine learning over big data has become an important problem, not only in web related applications, but also more commonly in other applications, e.g., in the data mining over huge amounts of user logs. The data in such applications are usually collected and stored in a decentralized fashion over a cluster of commodity machines (nodes) where communication times between nodes is significantly large. In such a setting it is natural for the examples to be partitioned over the nodes. The development of efficient distributed machine learning algorithms that minimize communication between nodes is an important problem.

In this paper we consider the distributed batch training of linear classifiers in which: (a) both, the number of examples and the number of features are large; (b) the data matrix is sparse; (c) the examples are partitioned over the nodes; (d) the loss function is convex and differentiable; and, (e) the $L_2$ regularizer is employed. This problem involves the large scale unconstrained minimization of a convex, differentiable objective function $f(w)$ where $w$ is the weight vector.
The minimization is usually performed using an iterative descent method in which an iteration starts from a point $w^r$, computes a direction $d^r$ that satisfies
\begin{equation}
\mbox{{\small\bf sufficient angle of descent:}}\; \phase{-g^r, d^r} \le \theta \label{angle}
\end{equation}
where $g^r=g(w^r)$, $g(w)=\grad f(w)$, $\phase{a,b}$ is the angle between vectors $a$ and $b$, and $0 \le \theta < \pi/2$, and then performs a line search along the direction $d^r$ to find the next point, $w^{r+1}=w^r+t d^r$. Let $w^\star=\arg\min_w f(w)$. A key side contribution of this paper is the proof that, when $f$ is convex and satisfies some additional weak assumptions, the method has global linear rate of convergence ({\it glrc})\footnote{We say a method has {\it glrc} if $\exists$ $0<\delta<1$ such that $(f(w^{r+1})-f(w^\star)) \le \delta (f(w^r)-f(w^\star))\;\forall r$.} and so it finds a point $w^r$ satisfying $f(w^r)-f(w^\star)\le\epsilon$ in $O(log(1/\epsilon))$ iterations. {\it The main theme of this paper is that the flexibility offered by this method with strong convergence properties allows us to build a class of useful distributed learning methods with good computation and communication trade-off capabilities.}

%Let us consider large scale learning in a distributed setting in which examples are partitioned over a number of computing nodes.
Take one of the most effective distributed methods, viz., SQM (Statistical Query Model) \citep{chu2006,agarwal2011}, which is a batch, gradient-based descent method. The gradient is computed in a distributed way with each node computing the gradient component corresponding to its set of examples. This is followed by an aggregation of the components. We are interested in systems in which the communication time between nodes is large relative to the computation time in each node.\footnote{This is the case when feature dimension is huge. Many applications gain performance when the feature space is expanded, say, via feature combinations, explicit expansion of nonlinear kernels etc.} For iterative algorithms such as SQM, the total training time is given by
\begin{equation}
\mbox{Training time} = (T^{cmp} + T^{com})\; T^{iter}
\end{equation}
where $T^{cmp}$ and $T^{com}$ are respectively, the computation time and the communication time per iteration and $T^{iter}$ is the total number of iterations. When $T^{com}$ is large, it is not optimal to work with an algorithm such as SQM that has $T^{cmp}$ small and due to which, $T^{iter}$ is large.
In such a scenario, it is useful to ask: {\bf Q1.} {\it In each iteration, can we do more computation in each node so that the number of iterations and hence the number of communication passes are decreased, thus reducing the total computing time?}
%We will show that this question can be answered positively by using the iterative descent method suitably.

There have been some efforts in the literature to reduce the amount of communication. In one class of such methods, the current $w^r$ is first passed on to all the nodes. Then, each node $p$ forms an approximation $\ftilde_p$ of $f$ using only its examples, followed by several optimization iterations (local passes over its examples) to decrease $\ftilde_p$ and reach a point $w_p$. The $w_p\;\forall p$ are averaged to form the next iterate $w^{r+1}$. One can stop after just one major iteration (going from $r=0$ to $r=1$); such a method is referred to as {\it parameter mixing (PM)}~\citep{mann2009}. Alternatively, one can do many major iterations; such a method is referred to as {\it iterative parameter mixing (IPM)}~\citep{hall2010}. Convergence theory for such methods is inadequate~\citep{mann2009, mcdonald2010}, which prompts us to ask: {\bf Q2.} {\it Is it possible to devise an IPM method that produces $\{w^r\} \rightarrow w^\star$?}

In another class of methods, the dual problem is solved in a distributed fashion~\citep{pechyony2011, yang2013a, yang2013b, jaggi2014}. Let $\alpha_p$ denote the dual vector associated with the examples in node $p$. The basic idea is to optimize $\{\alpha_p\}$ in parallel and then use a combination of the individual directions thus generated to take an overall step. In practice these methods tend to have slow convergence; see~Section~\ref{expts} for details.

%For large scale learning on a single machine, it is now well established that example-wise methods\footnote{These methods update $w$ after scanning each example.} such as stochastic gradient descent (SGD) and its variations~\citep{bottou2010, johnson2013} and dual coordinate ascent~\citep{hsieh2008} are much faster than batch gradient-based methods. However, example-wise methods are inherently sequential. If one employs a method such as SGD as the local optimizer for $\ftilde_p$ in PM/IPM, the result is, in essence, a parallel SGD method. However, convergence theory for such a method is limited, even that requiring a complicated analysis~\citep{zinkevich2010}. Thus, we ask: {\bf Q3.} {\it Can we form a parallel SGD method with strong convergence properties?}

We make a novel and simple use of the iterative descent method mentioned at the beginning of this section to design a distributed algorithm that answers Q1-Q2 positively. The main idea is to use distributed computation for generating a good search direction $d^r$ and not just for forming the gradient as in SQM. At iteration $r$, let us say each node $p$ has the current iterate $w^r$ and the gradient $g^r$. This information can be used together with the examples in the node to form a function $\fhat_p(\cdot)$ that approximates $f(\cdot)$ and satisfies $\grad\fhat_p(w^r)=g^r$. One simple and effective suggestion is:
\begin{equation}
\fhat_p(w)=f_p(w)+(g^r - \nabla f_p(w^r))\cdot(w-w^r)
\label{sugg1}
\end{equation}
where $f_p$ is the part of $f$ that does not depend on examples outside node $p$; the second term in (\ref{sugg1}) can be viewed as an approximation of the objective function part associated with data from the other nodes. In Section~\ref{distr} we give other suggestions for forming $\fhat_p$. Now $\fhat_p$ can be optimized within node $p$ using any method ${\cal M}$ which has {\it glrc}, e.g., Trust region method, L-BFGS, etc. There is no need to optimize $\fhat_p$ fully. We show (see Section~\ref{distr}) that, in a constant number of local passes over examples in node $p$, an approximate minimizer $w_p$ of $\fhat_p$ can be found such that the direction $d_p=w_p-w^r$ satisfies the sufficient angle of descent condition, (\ref{angle}). A convex combination of the set of directions generated in the nodes, $\{d_p\}$ forms the overall direction $d^r$ for iteration $r$. Note that $d^r$ also satisfies (\ref{angle}). The result is an overall distributed method that finds a point $w$ satisfying $f(w)-f(w^\star)\le\epsilon$ in $O(\log (1/\epsilon))$ time. This answers {\bf Q2}.

The method also reduces the number of communication passes over the examples compared with SQM, thus also answering {\bf Q1}. The intuition here is that, if each $\fhat_p$ is a good approximation of $f$, then $d^r$ will be a good global direction for minimizing $f$ at $w^r$, and so the method will move towards $w^\star$ much faster than SQM.

In summary, the paper makes the following contributions. 
First, for convex $f$ we establish {\it glrc} for a general iterative descent method. 
Second, and more important, we propose a distributed learning algorithm that: (a) converges in $O(\log (1/\epsilon))$ time, thus leading to an IPM method with strong convergence; (b) is more efficient than SQM when communication costs are high; and (c) flexible in terms of the local optimization method ${\cal M}$ that can be used in the nodes. 

There is also another interesting side contribution associated with our method. It is known that example-wise methods such as stochastic gradient descent (SGD) are inherently sequential and hard to parallelize~\citep{zinkevich2010}. By employing SGD as ${\cal M}$, the local optimizer for $\ftilde_p$ in our method, we obtain a parallel SGD method with good performance as well as strong convergence properties. This contribution is covered in detail in~\citet{mahajan2013}; we give a summarized view in Subsection~\ref{subsec:psgd}.

Experiments (Section~\ref{expts}) validate our theory as well as show the benefits of our method for large dimensional datasets where communication is the bottleneck. We give a discussion on unexplored possibilities for extending our distributed learning method in Section~\ref{disc} and conclude the paper in Section~\ref{conc}.

%We close this section with a brief discussion of works related to this paper.

%\noindent {\bf Related work.} Let us begin with papers related to the general descent method and its convergence.

\section{Basic descent method}
\label{general}

Let $f\in\Cone$, the class of continuously differentiable functions\footnote{It would be interesting future work to extend all the theory developed in this paper to non-differentiable convex functions, using sub-gradients.}, $f$ be convex, and the gradient $g$ satisfy the following assumptions.

\noindent {\bf A1.} $g$ is Lipschitz continuous, i.e., $\exists$ $L>0$ such that
$\|g(w)-g(\wtilde)\| \le L \|w-\wtilde\| \;\;\; \forall \; w, \wtilde$.

\noindent {\bf A2.} $\exists$ $\sigma >0$ such that
$(g(w)-g(\wtilde))\cdot(w-\wtilde) \ge \sigma \|w-\wtilde\|^2  \;\;\; \forall \; w, \wtilde$.

A1 and A2 are essentially second order conditions: if $f$ happens to be twice continuously differentiable, then $L$ and $\sigma$ can be viewed as upper and lower bounds on the eigenvalues of the Hessian of $f$. A convex function $f$ is said to be $\sigma$- strongly convex if $f(w)-\frac{\sigma}{2} \|w\|^2$ is convex. In machine learning, all convex risk functionals in $\Cone$ having the $L_2$ regularization term, $\frac{\lambda}{2} \|w\|^2$ are $\sigma$- strongly convex with $\sigma=\lambda$. It can be shown~\citep{smola2008}
%(see (3.18) there)
that, if $f$ is $\sigma$-strongly convex, then $f$ satisfies assumption A2.

Let $f^r=f(w^r)$, $g^r=g(w^r)$ and $w^{r+1}=w^r+t d^r$. Consider the following standard line search conditions.
\begin{eqnarray}
\mbox{{\small\bf Armijo:}}\; & f^{r+1} \le f^r + \alpha g^r\cdot(w^{r+1}-w^r) \label{ag} \\
\mbox{{\small\bf Wolfe:}}\;  & g^{r+1}\cdot d^r \ge \beta g^r\cdot d^r \label{wolfe}
\end{eqnarray}
where $0<\alpha<\beta<1$.

\begin{algorithm2e}
\caption{Descent method for $f$\label{GD}}
Choose $w^0$\;
\For{$r=0,1 \ldots$}{
1. Exit if $g^r=0$\;
2. Choose a direction $d^r$ satisfying (\ref{angle})\;
3. Do line search to choose $t>0$ so that $w^{r+1}=w^r+td^r$ satisfies the Armijo-Wolfe conditions (\ref{ag}) and (\ref{wolfe})\;
}
\end{algorithm2e}
Let us now consider the general descent method in Algorithm~\ref{GD} for minimizing $f$. The following result shows that the algorithm is well-posed. A proof is given in the appendix B.

%Standard results in optimization~\citep{dennis1996} (see Theorem 6.3.2) can be easily modified to show that there always exists an interval of $t$ values over which (\ref{wolfe}) and (\ref{ag}) hold.

\noindent {\bf Lemma 1.} Suppose $g^r\cdot d^r<0$.
Then $\{ t: $ (\ref{ag}) and (\ref{wolfe}) hold for $w^{r+1}=w^r+td^r\} = [t_\beta,t_\alpha]$, where $0<t_\beta<t_\alpha$, and $t_\beta$, $t_\alpha$ are the unique roots of
\begin{eqnarray}
g(w^r+t_\beta d^r)\cdot d^r = \beta g^r\cdot d^r, \label{tbeta} \\
f(w^r+t_\alpha d^r) = f^r + t_\alpha \alpha g^r\cdot d^r, \;\; t_\alpha>0. \label{talpha}
\end{eqnarray}

%If $\alpha$ is chosen very small, say $\alpha=10^{-4}$ and $\beta$ close to 1, say $0.99$, then this interval becomes large and it is easy to locate a $t$ satisfying the two conditions. (We need to say a bit more here, e.g., give good starting $t$ values and a clean line search algorithm that works and is also efficient. See section~\ref{lsearch} for some ideas for bracketing the minimum in the line search.)

\noindent {\bf Theorem 2.} Let $w^\star=\arg\min_w f(w)$ and $f^\star=f(w^\star)$.\footnote{Assumption A2 implies that $w^\star$ is unique.} Then $\{w^r\}\rightarrow w^\star$. Also, we have {\it glrc}, i.e., $\exists$ $\delta$ satisfying $0<\delta<1$ such that
$(f^{r+1} - f^\star) \le \delta\, (f^r-f^\star) \; \forall \; r\ge 0$,
and, $f^r-f^\star\le\epsilon$ is reached after at most
$\frac{\log ((f^0-f^\star)/\epsilon)}{\log(1/\delta)}$
iterations. An upper bound on $\delta$ is $(1 - 2\alpha(1-\beta)\frac{\sigma^2}{L^2} \cos^2\theta)$.

A proof of Theorem 2 is given in the appendix B. If one is interested only in proving convergence, it is easy to establish under the assumptions made; such theory goes back to the classical works of Wolfe~\citep{wolfe1969, wolfe1971}. But proving {\it glrc} is harder. There exist proofs for special cases such as the gradient descent method~\citep{boyd2004}. The {\it glrc} result in~\citet{wang2013} is only applicable to descent methods that are ``close" (see equations ($7$) and ($8$) in~\citet{wang2013}) to the gradient descent method. Though Theorem 2 is not entirely surprising, as far as we know, such a result does not exist in the literature.

It is important to note that the rate of convergence indicated by the upper bound on $\delta$ given in Theorem 2 is pessimistic since it is given for a very general descent algorithm that includes plain batch gradient descent which is known to have a slow rate of convergence. Depending on the method used for choosing $d^r$ the actual rate of convergence can be a lot better. For example, we observe very good rates for our distributed method; see Section~\ref{expts}.

\section{Distributed training}
\label{distr}

In this section we discuss full details of our distributed training algorithm.
Let $\{x_i,y_i\}$ be the training set associated with a binary classification problem ($y_i\in\{1,-1\}$). Consider a linear classification model, $y=\mysgn(w^Tx)$. Let $l(w\cdot x_i,y_i)$ be a continuously differentiable loss function that has Lipschitz continuous gradient. This allows us to consider loss functions such as least squares, logistic loss and squared hinge loss. Hinge loss is not covered by our theory since it is non-differentiable.

Suppose the training examples are distributed in $P$ nodes. Let: $I_p$ be the set of indices $i$ such that $(x_i,y_i)$ sits in the $p$-th node; $L_p(w) = \sum_{i\in I_p} l(w;x_i,y_i)$ be the total loss associated with node $p$; and, $L(w)=\sum_p L_p(w)$ be the total loss over all nodes. Our aim is to minimize the regularized risk functional $f(w)$ given by
\begin{equation}
f(w) = \frac{\lambda}{2} \|w\|^2 + L(w) = \frac{\lambda}{2} \|w\|^2 + \sum_p L_p(w),
\label{risk}
\end{equation}
where $\lambda>0$ is the regularization constant. It is easy to check that $g=\grad f$ is Lipschitz continuous.

\subsection{Our approach}
\label{subsec:approach}

Our distributed method is based on the descent method in Algorithm~\ref{GD}. We use a master-slave architecture.\footnote{An {\it AllReduce} arrangement of nodes~\citep{agarwal2011} may also be used.} Let the examples be partitioned over $P$ slave nodes. Distributed computing is used to compute the gradient $g^r$ as well as the direction $d^r$.  In the $r$-th iteration, let us say that the master has the current $w^r$ and gradient $g^r$. One can communicate these to all $P$ (slave) nodes. The direction $d^r$ is formed as follows. Each node $p$ constructs an approximation of $f(w)$ using only information that is available in that node, call it $\fhat_p(w)$, and (approximately) optimizes it (starting from $w^r$) to get the point $w_p$. Let $d_p=w_p-w^r$. Then $d^r$ is chosen to be any convex combination of $d_p\;\forall p$. Doing line search along the $d^r$ direction completes the $r$-th iteration. Line search involves distributed computation, but it is inexpensive; we give details in Subsection~\ref{subsec:practical}.

We want to point out that $\fhat_p$ can change with $r$, i.e., one is allowed to use a different $\fhat_p$ in each outer iteration. We just don't mention it as $\fhat_p^r$ to avoid clumsiness of notation. In fact, all the choices for $\fhat_p$ that we discuss below in Subsection~\ref{subsec:fhatp} are such that $\fhat_p$ depends on the current iterate, $w^r$.

\subsection{Choosing $\fhat_p$}
\label{subsec:fhatp}

Our method offers great flexibility in choosing $\fhat_p$ and the method used to optimize it. We only require $\fhat_p$ to satisfy the following.

\noindent {\bf A3.} $\fhat_p$ is $\sigma$-strongly convex, has Lipschitz continuous gradient and satisfies {\it gradient consistency at $w^r$:} $\grad\fhat_p(w^r)=g^r$.

Below we give several ways of forming $\fhat_p$. The $\sigma$-strongly convex condition is easily taken care of by making sure that the $L_2$ regularizer is always a part of $\fhat_p$. This condition implies that
\begin{equation}
\fhat_p(w_p) \ge \fhat_p(w^r) + \grad\fhat_p(w^r)\cdot (w_p-w^r) + \frac{\sigma}{2} \|w_p-w^r\|^2.
\label{fhatsig}
\end{equation}
The gradient consistency condition is motivated by the need to satisfy the angle condition (\ref{angle}). Since $w_p$ is obtained by starting from $w^r$ and optimizing $\fhat_p$, it is reasonable to assume that $\fhat_p(w_p)<\fhat_p(w^r)$. Using these in (\ref{fhatsig}) gives $-g^r\cdot d_p > 0$. Since $d^r$ is a convex combination of the $d_p$ it follows that $-g^r\cdot d^r > 0$. Later we will formalize this to yield (\ref{angle}) precisely.

\def\Ltilde{\tilde{L}}

A general way of choosing the approximating functional $\fhat_p$ is
\begin{equation}
\fhat_p(w) = \frac{\lambda}{2} \|w\|^2 + \Ltilde_p(w) +\Lhat_p(w),
\label{riskapp}
\end{equation}
where $\Ltilde_p$ is an approximation of $L_p$ and $\Lhat_p(w)$ is an approximation of $L(w)-L_p(w)=\sum_{q\not= p}$ $L_q(w)$. A natural choice for $\Ltilde_p$ is $L_p$ itself since it uses only the examples within node $p$; but there are other possibilities too. To maintain communication efficiency, we would like to design an $\Lhat_p$ such that it does not explicitly require any examples outside node $p$. To satisfy A3 we need $\Lhat_p$ to have Lipschitz continuous gradient. Also, to aid in satisfying gradient consistency, appropriate linear terms are added. We now suggest five choices for $\fhat_p$.

\noindent {\bf Linear Approximation.} Set $\Ltilde_p=L_p$ and choose $\Lhat_p$ based on the first order Taylor series. Thus,
\begin{equation}
\Ltilde_p(w) = L_p(w), \;\;\;\;\;\; \Lhat_p(w) = (\grad L(w^r) - \grad L_p(w^r))\cdot(w-w^r).
\label{linapp}
\end{equation}
(The zeroth order term needed to get $f(w^r)=\fhat(w^r)$ is omitted everywhere because it is a constant that plays no role in the optimization.) Note that $\grad L(w^r) = g^r - \lambda w^r$ and so it is locally computable in node $p$; this comment also holds for the methods below.

\noindent {\bf Hybrid approximation.}
This is an improvement over the linear approximation where we add a quadratic term to $\Lhat_p$. This is done by using $(P-1)$ copies of the quadratic term of $L_p$ to approximate the quadratic term of $\sum_{q\not=p} L_q$.
\begin{eqnarray}
\Ltilde_p(w) = L_p(w), \\
\Lhat_p(w) = (\grad L(w^r) - \grad L_p(w^r))\cdot(w-w^r) + \frac{P-1}{2} (w-w^r)^T H_p^r (w-w^r),
\label{hybapp}
\end{eqnarray}
where $H_p^r$ is the Hessian of $L_p$ at $w^r$.
This corresponds to using subsampling to approximate the Hessian of $L(w)-L_p(w)$ at $w^r$ utilizing only the local examples. Subsampling based Hessian approximation is known to be very effective in optimization for machine learning~\citep{byrd2012}.

\noindent {\bf Quadratic approximation.}
This is a pure quadratic variant where a second order approximation is used for $\Ltilde_p$ too.
\begin{eqnarray}
\Ltilde_p(w) = \grad L_p(w^r) \cdot (w-w^r) + \frac{1}{2} (w-w^r)^T H_p^r (w-w^r), \\
\Lhat_p(w) = (\grad L(w^r) - \grad L_p(w^r))\cdot(w-w^r) + \frac{P-1}{2} (w-w^r)^T H_p^r (w-w^r).
\label{quadapp}
\end{eqnarray}
The comment made earlier on the goodness of subsampling based Hessian for the Hybrid approximation applies here too.

\noindent {\bf Nonlinear approximation.} 
Here the idea is to use $P-1$ copies of $L_p$ to approximate $\sum_{q\not=p} L_q$.
\begin{eqnarray}
\label{sszapp1}
\Ltilde_p(w) = L_p(w), \\
\Lhat_p(w) = (\grad L(w^r) - P\grad L_p(w^r))\cdot(w-w^r) + (P-1)L_p(w).
\label{sszapp2}
\end{eqnarray}
A somewhat similar approximation is used in~\citet{sharir2014}. But the main algorithm where it is used does not have deterministic monotone descent like our algorithm. The gradient consistency condition, which is essential for establishing function descent, is not respected in that algorithm. In Section~\ref{expts} we compare our methods against the method in~\citet{sharir2014}.

\noindent {\bf BFGS approximation.}
For $\Ltilde_p$ we can either use $L_p$ or a second order approximation, like in the approximations given above. For $\Lhat_p$ we can use a second order term, $\frac{1}{2}(w-w^r)\cdot H(w-w^r)$ where $H$ is a positive semi-definite matrix; for $H$ we can use a diagonal approximation or keep a limited history of gradients and form a BFGS approximation of $L-L_p$.

The distributed method described above is an instance of Algorithm~\ref{GD} and so Theorem 2 can be used.
In Theorem 2 we mentioned a convergence rate, $\delta$. For $\cos\theta = \sigma/L$ this yields the rate $\delta = (1-2\alpha(1-\beta)(\frac{\sigma}{L})^4)$. This rate is obviously pessimistic given that it applies to general choices of $\fhat_p$ satisfying minimal assumptions. Actual rates of convergence depend a lot on the choice made for $\fhat_p$. Suppose we choose $\fhat_p$ via Hybrid, Quadratic or Nonlinear approximation choices mentioned in Subsection~\ref{subsec:fhatp} and minimize $\fhat_p$ exactly in the inner optimization. These approximations are invariant to coordinate transformations such as $w'=Bw$, where $B$ is a positive definite matrix. Note that Armijo-Wolfe line search conditions are also unaffected by such transformations. What this means is that, at each iteration, we can, without changing the algorithm, choose for analysis a coordinate transformation that gives the best rate. The Linear approximation choice for $\fhat_p$ does not enjoy this property. This explains why the Hybrid, Quadratic and Nonlinear approximations perform so well and give great rates of convergence in practice; see the experiments in Section~\ref{subsec:our} and Subsection~\ref{subsubsec:rate}.

In Section~\ref{expts} we evaluate some of these approximations in detail.

\subsection{Convergence theory}
\label{subsec:convth}

In practice, exactly minimizing $\fhat_p$ is infeasible. For convergence, it is not necessary for $w_p$ to be the minimizer of $\fhat_p$; we only need to find $w_p$ such that the direction $d_p=w_p-w^r$ satisfies (\ref{angle}). The angle $\theta$ needs to be chosen right. Let us discuss this first. Let $\what_p^\star$ be the minimizer of $\fhat_p$. It can be shown (see appendix B) that $\phase{\what_p^\star-w^r,-g^r} \le \cos^{-1}\frac{\sigma}{L}$. To allow for $w_p$ being an approximation of $\what_p^\star$, we choose $\theta$ such that
\begin{equation}
\frac{\pi}{2} > \theta > \cos^{-1} \frac{\sigma}{L}.
\label{thetadef}
\end{equation}
The following result shows that if an optimizer with {\it glrc} is used to minimize $\fhat_p$, then, only a constant number of iterations is needed to satisfy the sufficient angle of descent condition.

\noindent {\bf Lemma 3.} Assume $g^r\not=0$. Suppose we minimize $\fhat_p$ using an optimizer ${\cal M}$ that starts from $v^0=w^r$ and generates a sequence $\{v^k\}$ having {\it glrc}, i.e.,
$\fhat_p(v^{k+1}) - \fhat_p^\star \le \delta (\fhat_p(v^k) - \fhat_p^\star)$,
where $\fhat_p^\star = \fhat_p(\what_p^\star)$. Then, there exists $\khat$ (which depends only on $\sigma$ and $L$) such that
$\phase{-g^r,v^k-w^r} \le \theta \;\; \forall k\ge \khat$.

Lemma 3 can be combined with Theorem 2 to yield the following convergence theorem.

\noindent {\bf Theorem 4.} Suppose $\theta$ satisfies (\ref{thetadef}), ${\cal M}$ is as in Lemma 3 and, in each iteration $r$ and for each $p$, $\khat$ or more iterations of ${\cal M}$ are applied to minimize $\fhat_p$ (starting from $w^r$) and get $w_p$. Then the distributed method converges to a point $w$ satisfying $f(w)-f(w^\star)\le\epsilon$ in $O(\log(1/\epsilon))$ time.

Proofs of Lemma 3 and Theorem 4 are given in appendix B.

%{\bf Leon's comments to go here.} Proofs of the above results are given in appendix A. The bound on rate of convergence derived there should not be interpreted as the actual rates associated with our method. etc.

%\vspace*{0.1in}
%{\bf Related work.}
%\vspace*{0.05in}
%As already mentioned in section~\ref{intro} our distributed method can be viewed as an IPM method, but one which has strong convergence properties.

%The ADMM method~\citep{Boyd2011}, like our method, solves approximate problems in the nodes and iteratively reaches the full batch solution. But it has the following disadvantages: (a) it does not have {\it glrc}; (b) the approximation problems are dictated by the ADMM formulation and so there is little flexibility; (c) the approximation problems need to be solved precisely; and (d) its efficiency is sensitive to the choice of the penalty parameter which is problem dependent.

\subsection{Practical implementation.} 
\label{subsec:practical}

We refer to our method by the acronym, FADL - Function Approximation based Distributed Learning.
Going with the practice in numerical optimization, we replace (\ref{angle}) by the condition, $-g^r\cdot d^r > 0$ and use $\alpha=10^{-4}$, $\beta=0.9$ in (\ref{ag}) and (\ref{wolfe}). In actual usage,  Algorithm~\ref{GD} can be terminated when $\|g^r\|\le \epsilon_g\|g^0\|$ is satisfied at some $r$. Let us take line search next. On $w=w^r+t d^r$, the loss has the form $l(z_i+te_i,y_i)$ where $z_i=w^r\cdot x_i$ and $e_i=d^r\cdot x_i$. Once we have computed $z_i\;\forall i$ and $e_i\;\forall i$, the distributed computation of $f(w^r+t d^r)$ and its derivative with respect to $t$ is cheap as it does not involve any computation involving the data, $\{x_i\}$. Thus, many $t$ values can be explored cheaply. Since $d^r$ is determined by approximate optimization, $t=1$ is expected to give a decent starting point. We first identify an interval $[t_1,t_2]\subset [t_\beta,t_\alpha]$ (see Lemma 1) by starting from $t=1$ and doing forward and backward stepping. Then we check if $t_1$ or $t_2$ is the minimizer of $f(w^r+t d^r)$ on $[t_1,t_2]$; if not, we do several bracketing steps in $(t_1,t_2)$ to locate the minimizer approximately. Finally, when using method ${\cal M}$, we terminate it after a fixed number of steps, $\khat$. Algorithm~\ref{alg:fadl} gives all the steps of FADL while also mentioning the distributed communications and computations involved.

\begin{algorithm2e}
\label{alg:fadl}
\caption{FADL - Function Approximation based Distributed Learning.
{\it com:} communication; {\it cmp:} = computation; {\it agg:} aggregation. ${\cal M}$ is the optimizer used for minimizing $\fhat_p$.}
Choose $w^0$\;
\For{$r=0,1 \ldots$}{
1. Compute $g^r$ ({\it com}: $w^r$; {\it cmp:} Two passes over data; {\it agg:} $g^r$); By-product: $\{z_i=w^r\cdot x_i\}$\;
2. Exit if $\|g^r\| \le \epsilon_g \|g^0\|$\;
3. \For{$p=1,\ldots,P$ (in parallel)}{
4.   Set $v^0=w^r$\;
5.   \For{$k=0,1,\ldots,\khat$}{
        6. Find $v^{k+1}$ using one iteration of ${\cal M}$\;
     }
     7. Set $w_p=v^{\khat+1}$\;
   }
8. Set $d^r$ as any convex combination of $\{w_p\}$ ({\it agg:} $w_p$)\;
9. Compute $\{e_i=d^r\cdot x_i\}$ ({\it com:} $d^r$; {\it cmp:} One pass over data)\;
10. Do line search to find $t$ (for each $t$: {\it com:} $t$; {\it cmp:} $l$ and $\partial l/\partial t$ {\it agg:} $f(w^r+t d^r)$ and its derivative wrt $t$)\;
11. Set $w^{r+1} = w^r+t d^r$\;
}
\end{algorithm2e}

\noindent {\bf Choices for ${\cal M}$.}
There are many good methods having (deterministic) {\it glrc}: L-BFGS, TRON~\citep{lin2008}, Primal coordinate descent~\citep{chang2008}, etc. One could also use methods with {\it glrc} in the expectation sense (in which case, the convergence in Theorem 4 should be interpreted in some probabilistic sense). This nicely connects our method with recent literature on parallel SGD. We discuss this in the next subsection only briefly as it is outside the scope of the current paper. See our related work~\citep{mahajan2013} for details.

\subsection{Connections with parallel SGD}
\label{subsec:psgd}

For large scale learning on a single machine, example-wise methods\footnote{These methods update $w$ after scanning each example.} such as stochastic gradient descent (SGD) and its variations~\citep{bottou2010, johnson2013} and dual coordinate ascent~\citep{hsieh2008} perform quite well. However, example-wise methods are inherently sequential. If one employs a method such as SGD as ${\cal M}$, the local optimizer for $\ftilde_p$, the result is, in essence, a parallel SGD method. However, with parameter mixing and iterative parameter mixing methods~\citep{mann2009, hall2010, mcdonald2010} (we briefly discussed these methods in Section~\ref{intro}) that do not do line search, convergence theory is limited, even that requiring a complicated analysis~\citep{zinkevich2010}; see also~\citet{mann2009} for some limited results. Thus, the following has been an unanswered question: {\bf Q3.} {\it Can we form a parallel SGD method with strong convergence properties such as glrc?}

As one special instantiation of our distributed method, we can use, for the local optimization method ${\cal M}$, any variation of SGD with {\it glrc} (in expectation), e.g., the one in~\citet{johnson2013}. For this case, in a related work of ours~\citep{mahajan2013} we show that our method has $O(\log (1/\epsilon))$ time convergence in a probabilistic sense. The result is a strongly convergent parallel SGD method, which answers {\bf Q3}. An interesting side observation is that, the single machine version of this instantiation is very close to the variance-reducing SGD method in~\citet{johnson2013}. We discuss this next.

\noindent {\bf Connection with SVRG. }
Let us take the $\fhat_p$ in~(\ref{linapp}). Let $n_p=|I_p|$ be the number of examples in node $p$. Define $\psi_i(w) = n_p l(w\cdot x_i,y_i) + \frac{\lambda}{2} \|w\|^2$. It is easy to check that
\begin{equation}
\grad\fhat_p(w) = \frac{1}{n_p} \sum_{i\in I_p} ( \grad\psi_i(w) - \grad\psi_i(w^r) + g^r).
\label{psigrad}
\end{equation}
Thus, plain SGD updates applied to $\fhat_p$ has the form
\begin{equation}
w = w - \eta ( \grad\psi_i(w) - \grad\psi_i(w^r) + g^r),
\label{svrg}
\end{equation}
which is precisely the update in SVRG. In particular, the single node ($P=1$) implementation of our method using plain SGD updates for optimizing $\fhat_p$ is very close to the SVRG method.\footnote{Note the subtle point that applying SVRG method on $\fhat_p$ is different from doing (\ref{svrg}), which corresponds to plain SGD. It is the former that assures {\it glrc} (in expectation).} While~\citet{johnson2013} motivate the update in terms of variance reduction, we derive it from a functional approximation viewpoint.

\subsection{Computation-Communication tradeoff}
\label{compcomm}

In this subsection we do a rough analysis to understand the conditions under which our method (FADL) is faster than the SQM method~\citep{chu2006,agarwal2011} (see Section~\ref{intro}). This analysis is only for understanding the role of various parameters and not for getting any precise comparison of the speed of the two methods.

Compared to the SQM method, FADL does a lot more computation (optimize $\fhat_p$) in each node. On the other hand FADL reaches a good solution using a much smaller number of outer iterations. Clearly, FADL will be attractive for problems with high communication costs, e.g., problems with a large feature dimension. For a given distributed computing environment and specific implementation choices, it is easy to do a rough analysis to understand the conditions in which FADL will be more efficient than SQM. Consider a distributed grid of nodes in an {\it AllReduce} tree. Let us use a common method such as TRON for implementing SQM as well as for ${\cal M}$ in FADL. Assuming that $T_{\rm SQM}^{\rm outer}>3.0T_{\rm FADL}^{outer}$ (where $T_{\rm FADL}^{\rm outer}$ and $T_{\rm SQM}^{outer}$ are the number of outer iterations required by SQM and FADL), we can do a rough analysis of the costs of SQM and FADL (see appendix A for details) to show that FADL will be faster when the following condition is satisfied.
\begin{equation}
\frac{nz}{m} < \frac{\gamma P}{2\khat}
\label{comm}
\end{equation}
where: $nz$ is the number of nonzero elements in the data, i.e., $\{x_i\}$; $m$ is the feature dimension; $\gamma$ is the relative cost of communication to computation (e.g. $100-1000$); $P$ is the number of nodes; and $\khat$ is the number of inner iterations of FADL. Thus, the larger the dimension ($m$) is, and the higher the sparsity in the data is, FADL will be better than SQM.

\section{Experiments}
\label{expts}

\def\grad{\nabla}
\def\wtilde{\tilde{w}}
\def\Cone{{\cal{C}}^1}
\def\kappap{\kappa^\prime}
\def\Lhat{\hat{L}}
\def\fhat{\hat{f}}
\def\what{\hat{w}}
\def\dhat{\hat{d}}
\def\mysgn{\operatorname{sgn}}

In this section, we demonstrate the effectiveness of our method by comparing it against several existing distributed training methods on five large data sets. 
We first discuss our experimental setup. We then briefly list each method considered and then do experiments to decide the best overall setting for each method. This applies to our method too, for which the setting is mainly decided by the choice made for the function approximation, $\fhat_p$; see Subsection~\ref{subsec:fhatp} for details of these choices. Finally, we compare, in detail, all the methods under their best settings. This study clearly demonstrates scenarios under which our method performs better than other methods.

\subsection{Experimental Setup}
\label{subsec:setup}

We ran all our experiments on a Hadoop cluster with $379$ nodes and 10 Gbit interconnect speed. Each node has Intel (R) Xeon (R) E5-2450L (2 processors) running at 1.8 GHz.
Since iterations in traditional {\it MapReduce} are slower (because of job setup and disk access costs), as in Agarwal et al.~\citep{agarwal2011}, we build an {\it AllReduce} binary tree between the mappers\footnote{Note that we do not use the pipelined version and hence we incur an extra multiplicative $logP$ cost in communication.}. The communication bandwidth is $1 Gbps$ (gigabits per sec).

\begin{table}[ht]
\caption{Properties of datasets.} % title of Table
\centering % used for centering table
\begin{tabular}{c c c c c} % centered columns (4 columns)
\hline\hline %inserts double horizontal lines
Dataset & \#Examples ($n$) & \#Features ($m$) & \#Non-zeros ($nz$) & $\lambda$ \\ %[0.5ex] % inserts table
%heading
\hline
{\it{kdd2010}} & $8.41\times 10^6$ & $20.21\times 10^6$ & $0.31\times 10^9$ & $1.25\times 10^{-6}$ \\
{\it{url }} & $1.91\times 10^6$ &  $3.23\times 10^6$ &  $0.22\times 10^9$ & $0.11\times10^{-6}$ \\
{\it{webspam}} & $0.35\times 10^6$ & $16.6\times 10^6$ & $0.98\times 10^9$  & $1.0\times10^{-4}$ \\
{\it{mnist8m}} & $8.1\times 10^6$ & 784 & $6.35\times 10^9$ & $1.0\times10^{-4}$ \\
{\it{rcv}} & $0.5\times 10^6$ & $47236$ & $0.50\times 10^8$ & $1.0\times10^{-4}$ \\
\hline
\end{tabular}
\label{tab:params}
\end{table}

\noindent {\bf Data Sets.} We consider the following publicly available datasets having a large number of examples:\footnote{These datasets are available at: \url{http://www.csie.ntu.edu.tw/~cjlin/libsvmtools/datasets/}. For {\it mnist8m} we solve the binary problem of separating class ``3" from others.}
{\it{kdd2010}}, {\it{url}}, {\it{webspam}}, {\it{mnist8m}} and {\it{rcv}}. Table~\ref{tab:params} shows the numbers of examples, features, nonzero in data matrix and the values of regularizer $\lambda$ used. The regularizer for each dataset is chosen to be the optimal value that gives the best performance on a small validation set. We use these datasets mainly to illustrate the validity of theory, and its utility to distributed machine learning. In real scenarios of Big data, the datasets are typically much larger. Note that {\it kdd2010}, {\it url} and {\it webspam} are large dimensional ($m$ is large) while {\it mnist8m} and {\it rcv} are low/medium dimensional ($m$ is not high). This division of the datasets is useful because communication cost in example-partitioned distributed methods is mainly dependent on $m$ (see Appendix A) and so these datasets somewhat help see the effect of communications cost.

We use the {\it squared-hinge loss} function for all the experiments. Unless stated differently, for all numerical optimizations we use the Trust Region Newton method (TRON) proposed in~\citet{lin2008}.

\noindent {\bf Evaluation Criteria.} We use the relative difference to the optimal function value and the Area under Precision-Recall Curve (AUPRC) as the evaluation criteria. The former is calculated as $(f-f^*)/f^*$ in log scale, where $f^*$ is the optimal function value obtained by running the {\it{TERA}} algorithm (see below) for a very large number of iterations.

\subsection{Methods for comparison}
\label{subsec:methods}

We compare the following methods.
\begin{itemize}
\item {\it{TERA:}}
The Terascale method (TERA)~\citep{agarwal2011} is the best representative method from the SQM class~\citep{chu2006}. It can be considered as the state-of-the-art distributed solver and therefore an important baseline.
\item {\it{ADMM:}}
We use the example partitioning formulation of the Alternating Direction Method of Multipliers (ADMM)~\citep{Boyd2011, zhang2012}. ADMM is a dual method which is very different from our primal method; however, like our method, it solves approximate problems in the nodes and iteratively reaches the full batch solution.
\item {\it{CoCoA:}}
This method~\citep{jaggi2014} represents the class of distributed dual methods~\citep{pechyony2011, yang2013a, yang2013b, jaggi2014} that, in each outer iteration, solve (in parallel) several local dual optimization problems.
\item {\it{Our method (FADL):}}
This is our method described in detail in Section~\ref{distr} and more specifically, in Algorithm~\ref{alg:fadl}.
\end{itemize}

\subsection{Study of TERA}
\label{subsec:tera}

\begin{figure}[t]
\centering
\subfigure[{\it{kdd2010}} - 8 nodes]{
\includegraphics[width=0.46\linewidth]{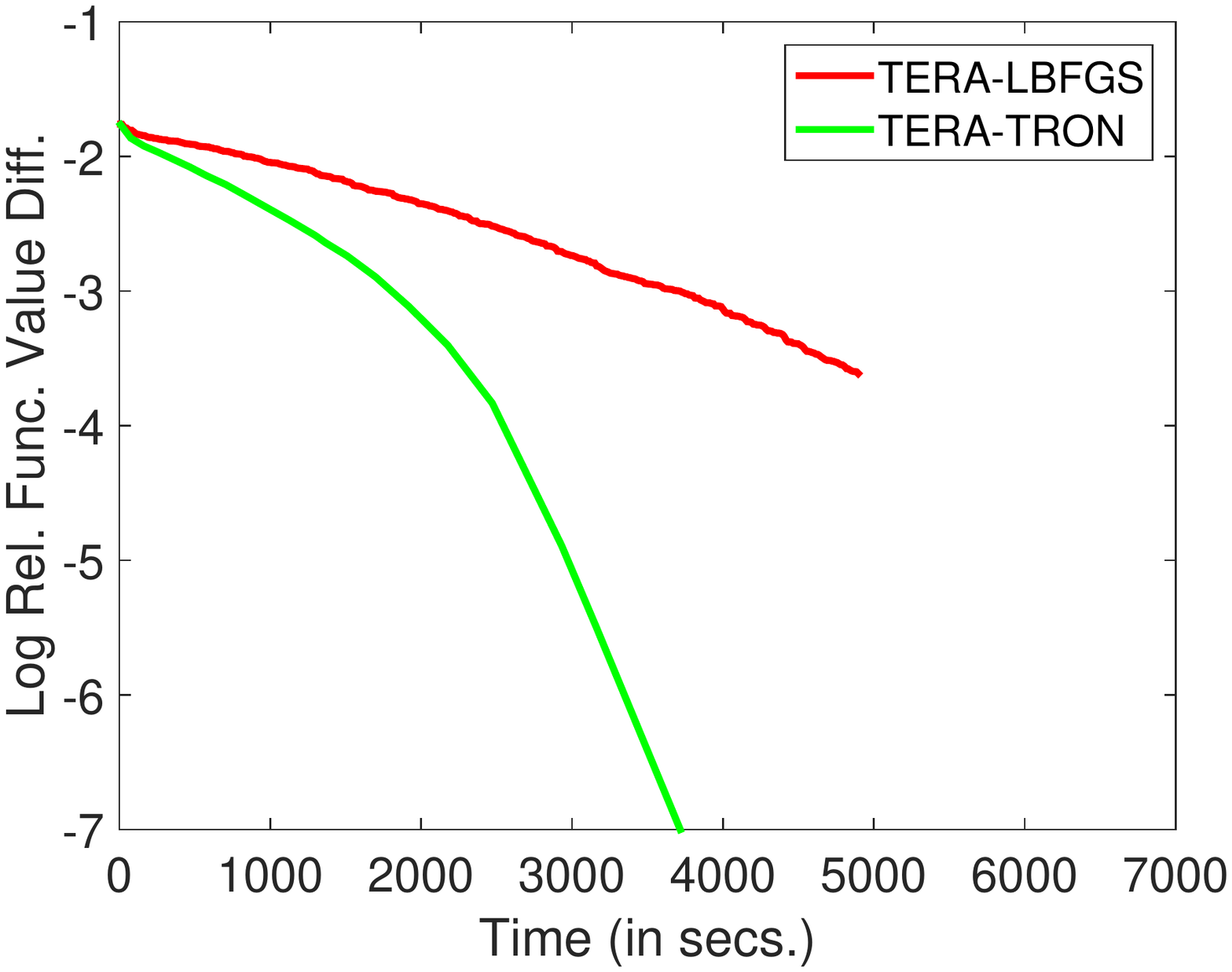}
}
\subfigure[{\it{kdd2010}} - 128 nodes]{
\includegraphics[width=0.46\linewidth]{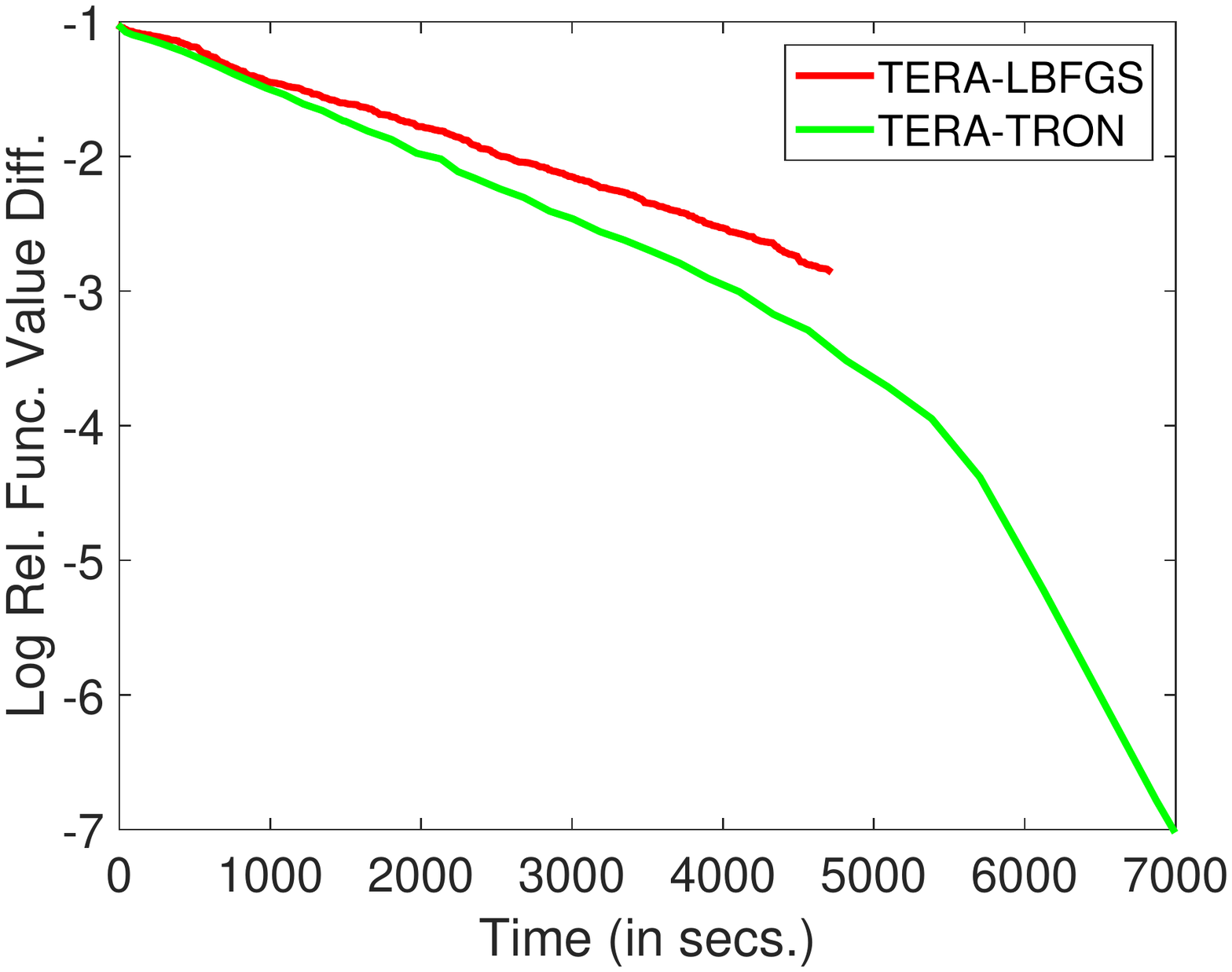}
}

\caption{Plots showing the time efficiency of TERA methods for {\it kdd2010}.}
\label{fig:tera}
\end{figure}

A key attractive property of TERA is that the number of outer iterations pretty much remains constant with respect to the number of distributed nodes used. As recommended by~\citet{agarwal2011}, we find a local weight vector per node by minimizing the local objective function (based only on the examples in that node) using five epochs of SGD~\citep{bottou2010}. (The optimal step size is chosen by running SGD on a subset of data.) We then average the weights from all the nodes (on a per-feature basis as explained in~\citet{agarwal2011}) and use the averaged weight vector to warm start {\it{TERA}}.\footnote{We use this inexpensive initialization for FADL and ADMM too. It is not applicable to CoCoA. Because of this, CoCoA starts with a different primal objective function value than others.} \citet{agarwal2011} use the LBFGS method as the trainer whereas we use TRON. To make sure that this does not lead to bias, we try both, TERA-LBFGS and TERA-TRON. Figure~\ref{fig:tera} compares the progress of objective function for these two choices. Clearly, TERA-TRON is superior. We observe similar behavior on the other datasets also. Given this, we restrict ourselves to TERA-TRON and simply refer to it as TERA.

\subsection{Study of ADMM}
\label{subsec:admm}

The ADMM objective function~\citep{Boyd2011} has a quadratic proximal term called augmented Lagrangian with a penalty parameter $\rho$ multiplying it. In general, the performance of ADMM is very sensitive to the value of $\rho$ and hence making a good choice for it is crucial. We consider three methods for choosing $\rho$.

Even though there is no supporting theory, Boyd et al~\citep{Boyd2011} suggest an approach by which $\rho$ is adapted in each iteration; see Equation (3.13) in Section 3.4.1 of that paper. We will refer to this choice as {\it Adap}.

Recently, \citet{deng2012} proved a linear rate of convergence for ADMM under assumptions A1 and A2 (see Section~\ref{general}) on ADMM functions. As a result, their analysis also hold for the objective function in~(\ref{risk}). They also give an analytical formula to set $\rho$ in order to get the best theoretical linear rate constant. We will refer to this choice of $\rho$ as {\it Analytic}.

We also consider a third choice, {\it{ADMM-Search}} in which, we start with the value of $\rho$ given by {\it Analytic}, choose several values of $\rho$ in its neighborhood and select the best $\rho$ by running ADMM for 10 iterations and looking at the objective function value.  Note that this step takes additional time and causes a late start of ADMM. 

\begin{figure}[t]
\centering
\subfigure[{\it{kdd2010}} - 8 nodes]{
\includegraphics[width=0.46\linewidth]{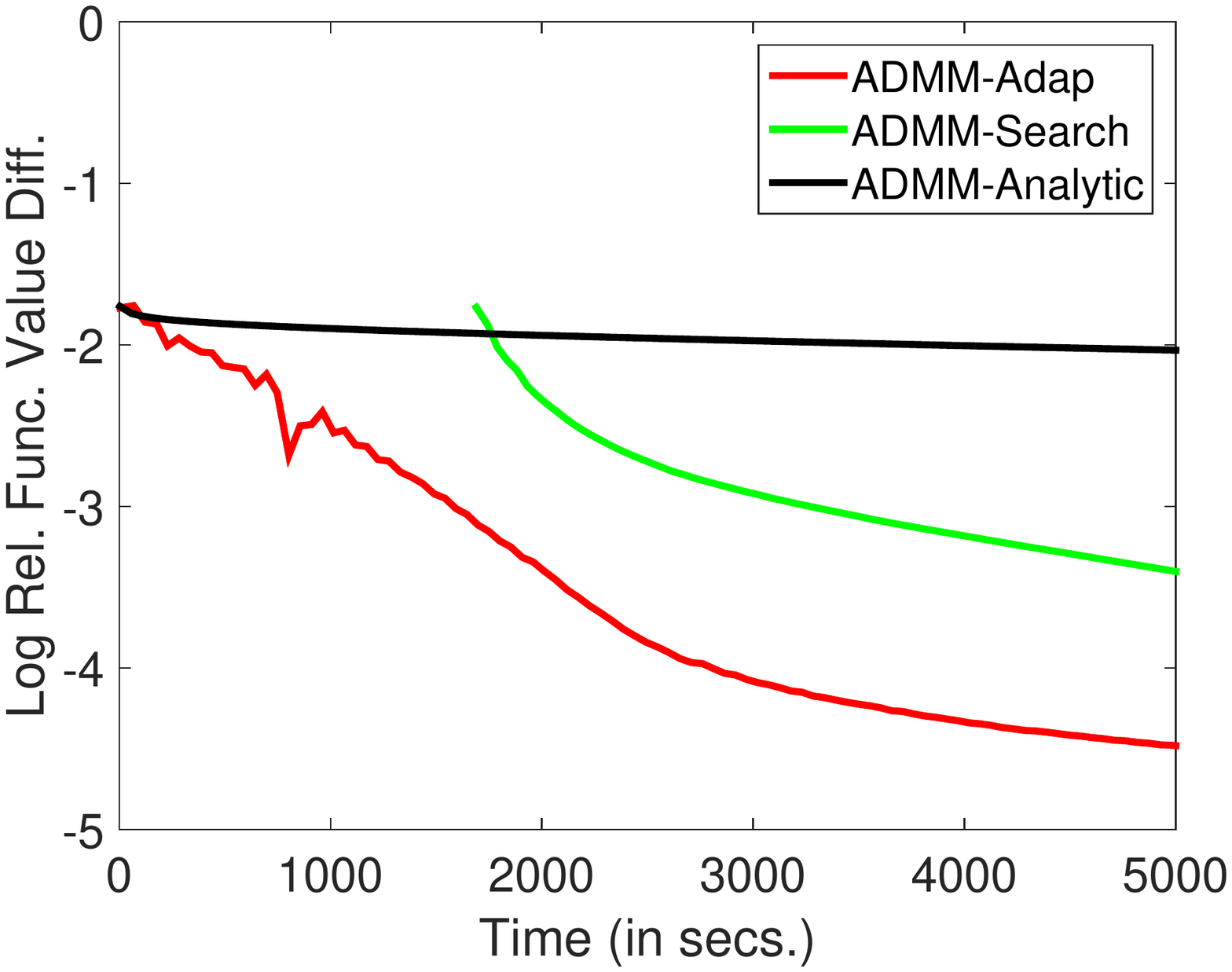}
}
\subfigure[{\it{kdd2010}} - 128 nodes]{
\includegraphics[width=0.46\linewidth]{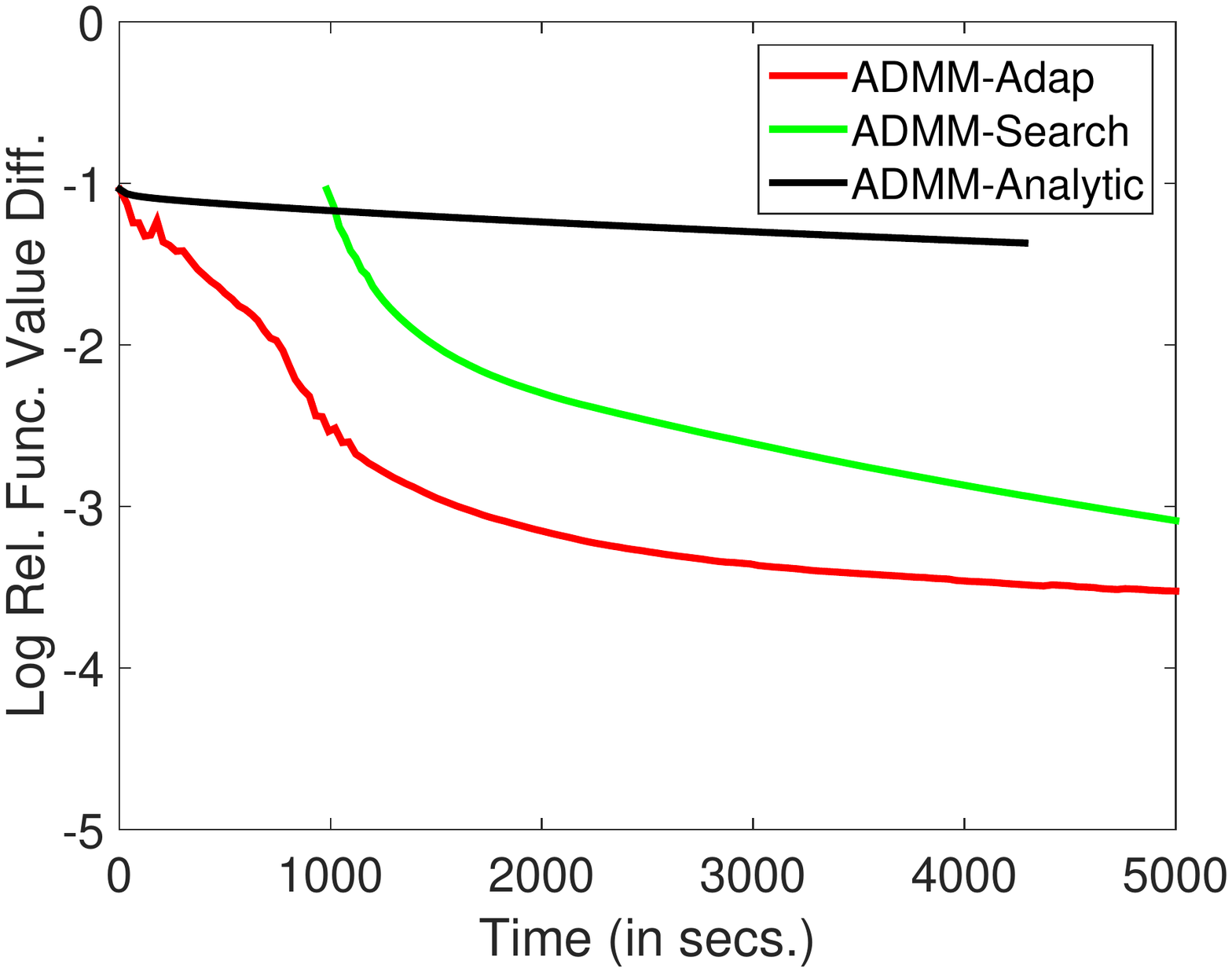}
}

\caption{Plots showing the time efficiency of ADMM methods for {\it kdd2010}.}
\label{fig:admm}
\end{figure}

Figure~\ref{fig:admm} compares the progress of the training objective function for the three choices on {\it kdd2010} for $P=8$ and $P=128$.
{\it Analytic} is an order of magnitude slower than the other two choices. {\it Search} works well. However, a significant amount of initial time is spent on finding a good value for $\rho$, thus making the overall approach slow. {\it Adap} comes out to be the best performer among the three choices. Similar observations hold for other datasets and other choices of $P$. So, for ADMM, we will fix {\it Adap} as the way of choosing $\rho$ and refer to ADMM-{\it Adap} simply as ADMM. 

It is also worth commenting on methods related to ADMM. Apart from ADMM,~\citet{bertsekas1997} discuss several other classic optimization methods for separable convex programming, based on proximal and Augmented Lagrangian ideas which can be used for distributed training of linear classifiers. ADMM represents the best of these methods. Also, Gauss-Seidel and Jacobi methods given in~\citet{bertsekas1997} are related to feature partitioning, which is very different from the example partitioning scenario studied in this paper. Therefore we do not consider these methods.

\subsection{Study of CoCoA}
\label{subsec:dbca}

\begin{figure}[t]
\centering
\subfigure[{\it{kdd2010}} - 8 nodes]{
\includegraphics[width=0.46\linewidth]{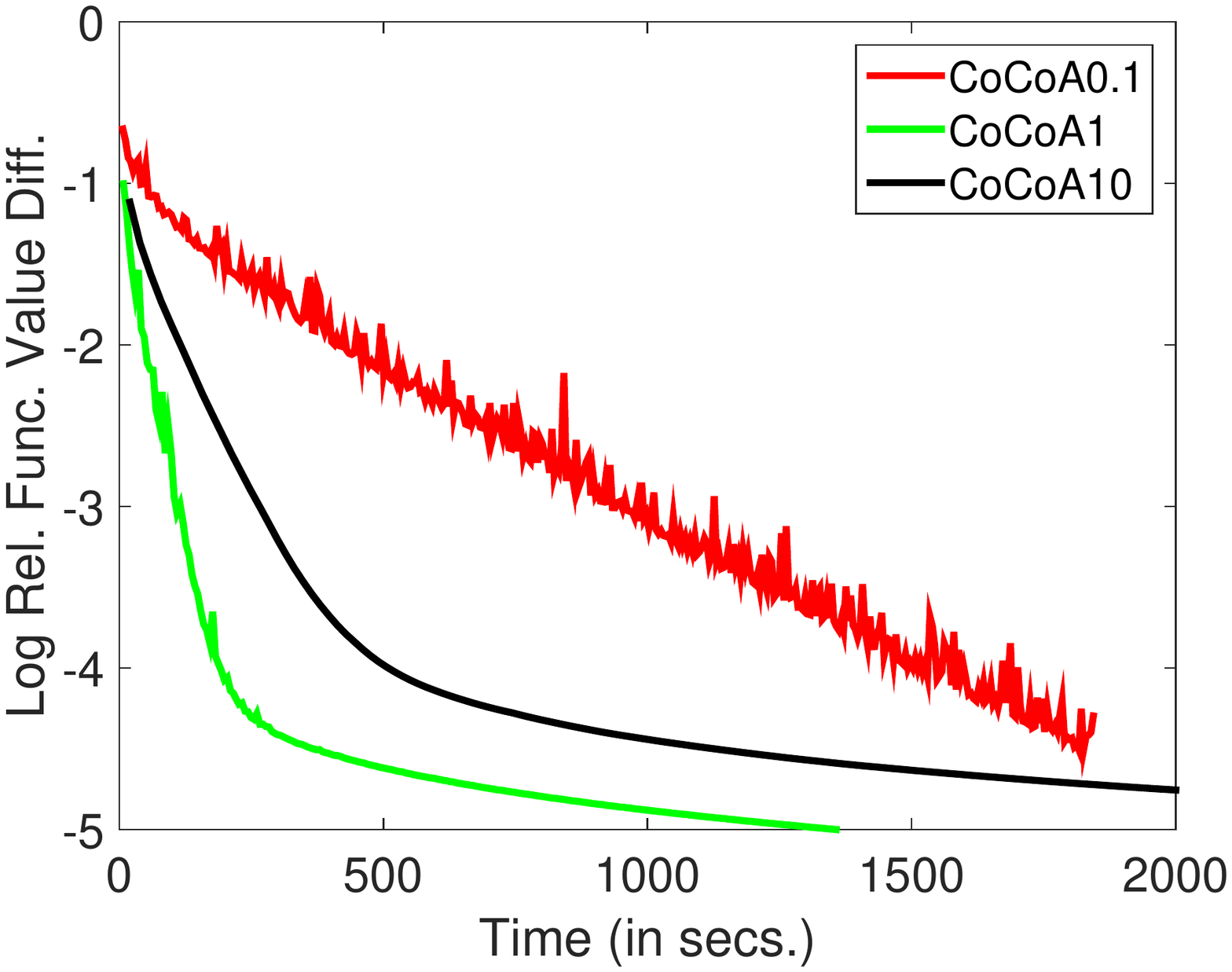}
}
\subfigure[{\it{kdd2010}} - 128 nodes]{
\includegraphics[width=0.46\linewidth]{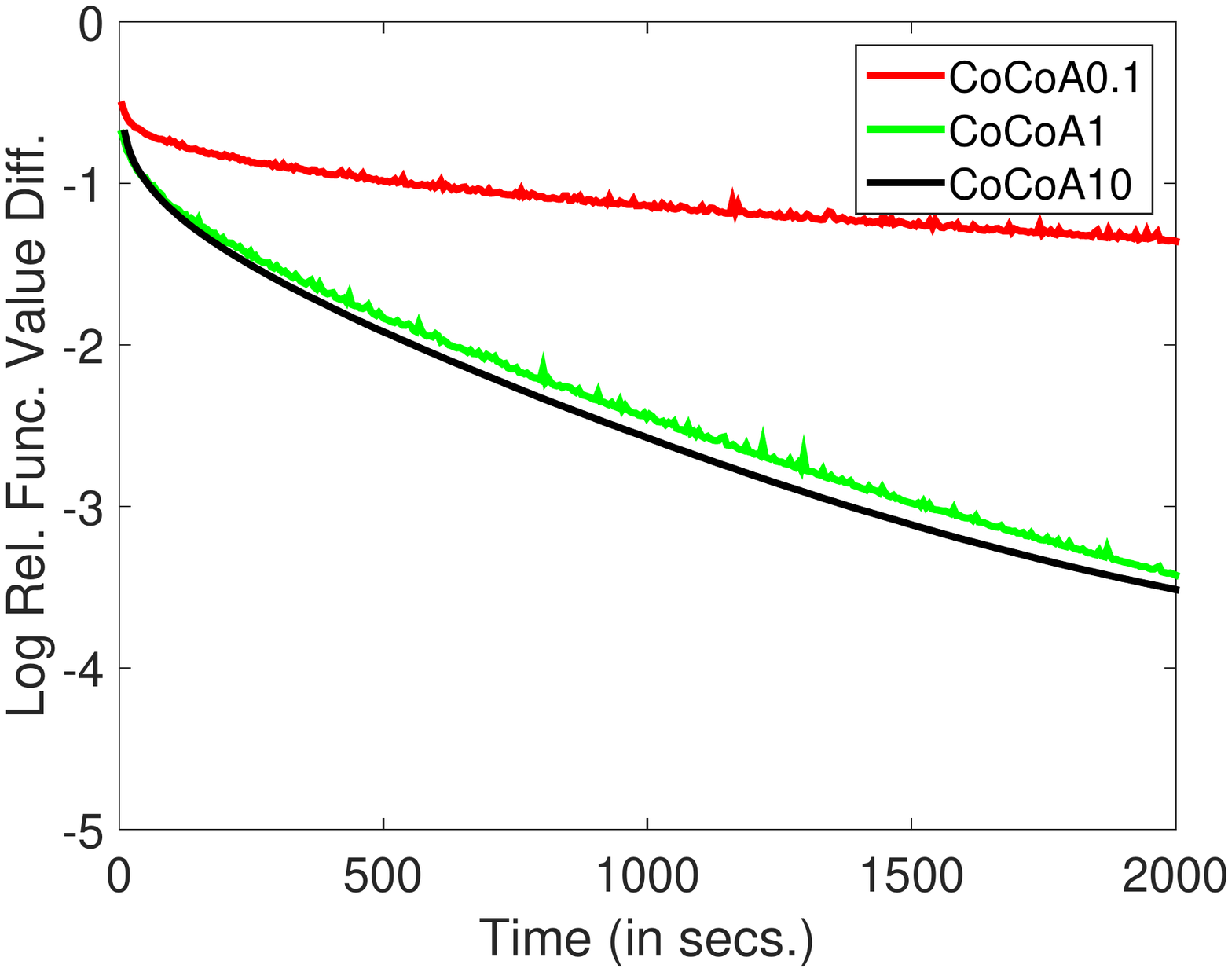}
}

\caption{Plots showing the time efficiency of CoCoA settings for {\it kdd2010}.}
\label{fig:cocoa}
\end{figure}

In CoCoA~\citep{jaggi2014} the key parameter is the approximation level of the inner iterations used to solve each projected dual sub-problem. The number of epochs of coordinate dual ascent inner iterations plays a crucial role. We try the following choices for it: 0.1, 1 and 10. Figure~\ref{fig:cocoa} compares the progress of the objective function on {\it kdd2010} for two choices of nodes, $P=8$ and $P=128$. We find the choice of 1 epoch to work well reasonably consistently over all the five datasets and varying number of nodes. So we fix this choice and refer to the resulting method simply as CoCoA. Note in Figure~\ref{fig:cocoa} that the (primal) objective function does not decrease continuously with time. This is because it is a dual method and so monotone descent of the objective function is not assured.\footnote{The same comment holds for ADMM which is also a dual method; see for examples the jumps in objective function values for ADMM in Figures~\ref{fig:commpass1} and \ref{fig:timepass1} for {\it kdd2010}.}

\subsection{Study of function approximation methods}
\label{subsec:our}

Recall from Subsection~\ref{subsec:fhatp} the various choices that we suggested for $\fhat_p$. 
We are yet to implement and study the BFGS approximation; we leave it out for future work.
Our original work~\citep{dhruv2013} focused on the Linear approximation, but we found the Quadratic, Hybrid and Nonlinear choices to work much better. So we study only these three methods. The implementation of these methods is as described in Subsection~\ref{subsec:practical} and Algorithm~\ref{alg:fadl}.

We will also include another method called SSZ in the analysis of this subsection. This is the Newton based method described in~\citet{sharir2014}. Even though this method is very different in spirit from our method, it uses an approximation similar to our Nonlinear idea. SSZ is a non-monotone method that is based on fixed step sizes, with a probabilistic convergence theory.  The method has two parameters, $\mu$ and $\eta$ in the approximation; $\mu$ is the coefficient for the proximal term and $\eta$ is used for defining the direction. \citet{sharir2014} do not prove convergence for any possible choices of $\mu$ and $\eta$ values. We go with their practical recommendation and employ $\mu=3\lambda$ and $\eta=1$ for their method.

\begin{figure}[t]
\centering
\subfigure[{\it{kdd2010}} - 8 nodes]{
\includegraphics[width=0.46\linewidth]{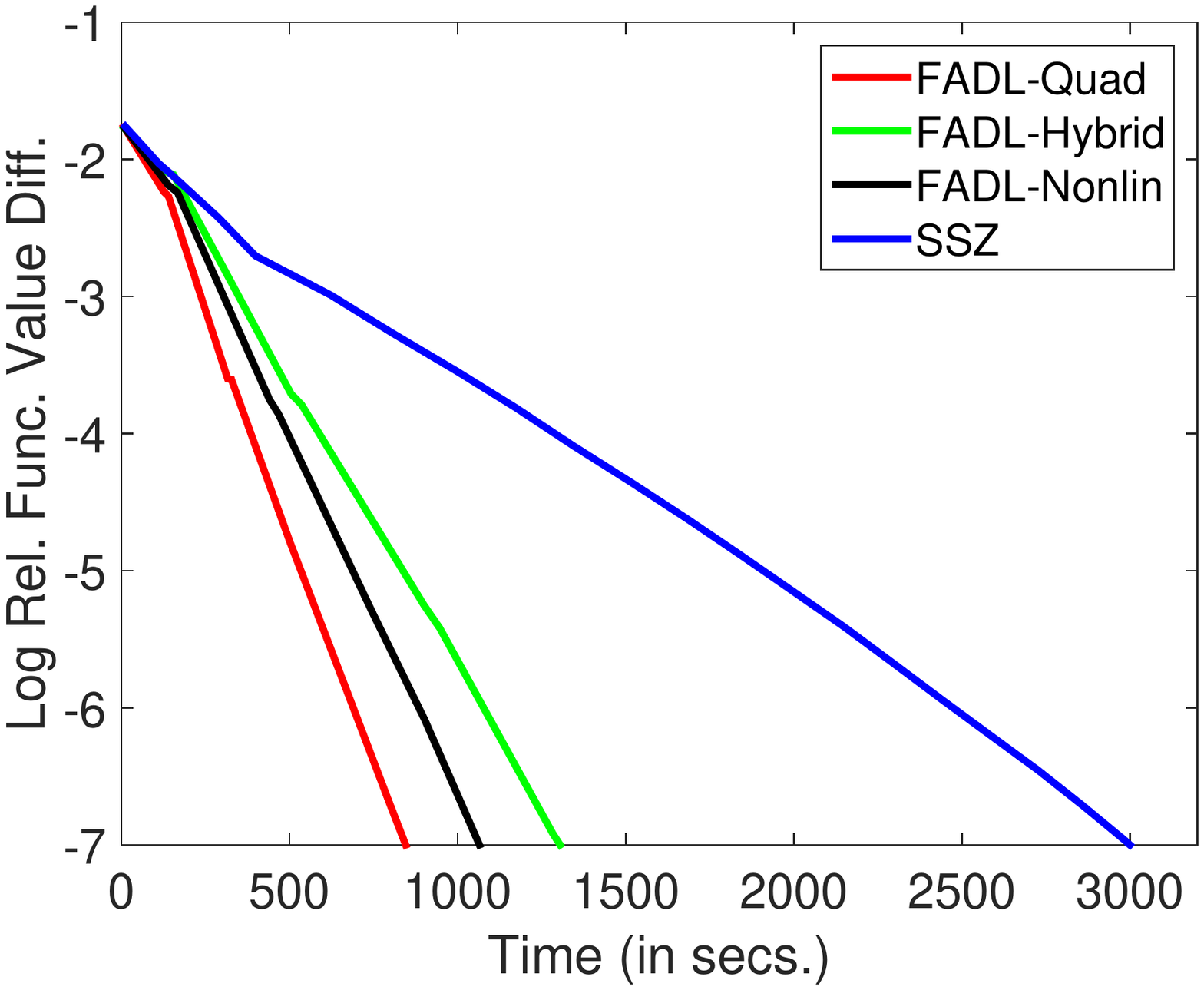}
}
\subfigure[{\it{kdd2010}} - 128 nodes]{
\includegraphics[width=0.46\linewidth]{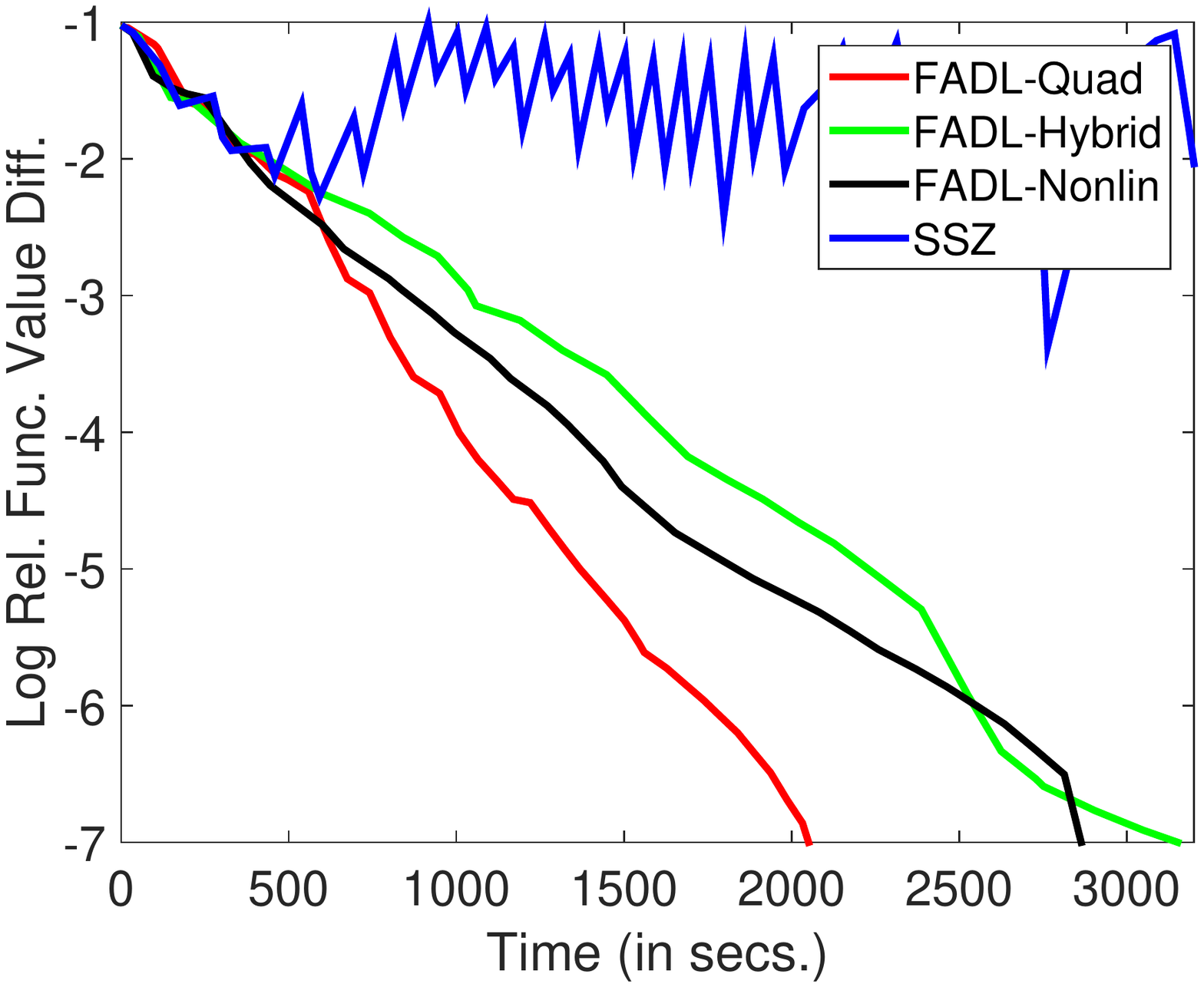}
}

\caption{Plots showing the time efficiency of FADL and SSZ methods for {\it kdd2010}.}
\label{fig:ipm}
\end{figure}

Figure~\ref{fig:ipm} compares the progress of the training objective function for various choices of $\fhat_p$, and SSZ. SSZ shows unstable behavior when the number of nodes is large. We have observed similar behavior in other datasets too. Among our methods, the quadratic approximation for $\fhat_p$ gives the best performance, although the Hybrid and Nonlinear approximations also do quite well. We observe this reasonably consistently in other datasets too. Hence, from the set of methods considered in this subsection we choose FADL-Quadratic approximation as the only method for further analysis, and simply refer to this method as FADL hereafter.

Why does the quadratic approximation do better than hybrid and nonlinear approximations? We do not have a precise answer to this question, but we give some arguments in support.
In each outer iteration, the function approximation idea is mainly used to get a good direction. Recall from Subsection~\ref{subsec:fhatp} that different choices use different approximations for $\Ltilde_p$ and $\Lhat_p$. Using the same ``type" (meaning linear, nonlinear or quadratic) for both, $\Ltilde_p$ and $\Lhat_p$ is possibly better for direction finding. Second, the direction finding could be more sensitive to the nonlinear approximation compared to the quadratic approximation; this could become more severe as the number of nodes becomes larger. Literature shows that quadratic approximations have good robustness properties; for example, subsampling in Hessian computation~\citep{byrd2012} doesn't worsen direction finding much.

\subsection{Comparison of FADL against TERA, ADMM and CoCoA}
\label{subsec:overallcomp}

\begin{figure}[H]
\centering
\subfigure[{\it{kdd2010}} - 8 nodes]{
\includegraphics[width=0.46\linewidth]{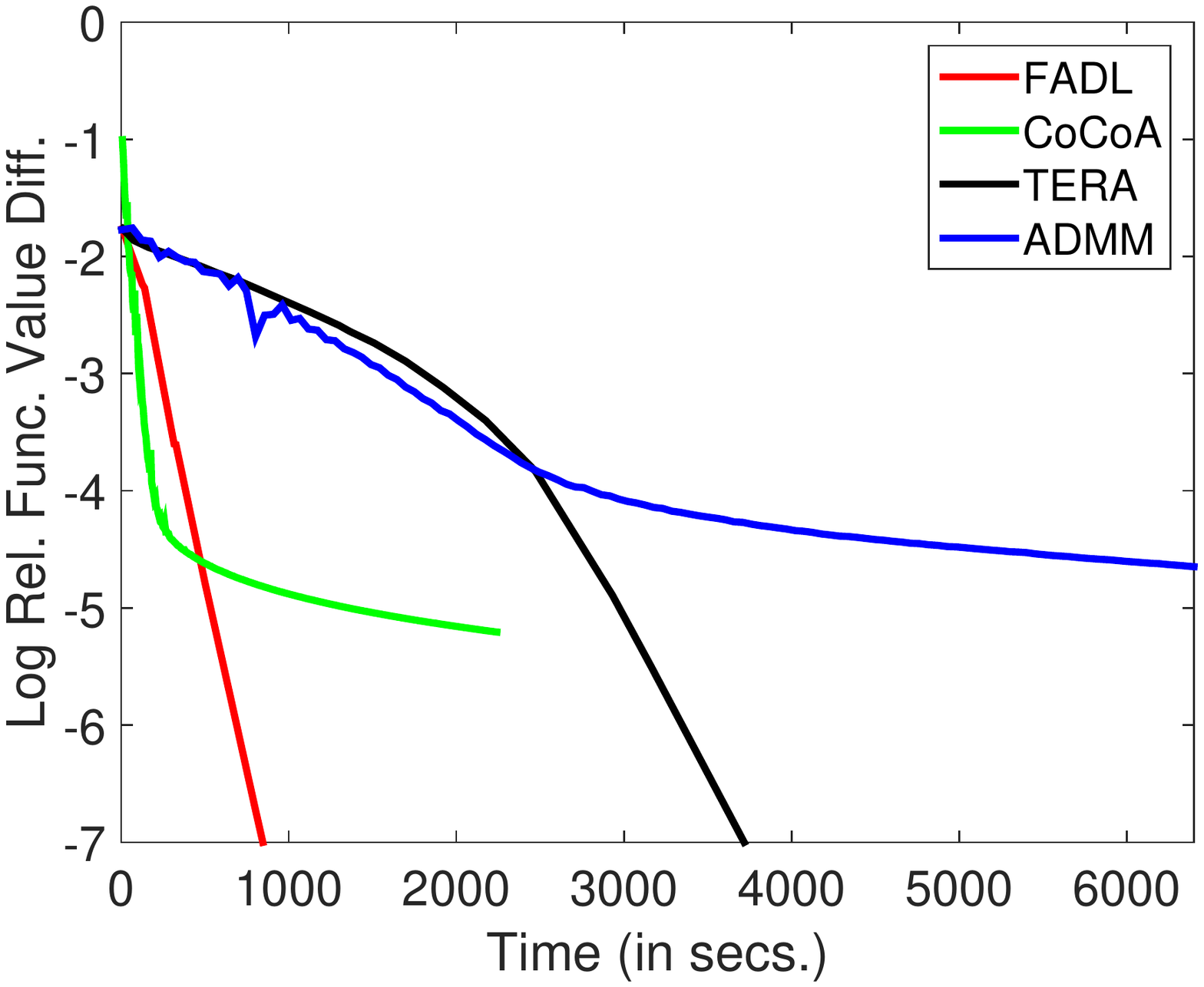}
}
\subfigure[{\it{kdd2010}} - 128 nodes]{
\includegraphics[width=0.46\linewidth]{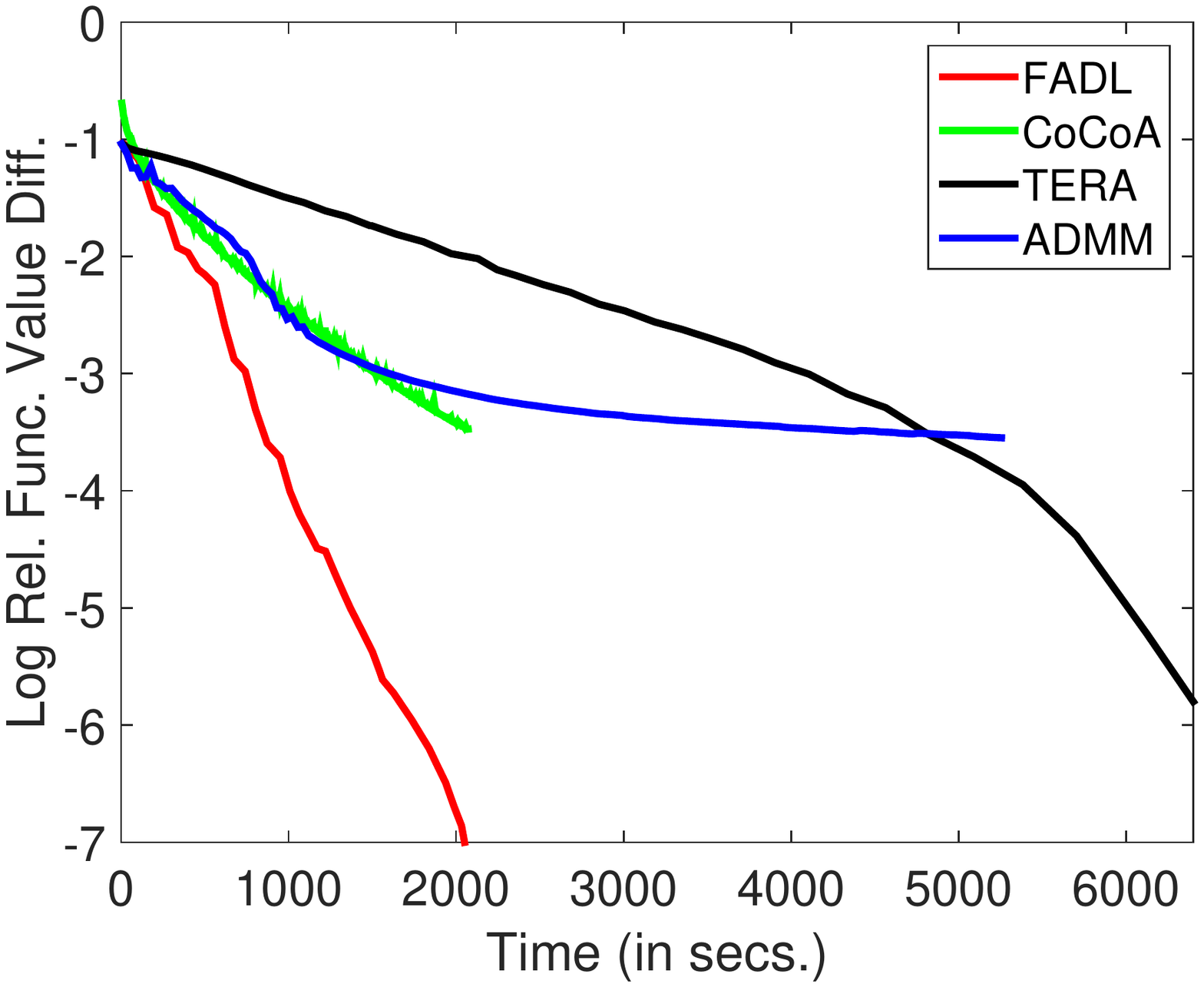}
}

\subfigure[{\it{url}} - 8 nodes]{
\includegraphics[width=0.46\linewidth]{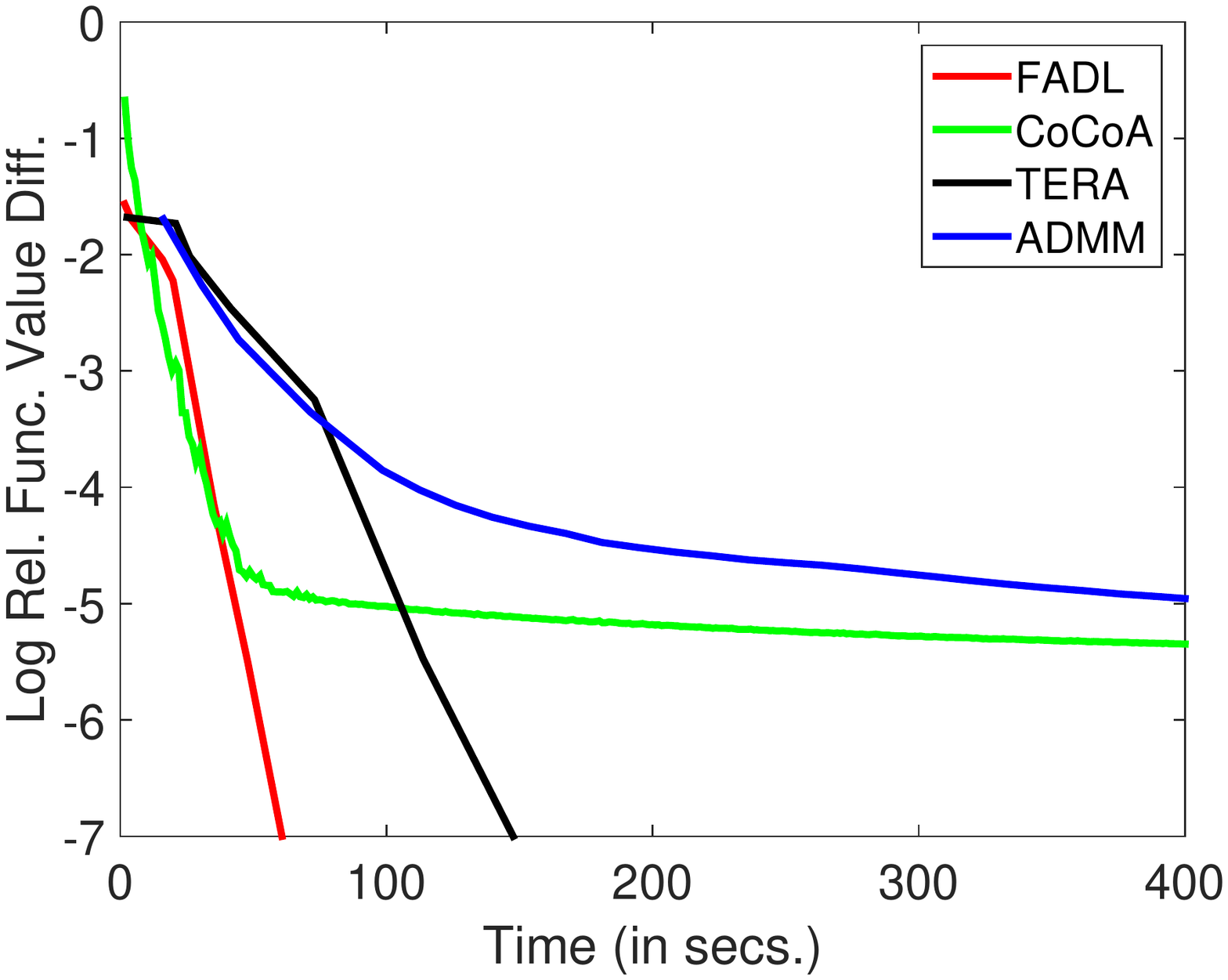}
}
\subfigure[{\it{url}} - 128 nodes]{
\includegraphics[width=0.46\linewidth]{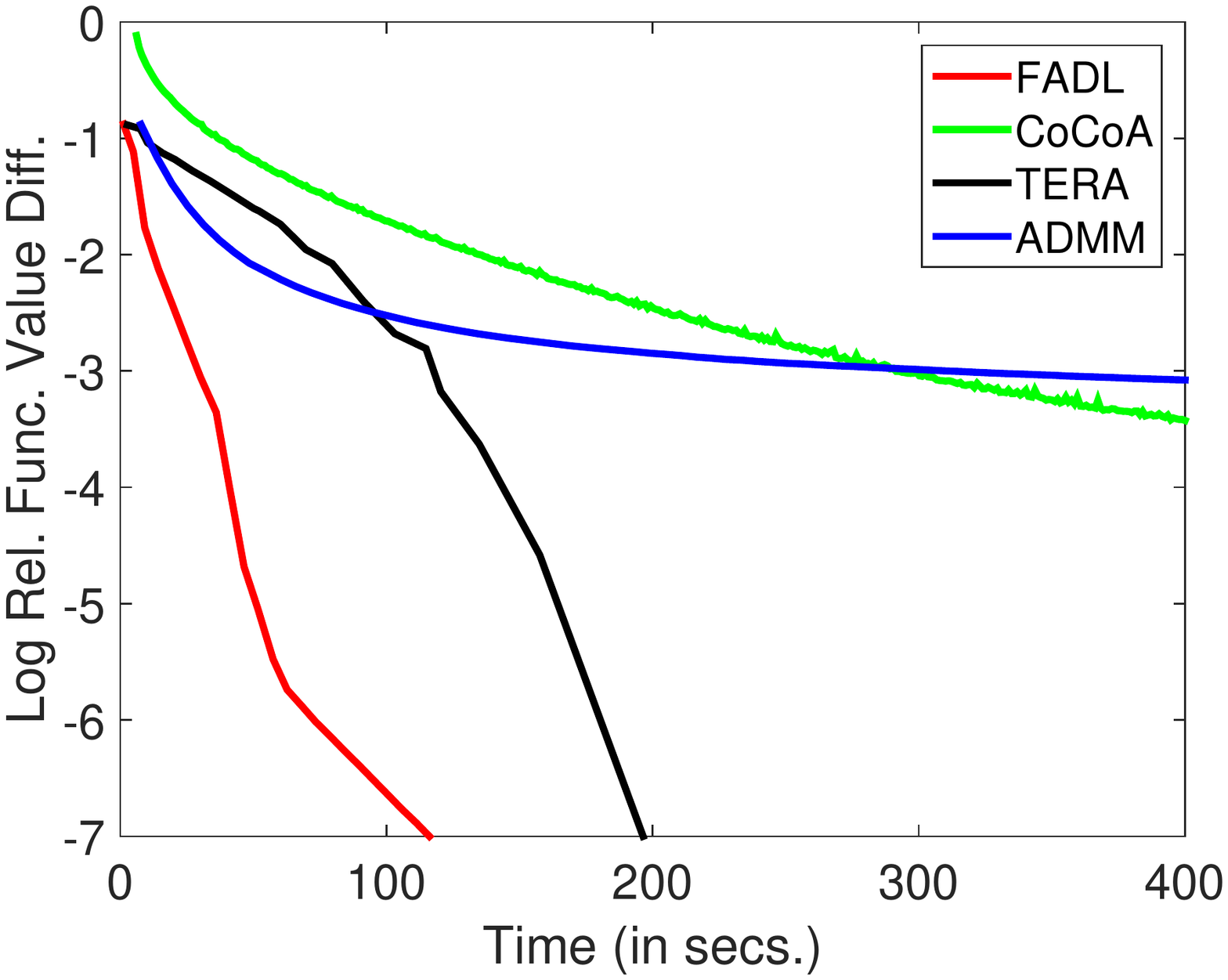}
}

\subfigure[{\it{webspam}} - 8 nodes]{
\includegraphics[width=0.46\linewidth]{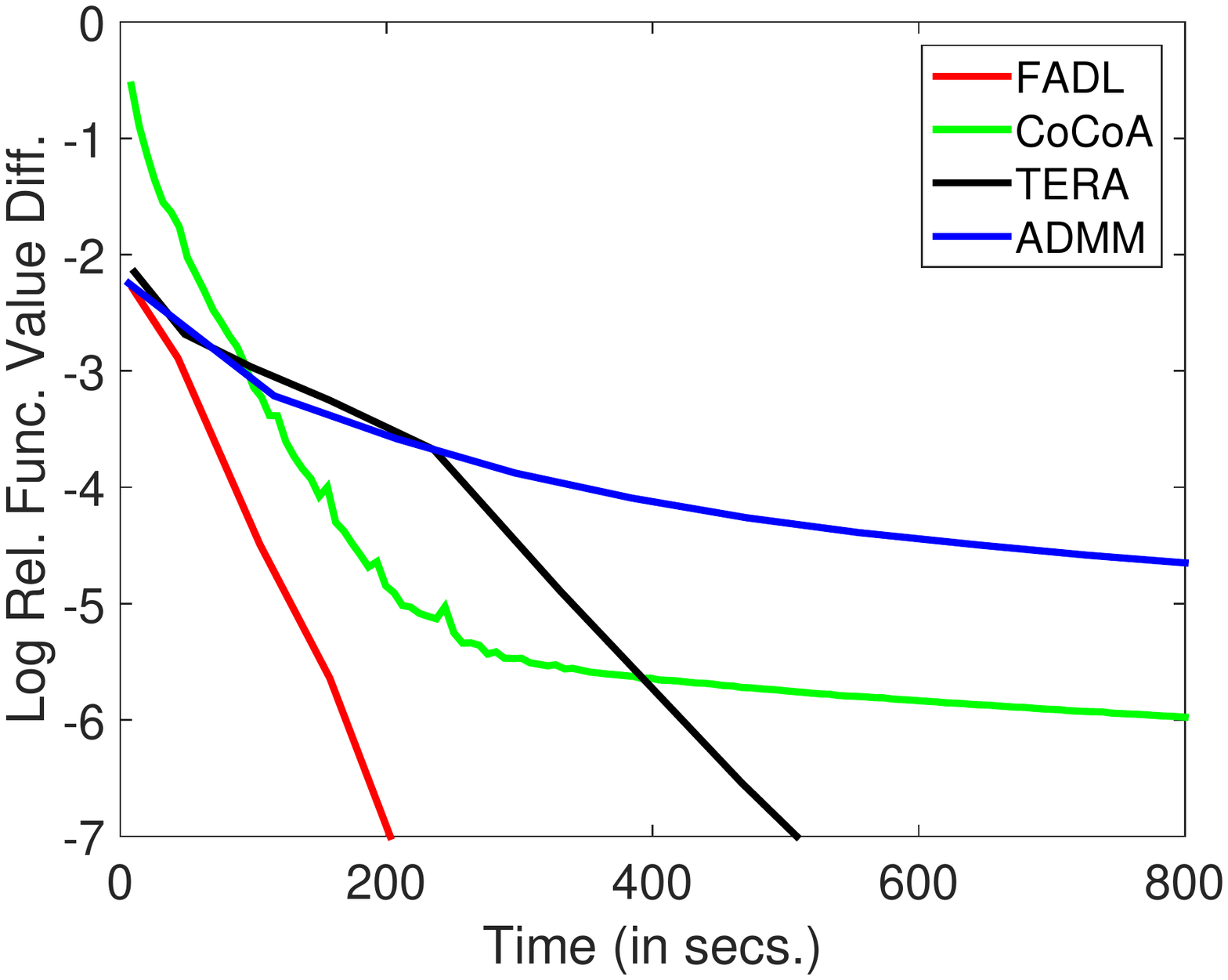}
}
\subfigure[{\it{webspam}} - 128 nodes]{
\includegraphics[width=0.46\linewidth]{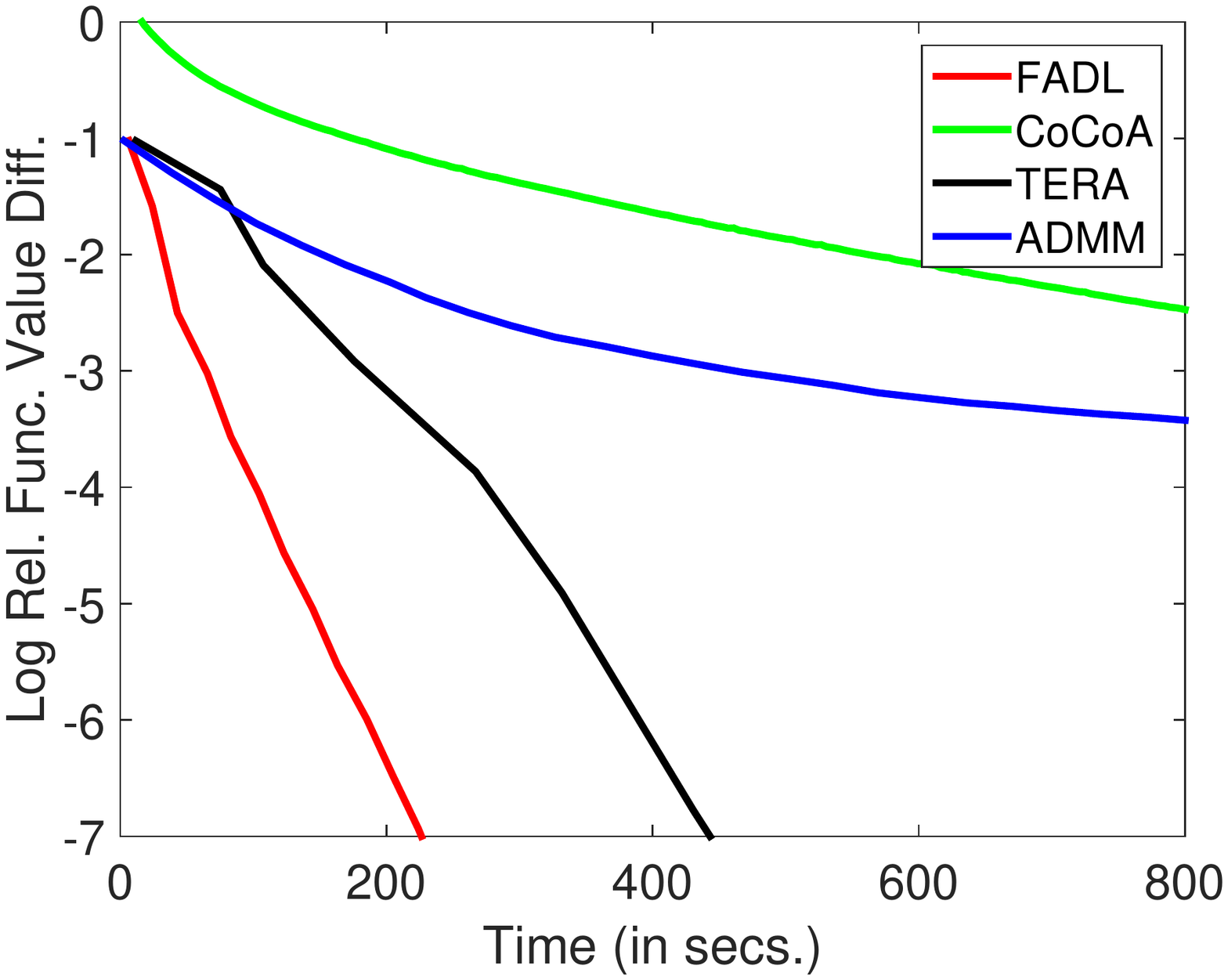}
}

\caption{Plots showing the time efficiency of various methods for the three high dimensional datasets.}
\label{fig:timepass1}
\end{figure}

\begin{figure}[H]
\centering

\subfigure[{\it{mnist8m}} - 8 nodes]{
\includegraphics[width=0.46\linewidth]{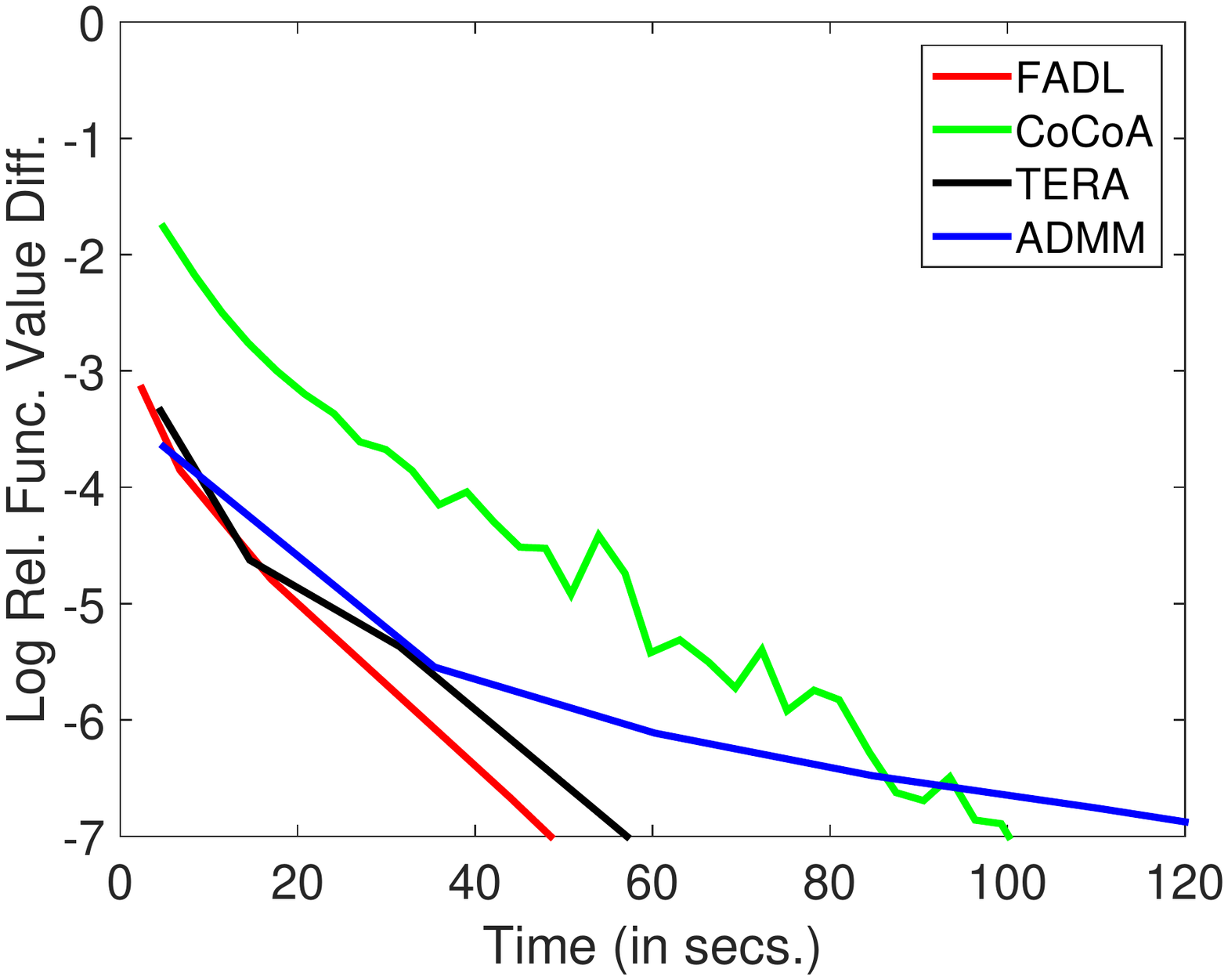}
}
\subfigure[{\it{mnist8m}} - 128 nodes]{
\includegraphics[width=0.46\linewidth]{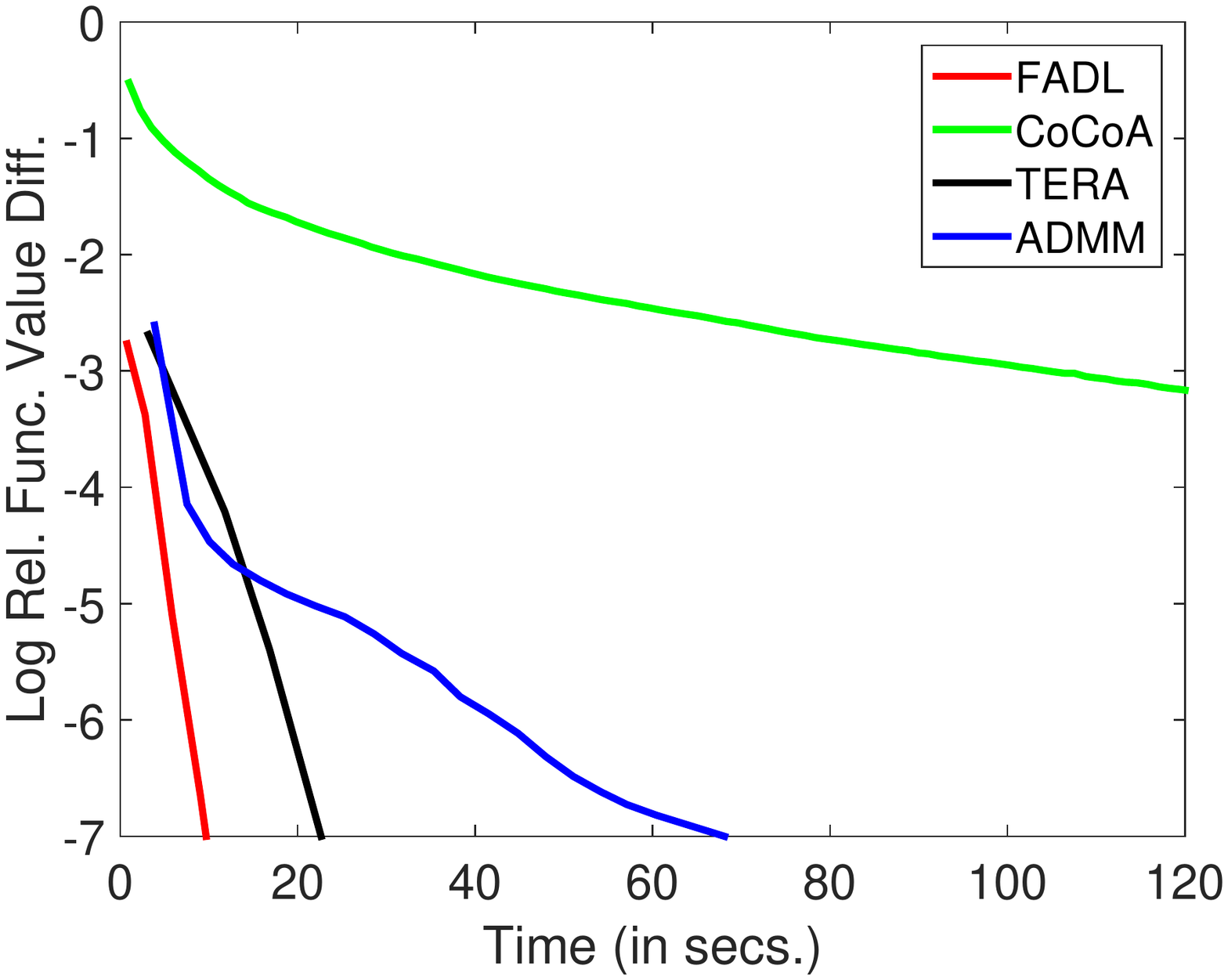}
}

\subfigure[{\it{rcv}} - 8 nodes]{
\includegraphics[width=0.46\linewidth]{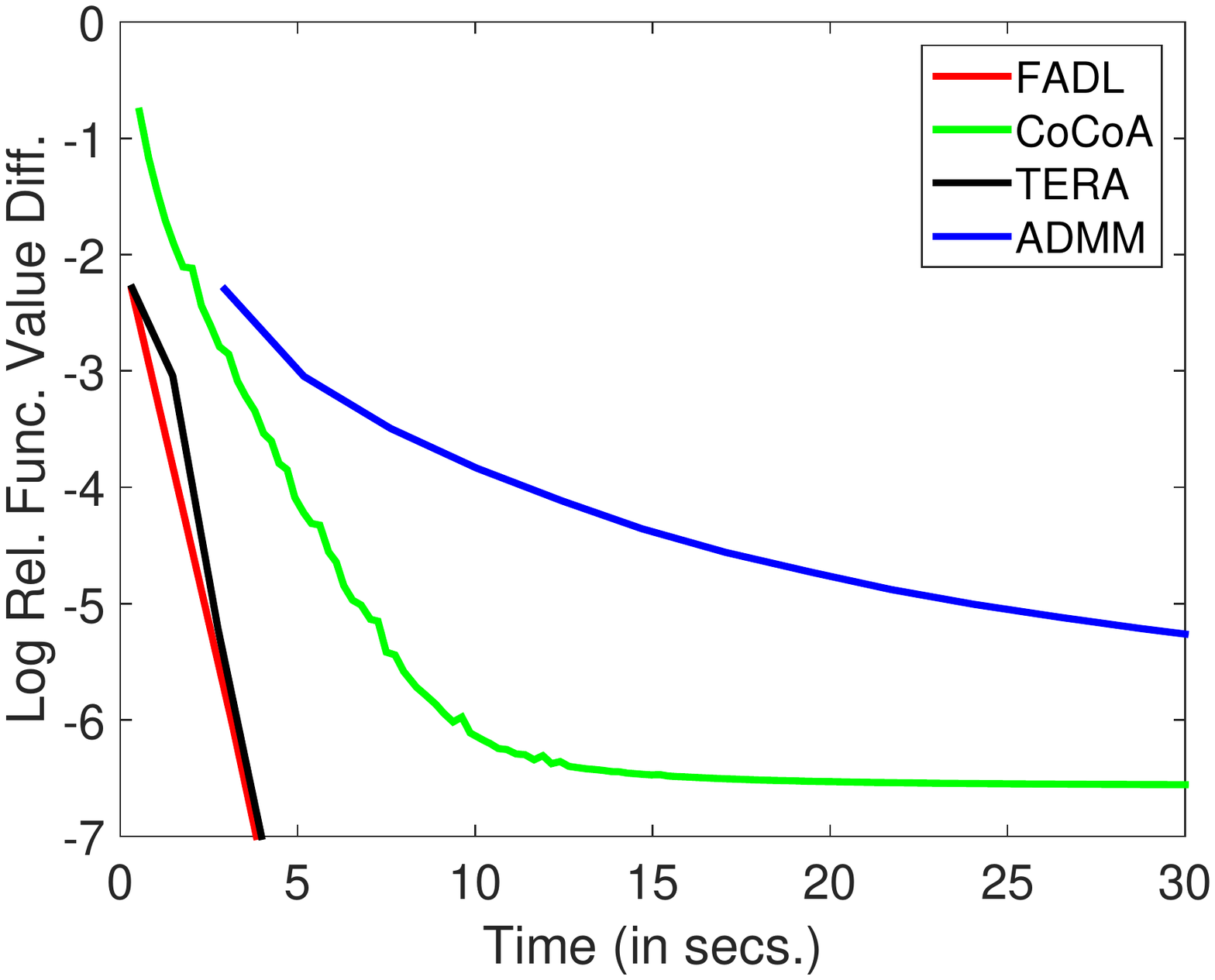}
}
\subfigure[{\it{rcv}} - 128 nodes]{
\includegraphics[width=0.46\linewidth]{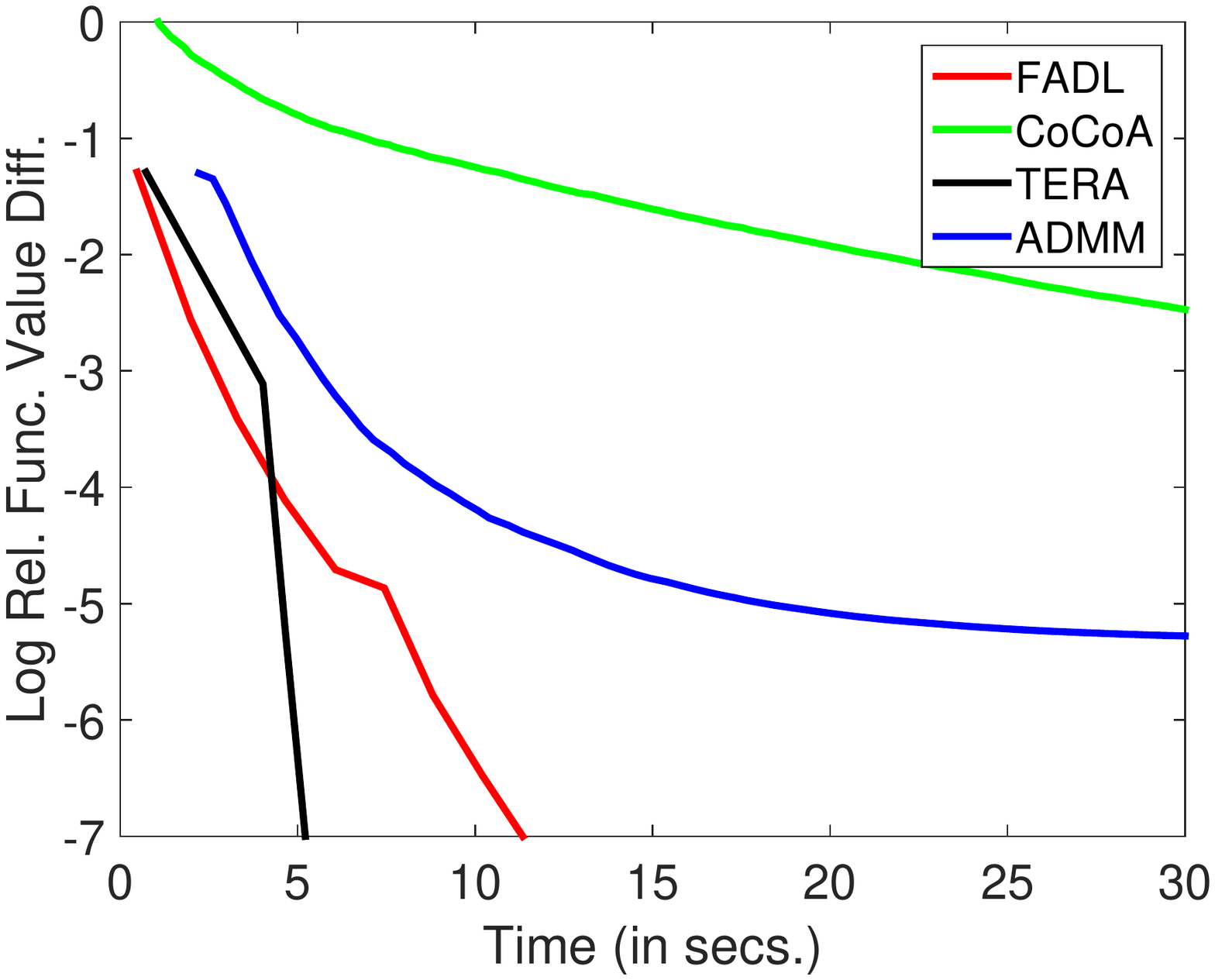}
}

\caption{Plots showing the time efficiency of various methods for the two low/medium dimensional datasets.}
\label{fig:timepass2}
\end{figure}

Having made the best choice of settings for the methods, we now evaluate FADL against TERA, ADMM and CoCoA in more detail. We do this using three sets of plots.
\begin{enumerate}
\item {\it Communication passes.} We plot the variation of the training objective function as a function of the number of communication passes. We do this only for $P=8$ and $P=128$ to give an idea of how performance varies for small and large number of nodes. For the $x$-axis we prefer the number of communication passes instead of the number of outer iterations since the latter has a different meaning for different methods while that former is quite uniform for all methods. Figures~\ref{fig:commpass1} and~\ref{fig:commpass2} give the plots respectively for the large dimensional ($m$ large) datasets ({\it kdd2010}, {\it url} and {\it webspam}) and medium/small dimensional ($m$ medium/small) datasets ({\it mnist8m} and {\it rcv}).
\item {\it Time.} We plot the variation of the training objective function as a function of the actual solution time. Figures~\ref{fig:timepass1} and~\ref{fig:timepass2} give the plots respectively for the large dimensional and medium/small dimensional datasets. We do this only for $P=8$ and $P=128$.
\item {\it Speed-up over TERA.} TERA is an established strong baseline method. So it is useful to ask how other methods fare relative to TERA and study this as a function of the number of nodes. For doing this we need to represent each method by one or two real numbers that indicate performance. Since generalization performance is finally the quantity of interest, we stop a method when it reaches within 0.1\% of the steady state AUPRC value achieved by full, perfect training of~(\ref{risk}) and record the following two measures: the total number of communication passes and the total time taken. For each measure, we plot the ratio of the measure's value for TERA to the corresponding measure's value for a method, as a function of the number of nodes, and repeat this for each method. Larger this ratio, better is a method; also, ratio greater than one means a method is faster than TERA. Figures~\ref{fig:auprcpass} and~\ref{fig:auprctime} give the plots for all the five datasets.
\end{enumerate}
Let us now use these plots to compare the methods.

\subsubsection{Rate of Convergence}
\label{subsubsec:rate}

\begin{figure}[H]
\centering
\subfigure[{\it{kdd2010}} - 8 nodes]{
\includegraphics[width=0.46\linewidth]{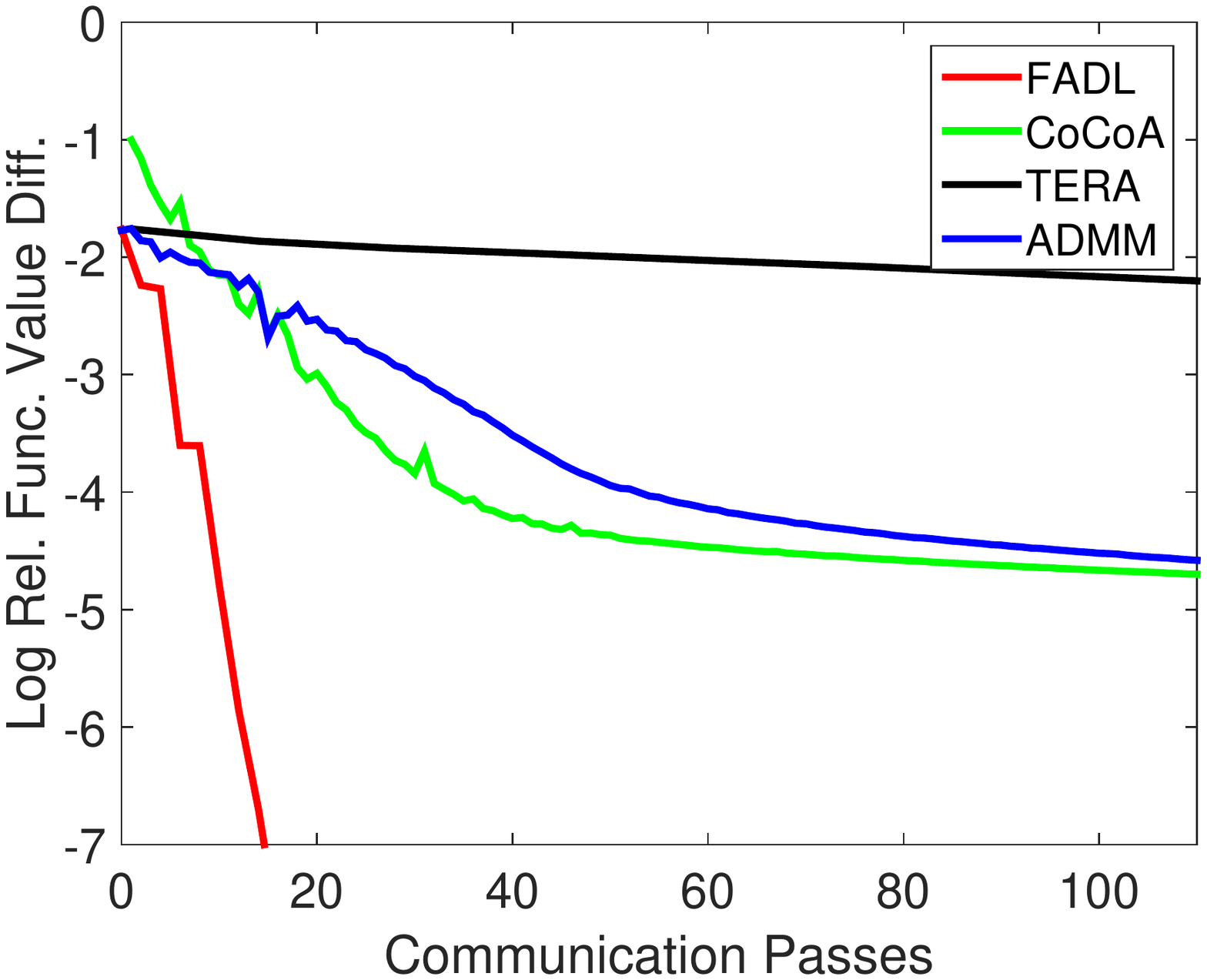}
}
\subfigure[{\it{kdd2010}} - 128 nodes]{
\includegraphics[width=0.46\linewidth]{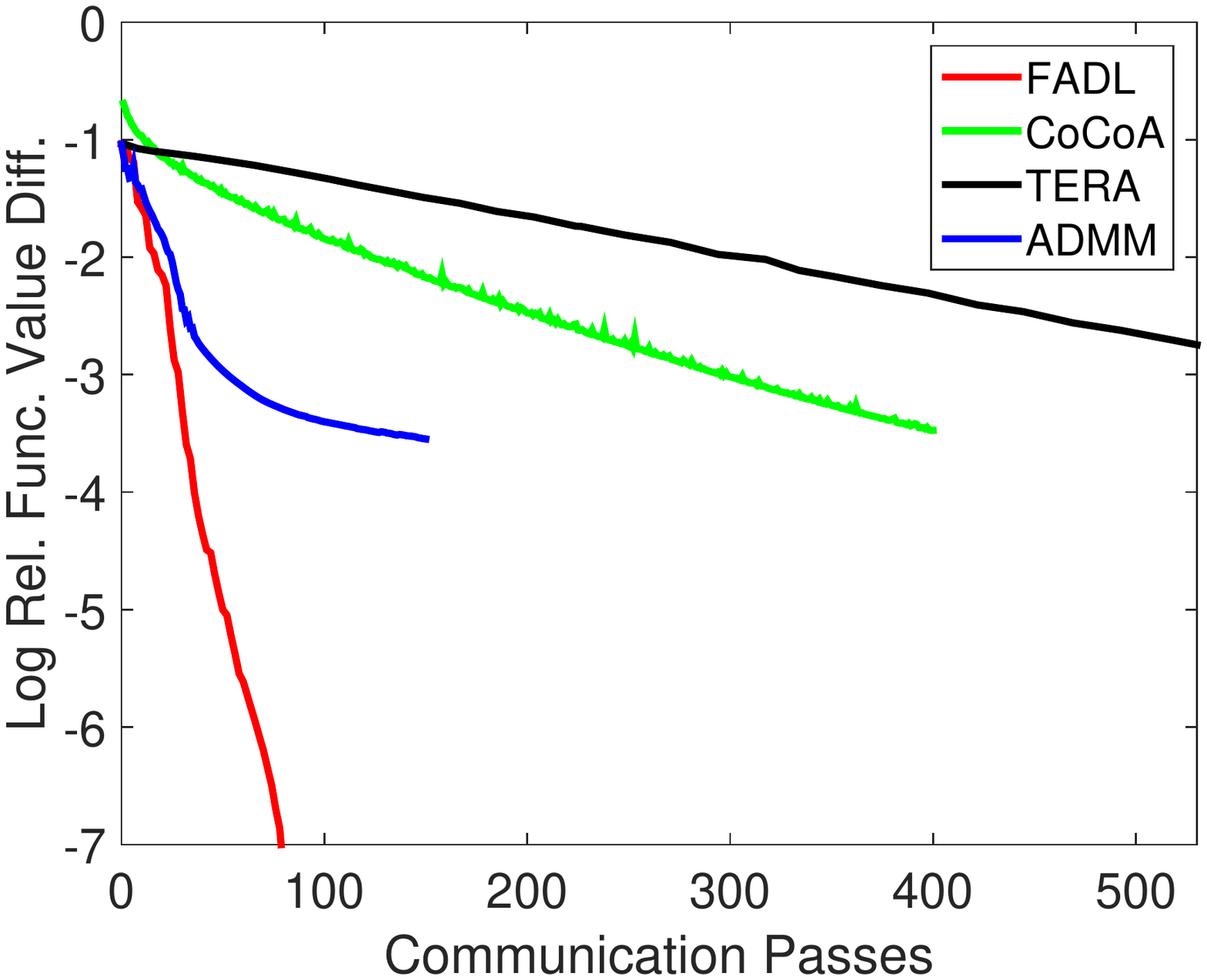}
}

\subfigure[{\it{url}} - 8 nodes]{
\includegraphics[width=0.46\linewidth]{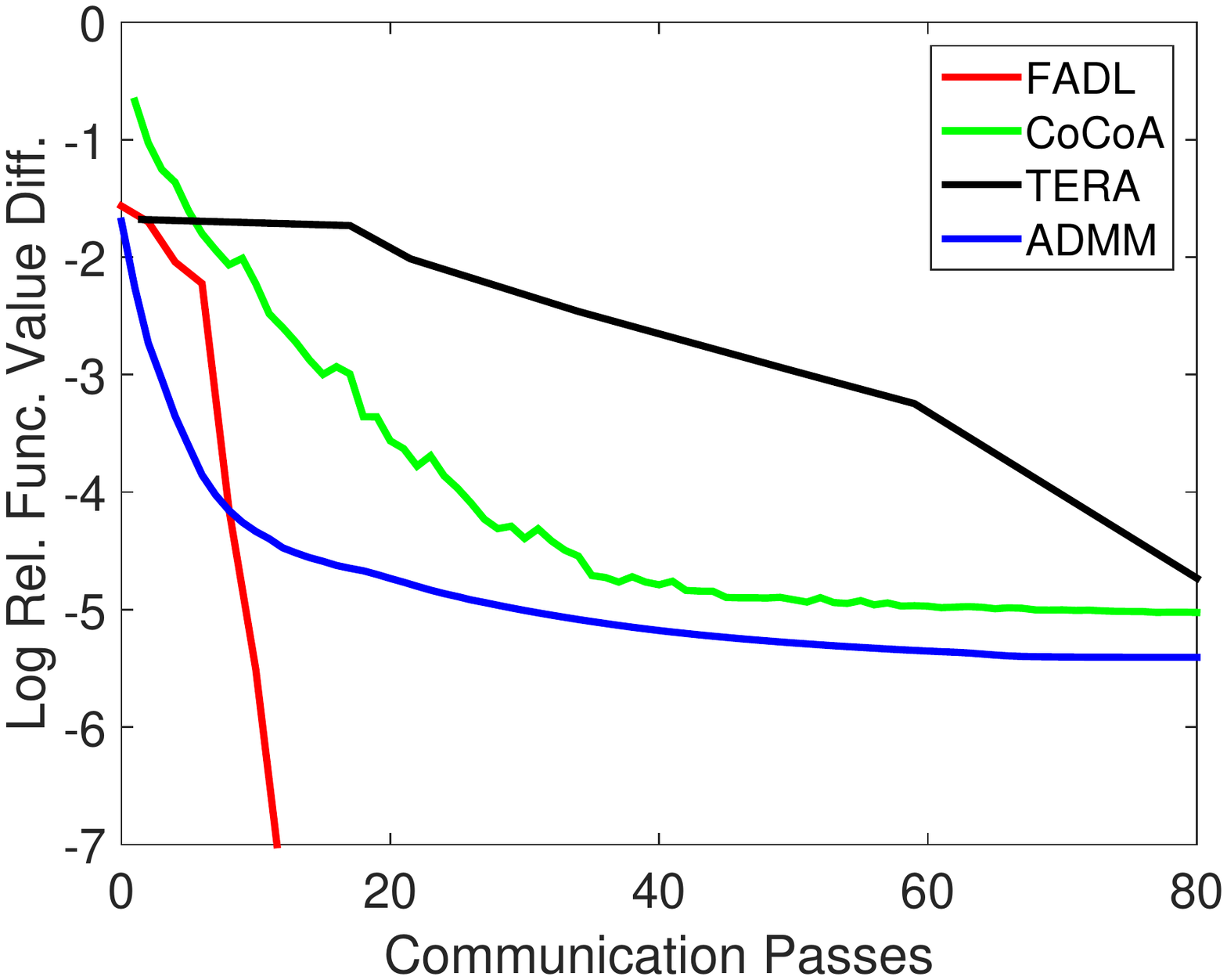}
}
\subfigure[{\it{url}} - 128 nodes]{
\includegraphics[width=0.46\linewidth]{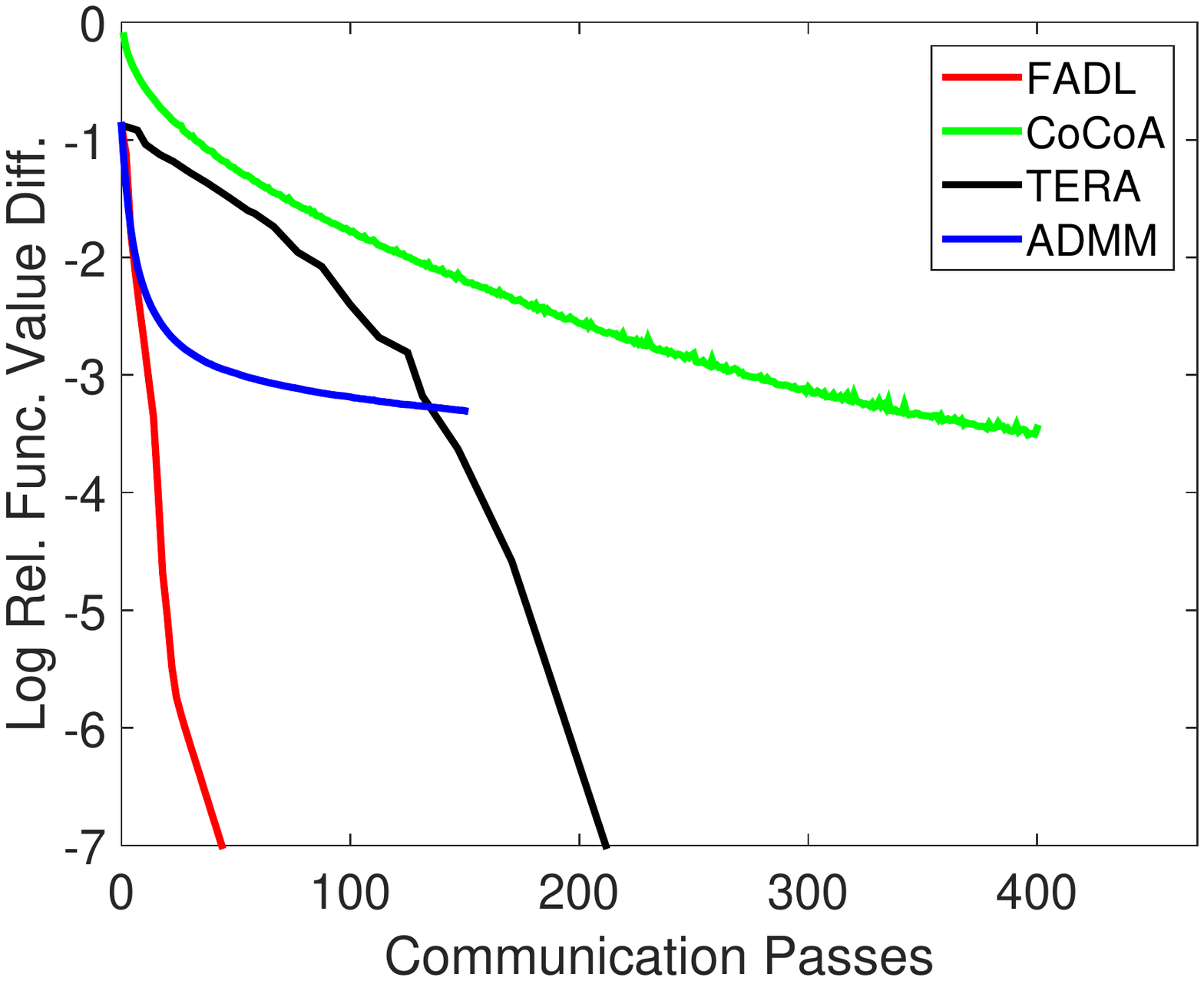}
}

\subfigure[{\it{webspam}} - 8 nodes]{
\includegraphics[width=0.46\linewidth]{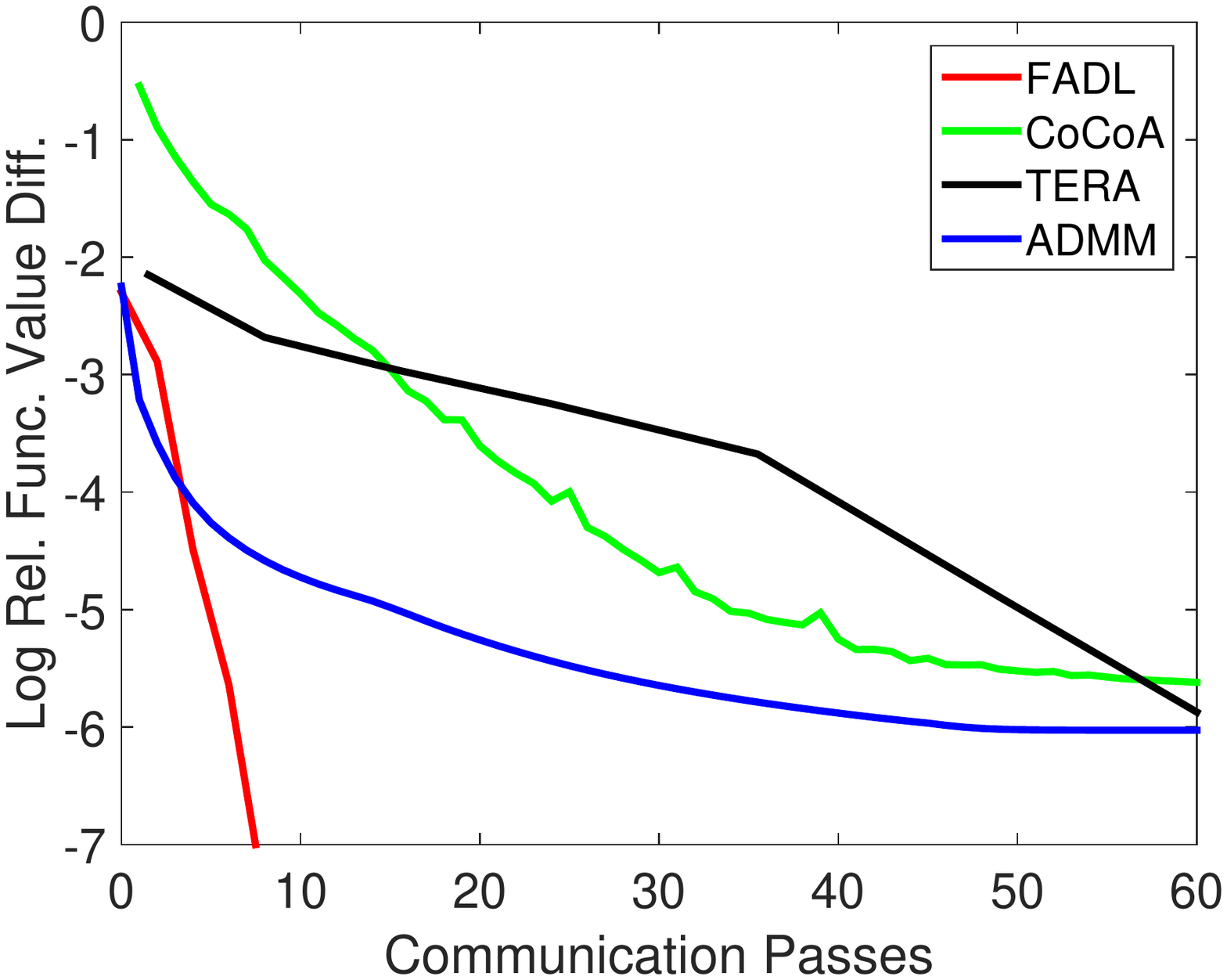}
}
\subfigure[{\it{webspam}} - 128 nodes]{
\includegraphics[width=0.46\linewidth]{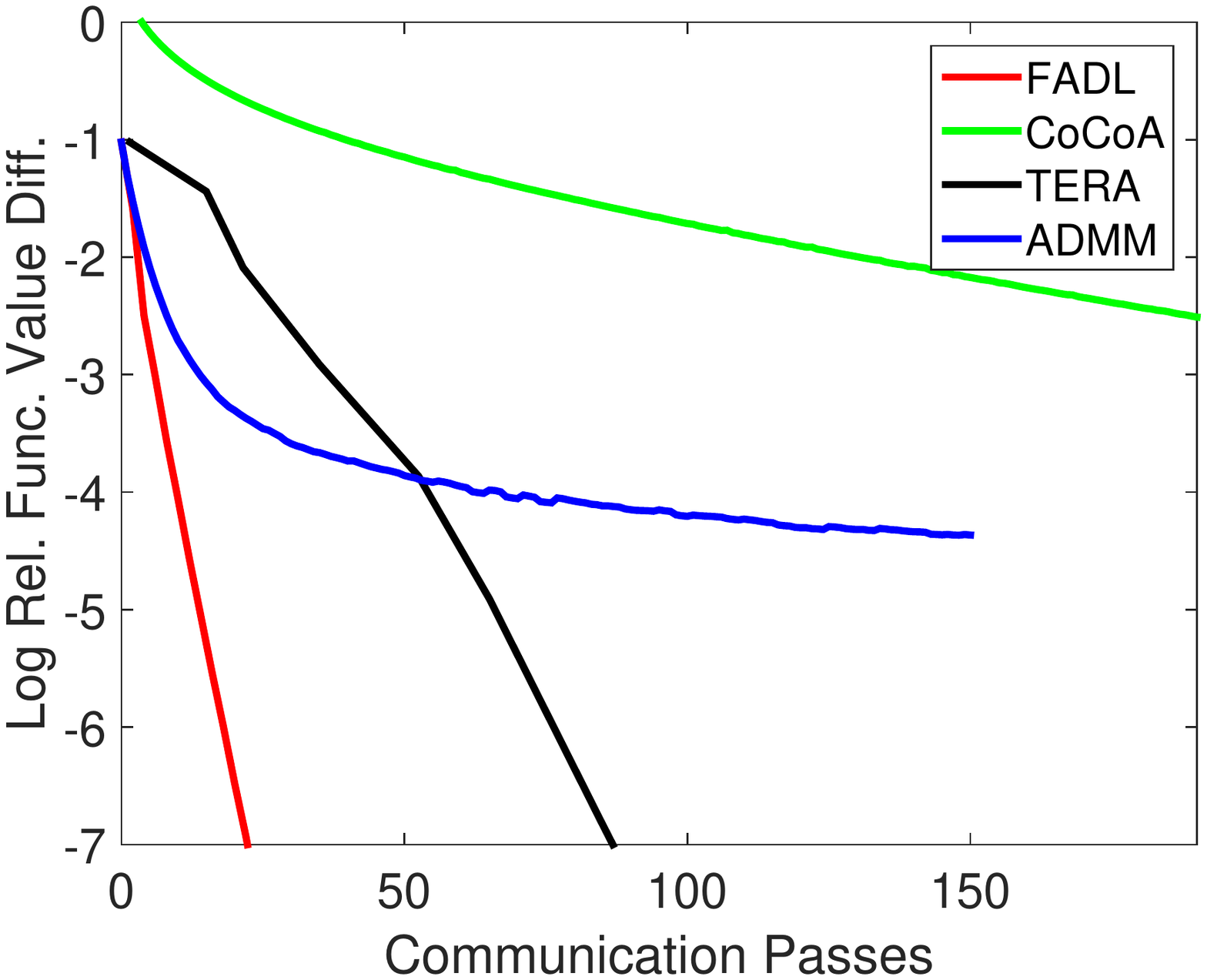}
}

\caption{Plots showing the linear convergence of various methods for the three high dimensional datasets.}
\label{fig:commpass1}
\end{figure}

\begin{figure}[H]
\centering

\subfigure[{\it{mnist8m}} - 8 nodes]{
\includegraphics[width=0.46\linewidth]{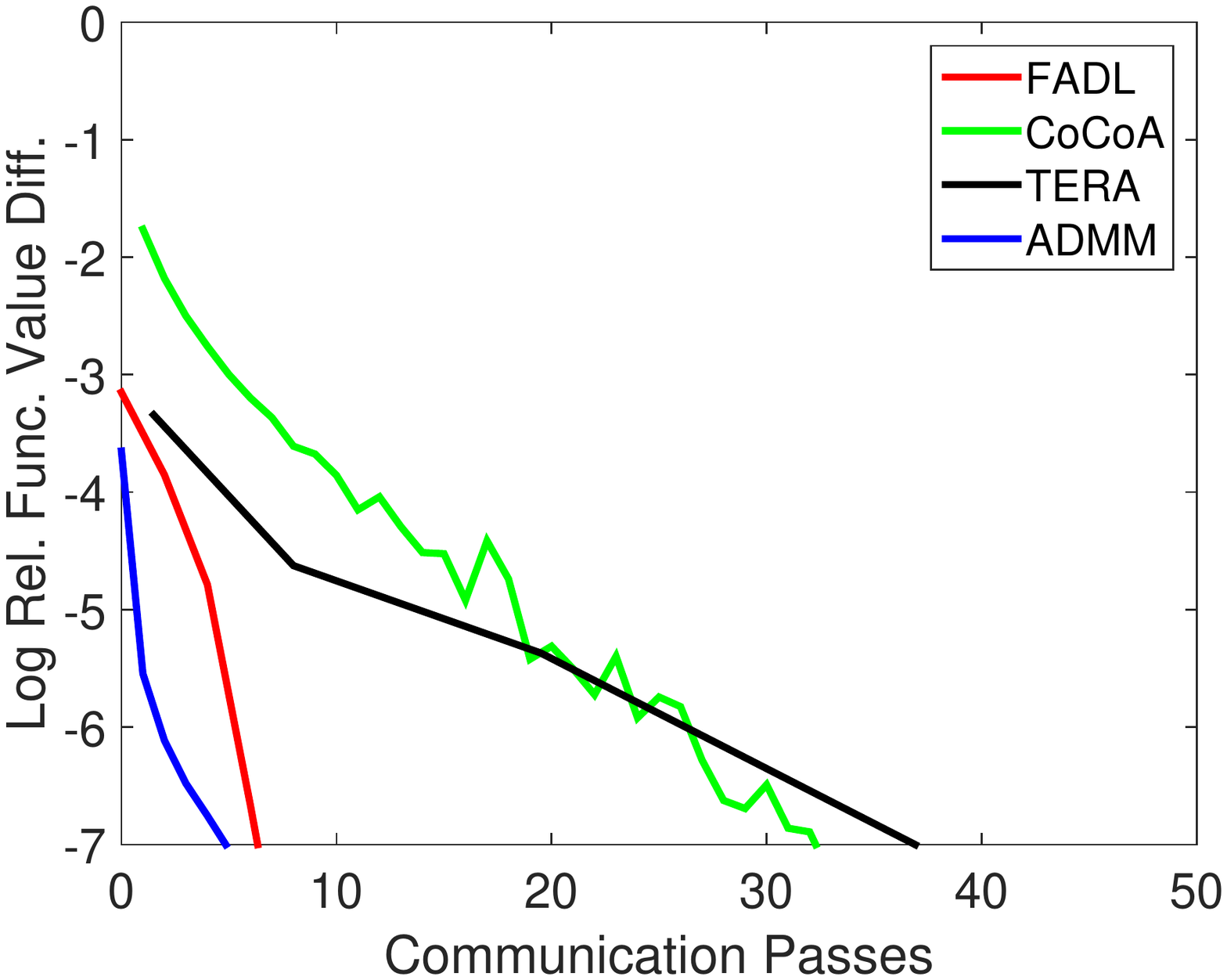}
}
\subfigure[{\it{mnist8m}} - 128 nodes]{
\includegraphics[width=0.46\linewidth]{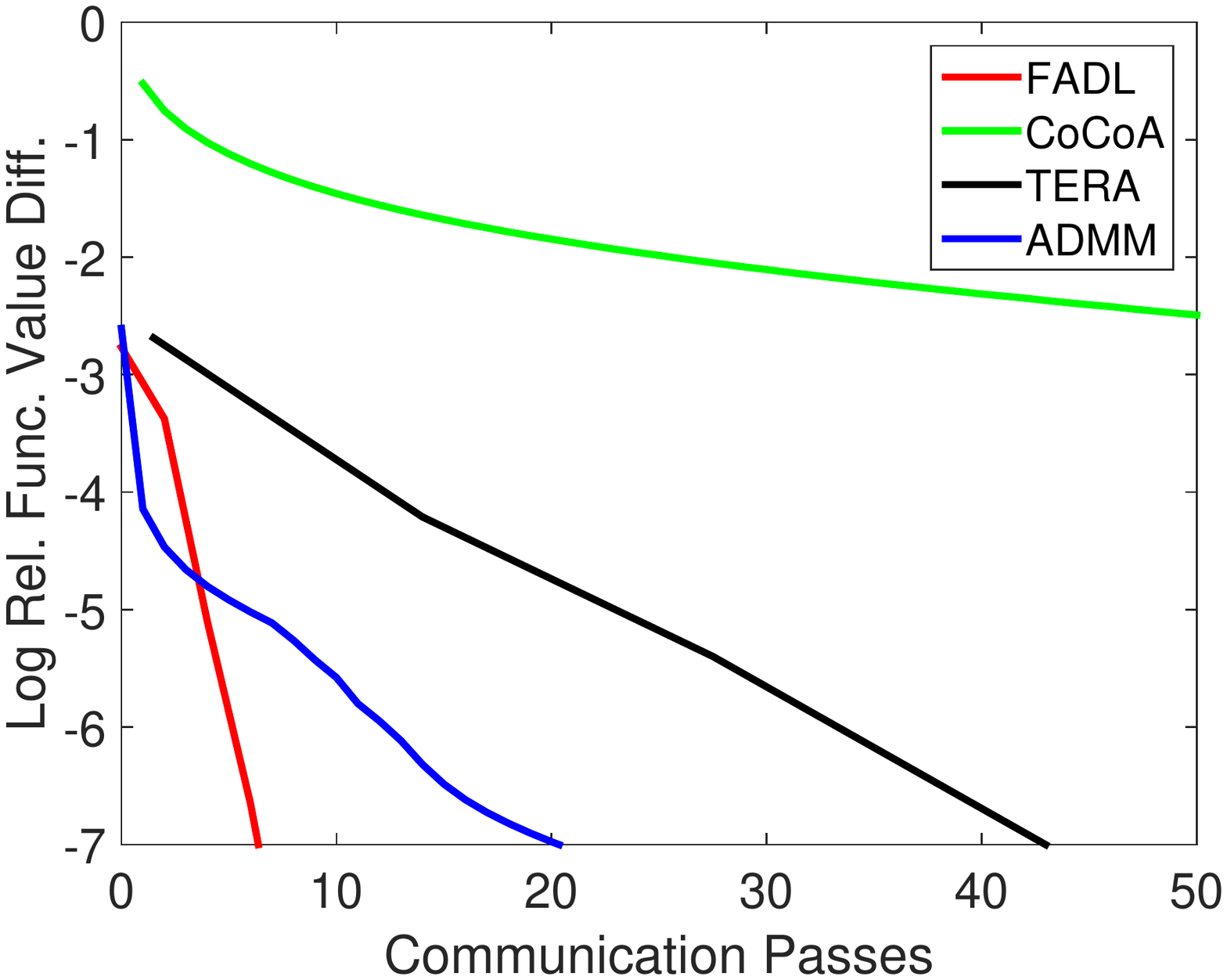}
}

\subfigure[{\it{rcv}} - 8 nodes]{
\includegraphics[width=0.46\linewidth]{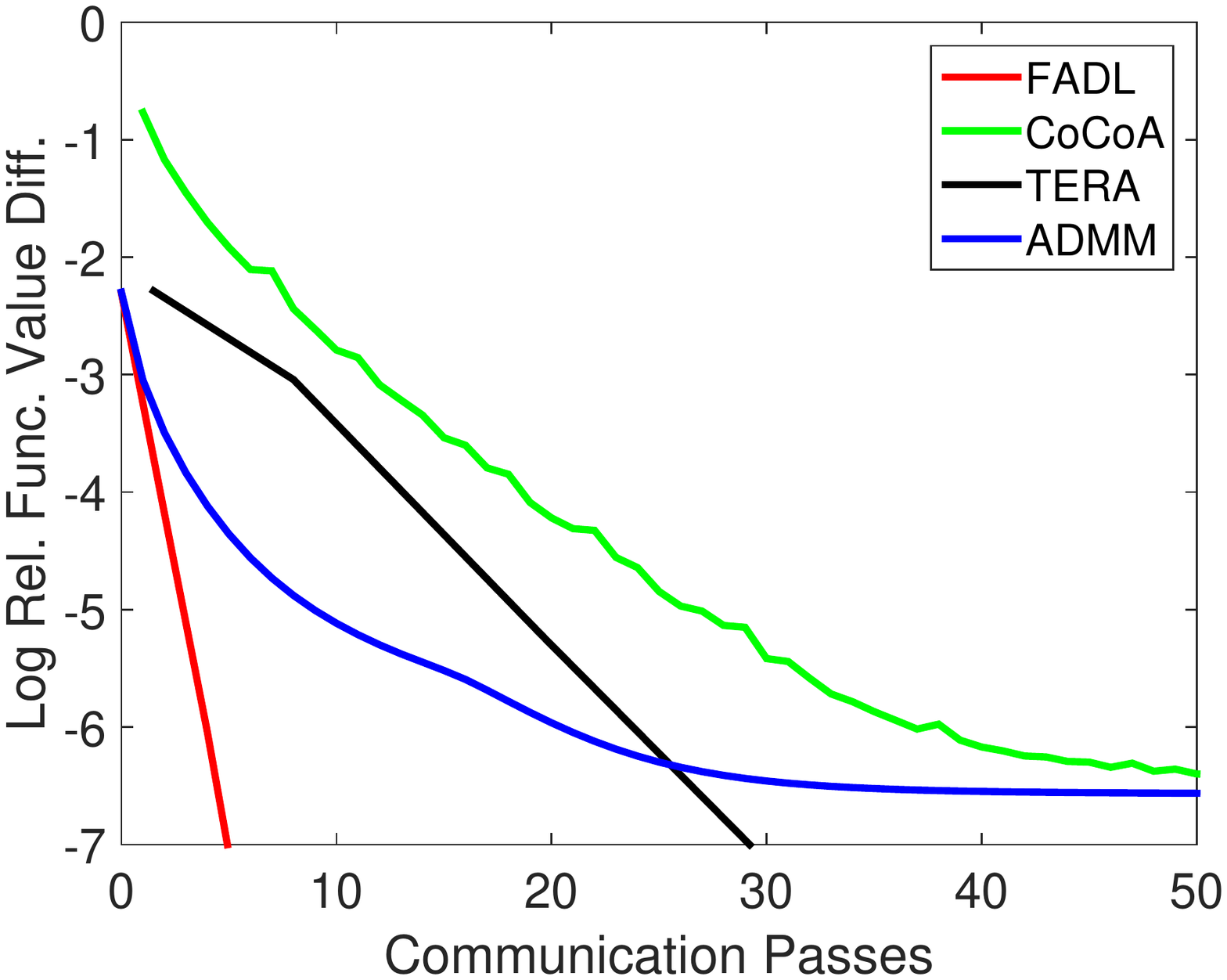}
}
\subfigure[{\it{rcv}} - 128 nodes]{
\includegraphics[width=0.46\linewidth]{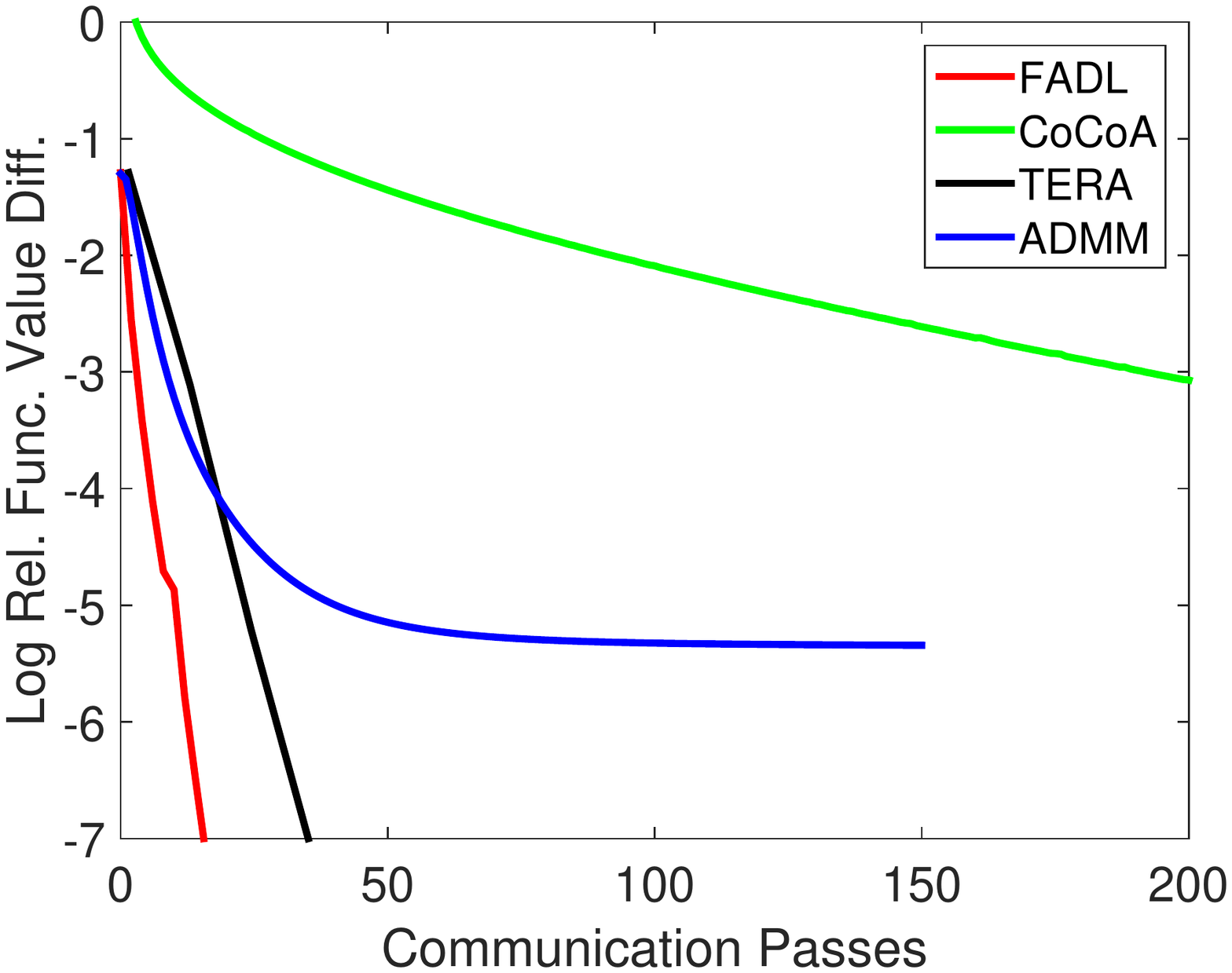}
}

\caption{Plots showing the linear convergence of various methods for the two low/medium dimensional datasets.}
\label{fig:commpass2}
\end{figure}

Analysis of the rate of convergence is better done by studying the behavior of the training objective function with respect to the number of communication passes. So it is useful to look at Figures~\ref{fig:commpass1} and~\ref{fig:commpass2}. Clearly, as predicted by theory, the rate of convergence is linear for all methods. 

TERA uses distributed computation only to compute the gradient and so the plots should be unaffected by $P$. But in the plots we do see differences between the plots for $P=8$ and $P=128$. This is because of their different initialization (average of one pass SGD solutions from nodes): the initialization with lower number of nodes is better; note also the better starting objective function value at the start (left most point) for $P=8$.

For FADL, the rate is steeper for $P=8$ than for $P=128$. This steeper behavior for lower number of nodes is expected because the functional approximation in each node becomes better as the number of nodes decreases.  

Recall from Section~\ref{intro} that, our main aim behind the design of the function approximation based methods is to reduce the number of communication passes significantly. The plots in Figures~\ref{fig:commpass1} and~\ref{fig:commpass2} clearly confirm such a reduction.

\begin{figure}[H]
\centering

\subfigure[kdd2010]{
\includegraphics[width=0.46\linewidth]{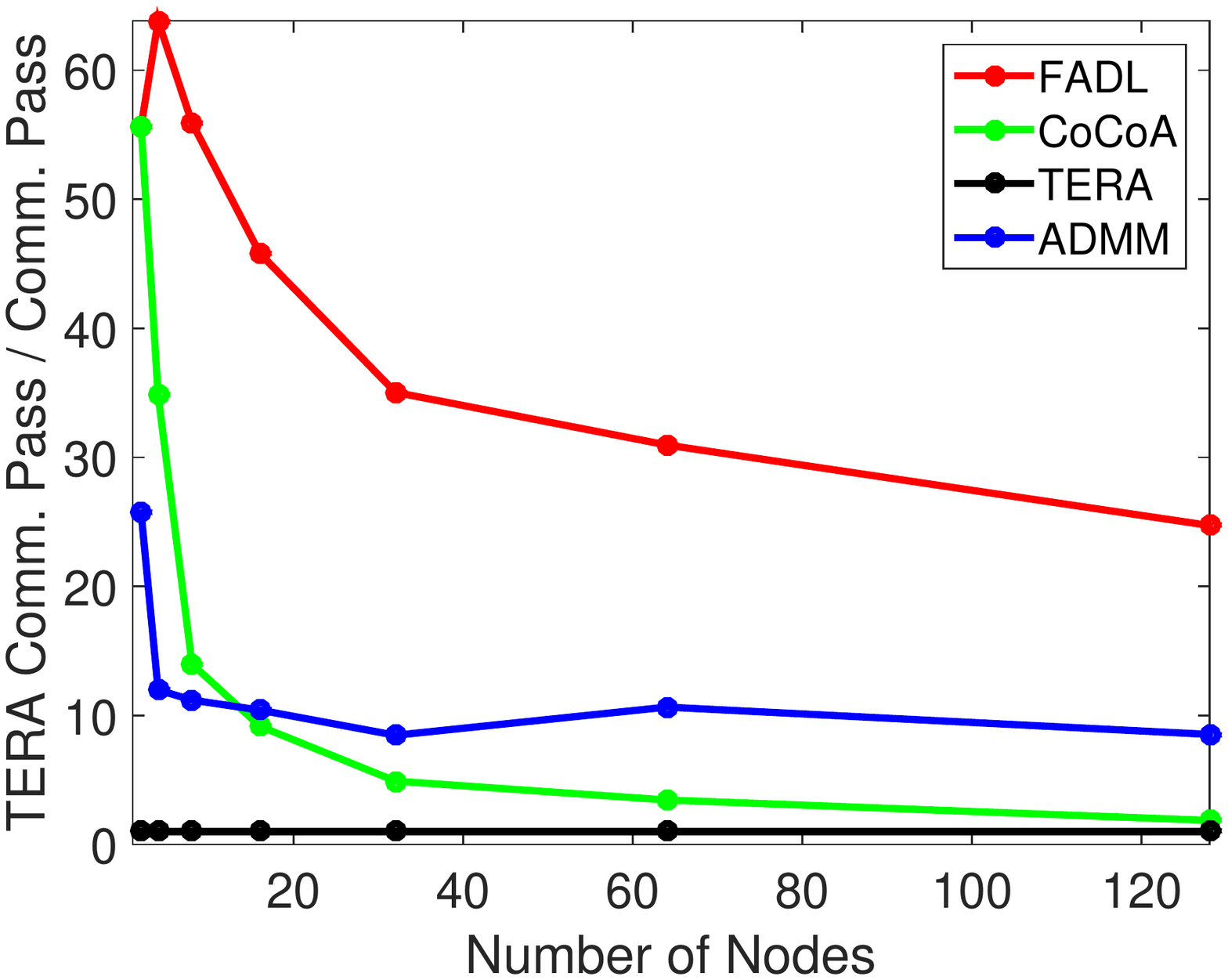}
}
\subfigure[url]{
\includegraphics[width=0.46\linewidth]{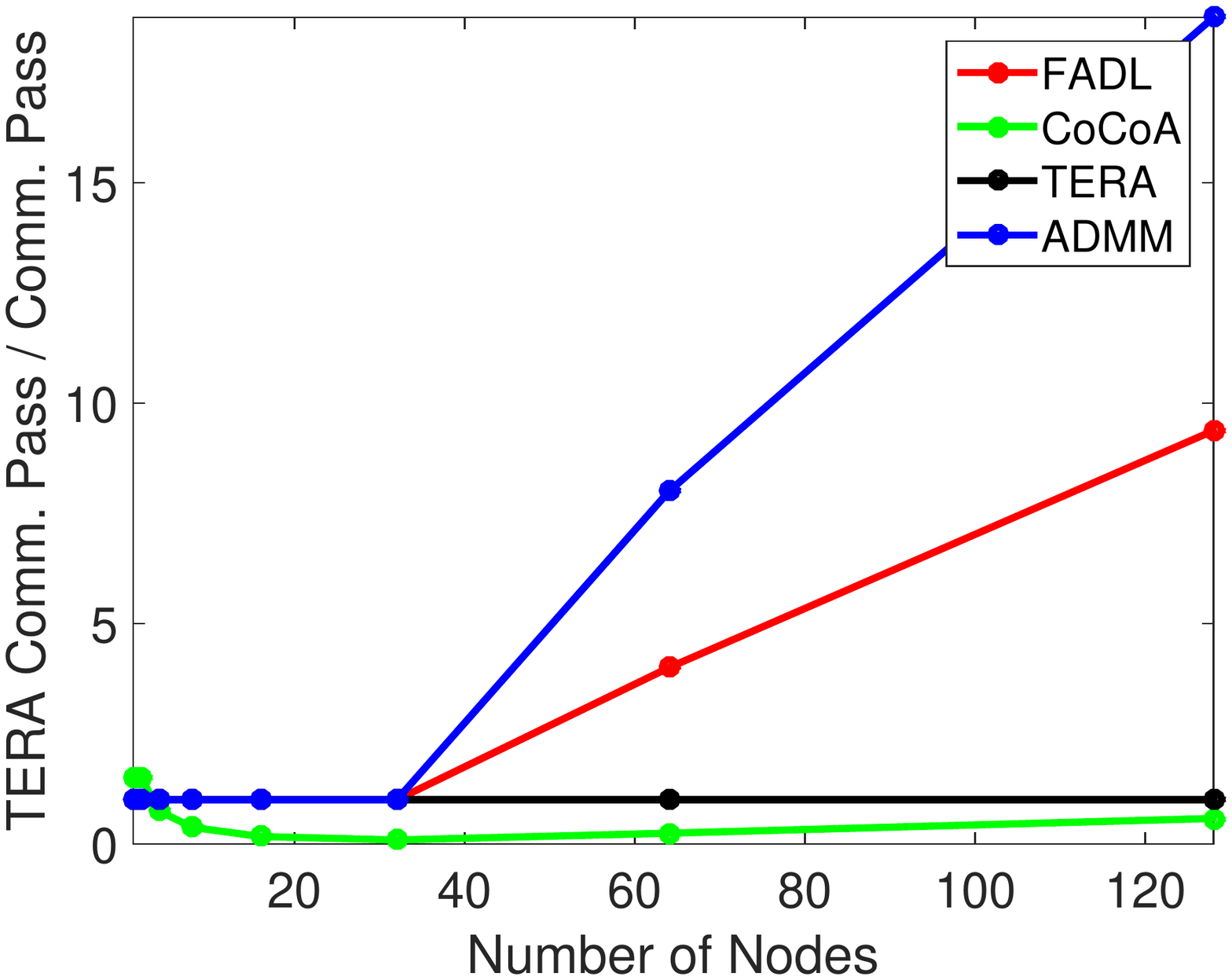}
}

\subfigure[webspam]{
\includegraphics[width=0.46\linewidth]{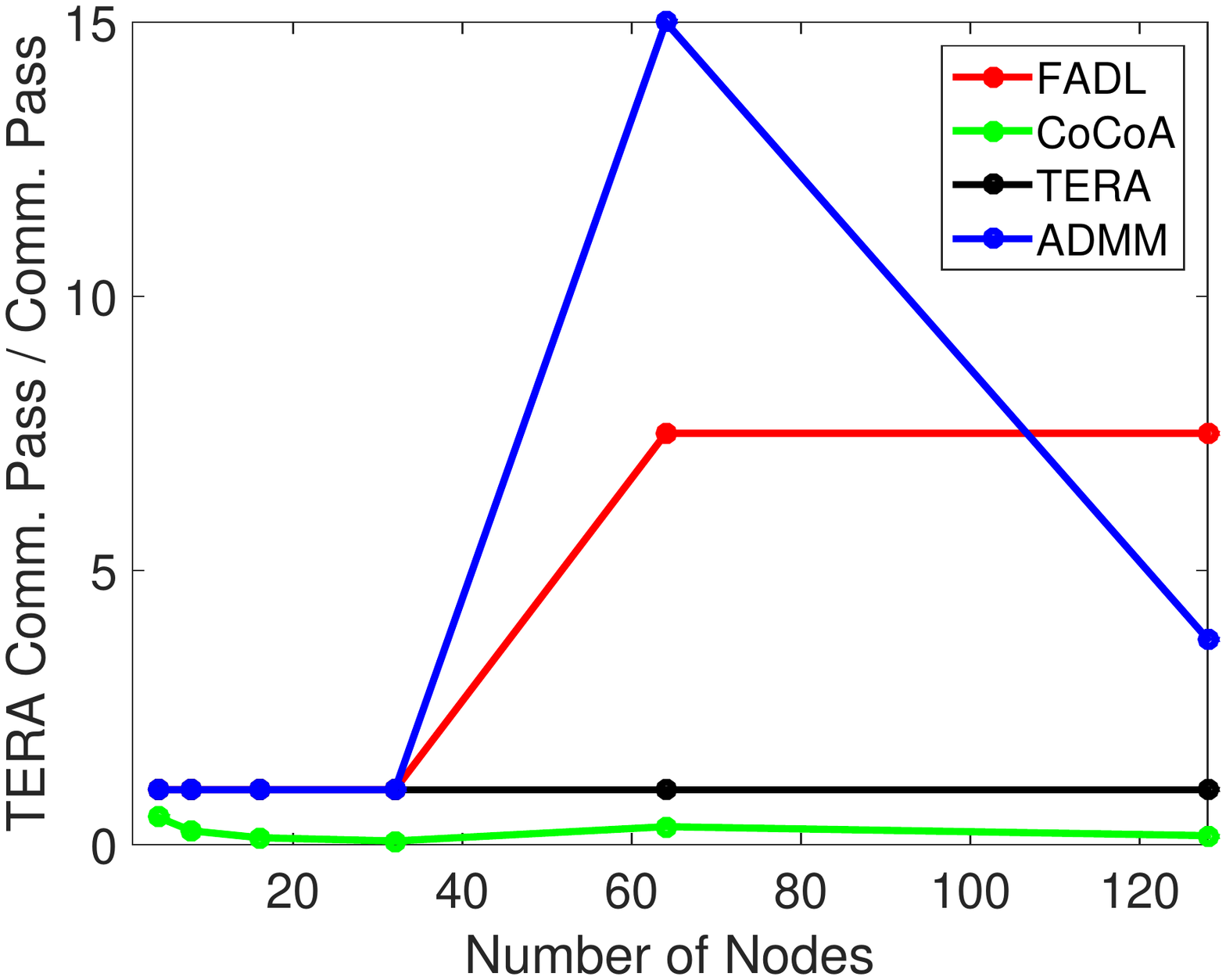}
}
\subfigure[mnist8m]{
\includegraphics[width=0.46\linewidth]{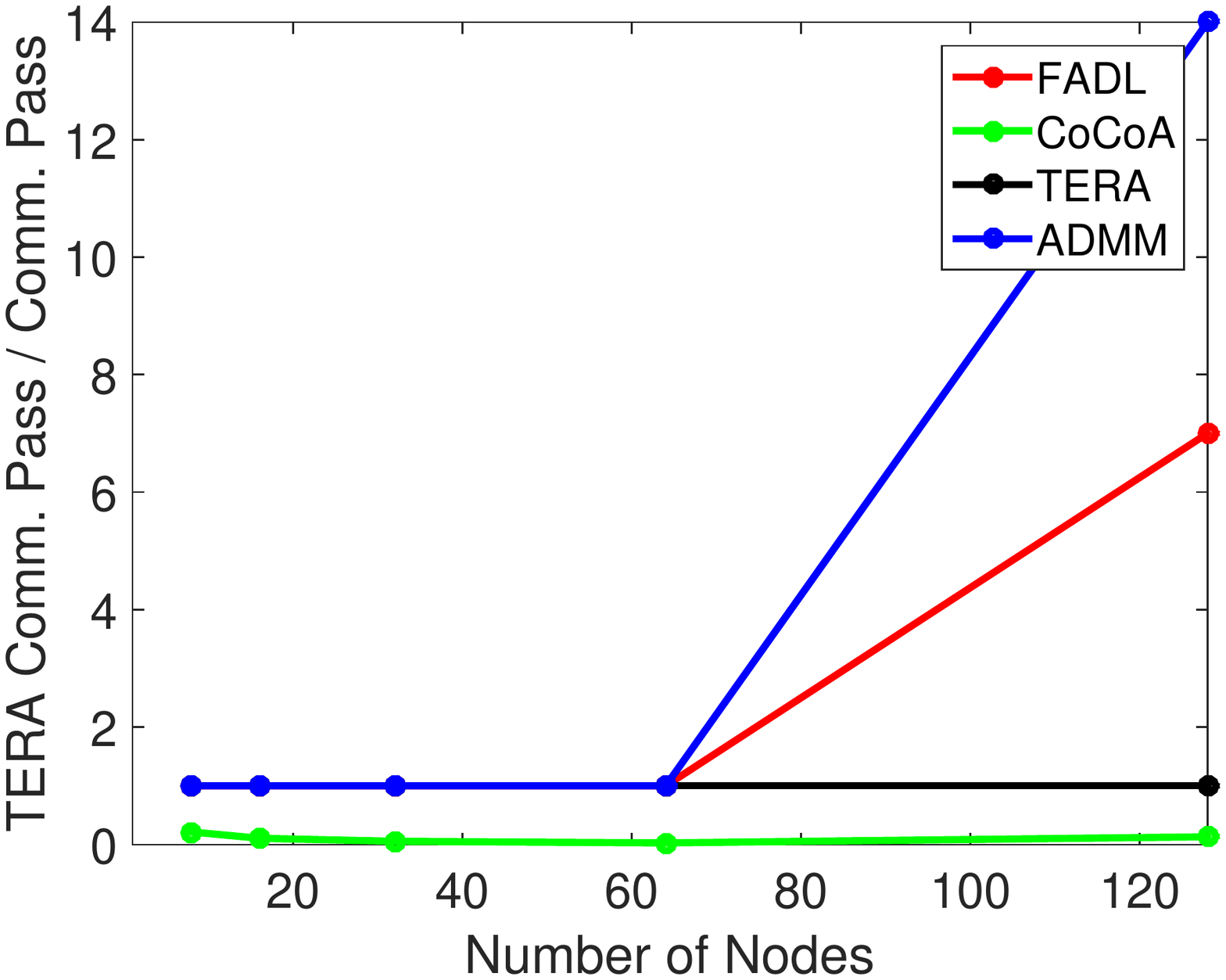}
}

\subfigure[rcv]{
\includegraphics[width=0.46\linewidth]{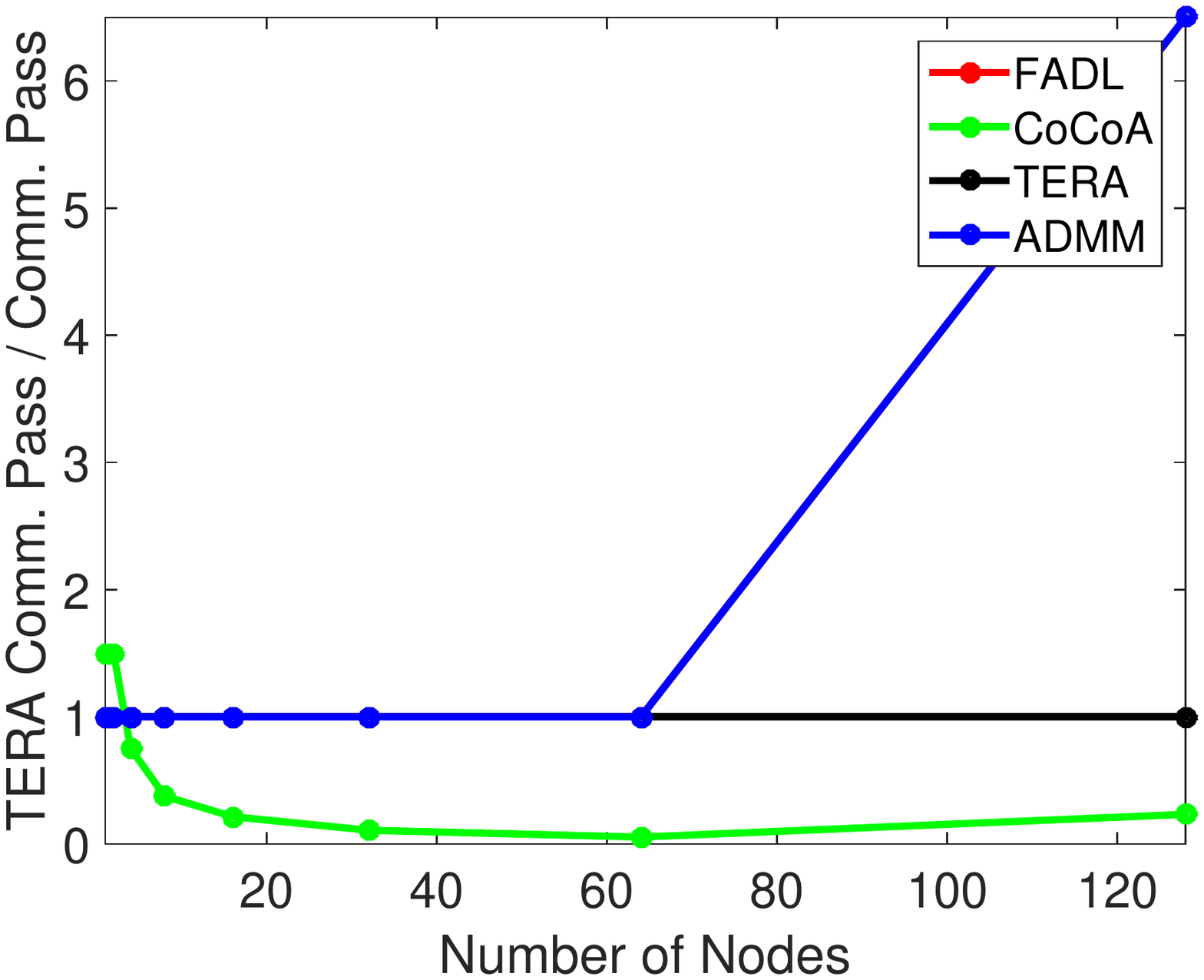}
}

\caption{Plots showing communication passes (relative to TERA) as a function of the number of nodes. Each method was terminated when it reached within 0.1\% of the steady state AUPRC value achieved by full, perfect training of~(\ref{risk}). For {\it rcv}, the FADL and ADMM curves coincide.}
\label{fig:auprcpass}
\end{figure}

Even though the end convergence rate of ADMM is slow, it generally shows good rates of convergence in the initial stages of training. This is a useful behavior because generalization measures such as AUPRC tend to achieve steady state values quickly in the early stages. This usefulness is seen in Figure~\ref{fig:auprcpass} too.

Overall, FADL gives much better rates of convergence (both, in the early training stage as well as in the end stage) compared to TERA, CoCoA and ADMM methods. FADL shows a large reduction in the number of communication passes over TERA, especially when the number of nodes is small. Against CoCoA the trend is the other way: FADL needs a much smaller number of communication passes than CoCoA especially when the number of nodes is large. These observations can also be seen from Figure~\ref{fig:auprcpass}. Clearly CoCoA seems to be very slow with increasing number of nodes.

\subsubsection{Time Taken}
\label{subsubsec:time}

In the previous analysis we ignored computation costs within each iteration. But these costs play a key role when we analyze overall efficiency in terms of the actual time taken. We study this next. Figures~\ref{fig:timepass1} and~\ref{fig:timepass2} are relevant for this study. FADL, ADMM and CoCoA involve much more extensive computations in the inner iterations than TERA; this is especially true when the number of nodes is small because of the large amount of local data in each node. Thus TERA fares much better in the time analysis than what we saw while studying using communication passes only. Compare, for example, TERA and ADMM with respect to communication passes and time. Although ADMM is much more efficient than TERA with respect to the number of communication passes, TERA catches up nicely with ADMM on the time taken. 

CoCoA does well sometimes; for example, on {\it kdd2010} and {\it url}, when the number of nodes is small, say $P=8$. But it is slow otherwise, especially when the number of nodes is large.

FADL is uniformly better than ADMM with respect to the total time taken.
Overall, FADL shows the best performance, performing equally or much better than other methods in different situations. With medium/low dimensional datasets (see Figure~\ref{fig:timepass2}), communication time is less of an issue and so the expectation is that FADL is less of value for them. Even on these datasets, FADL does equally or better than TERA.

\subsubsection{Relative performance of the methods} 
\label{subsubsec:relative}

\begin{figure}[H]
\centering

\subfigure[kdd2010]{
\includegraphics[width=0.46\linewidth]{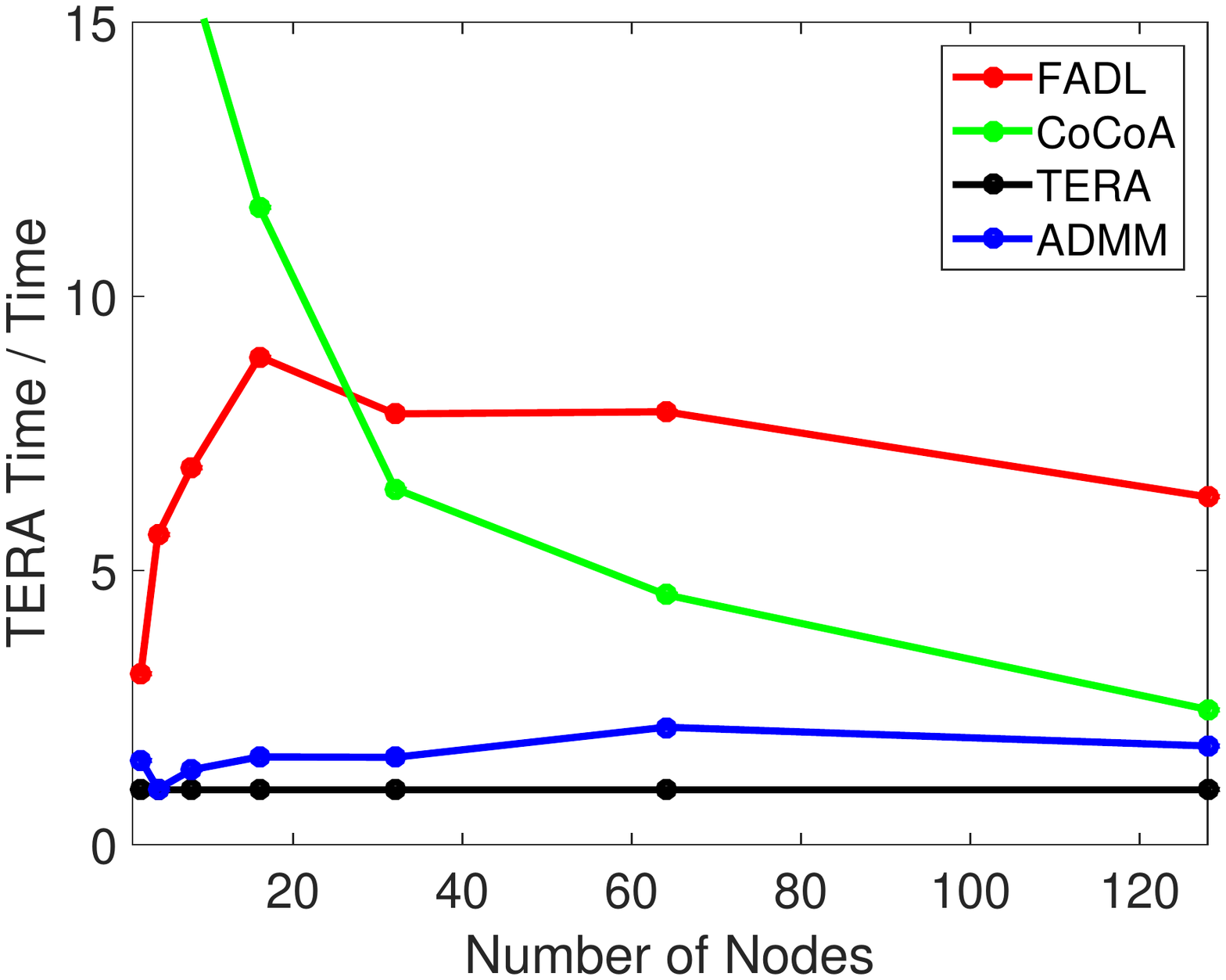}
}
\subfigure[url]{
\includegraphics[width=0.46\linewidth]{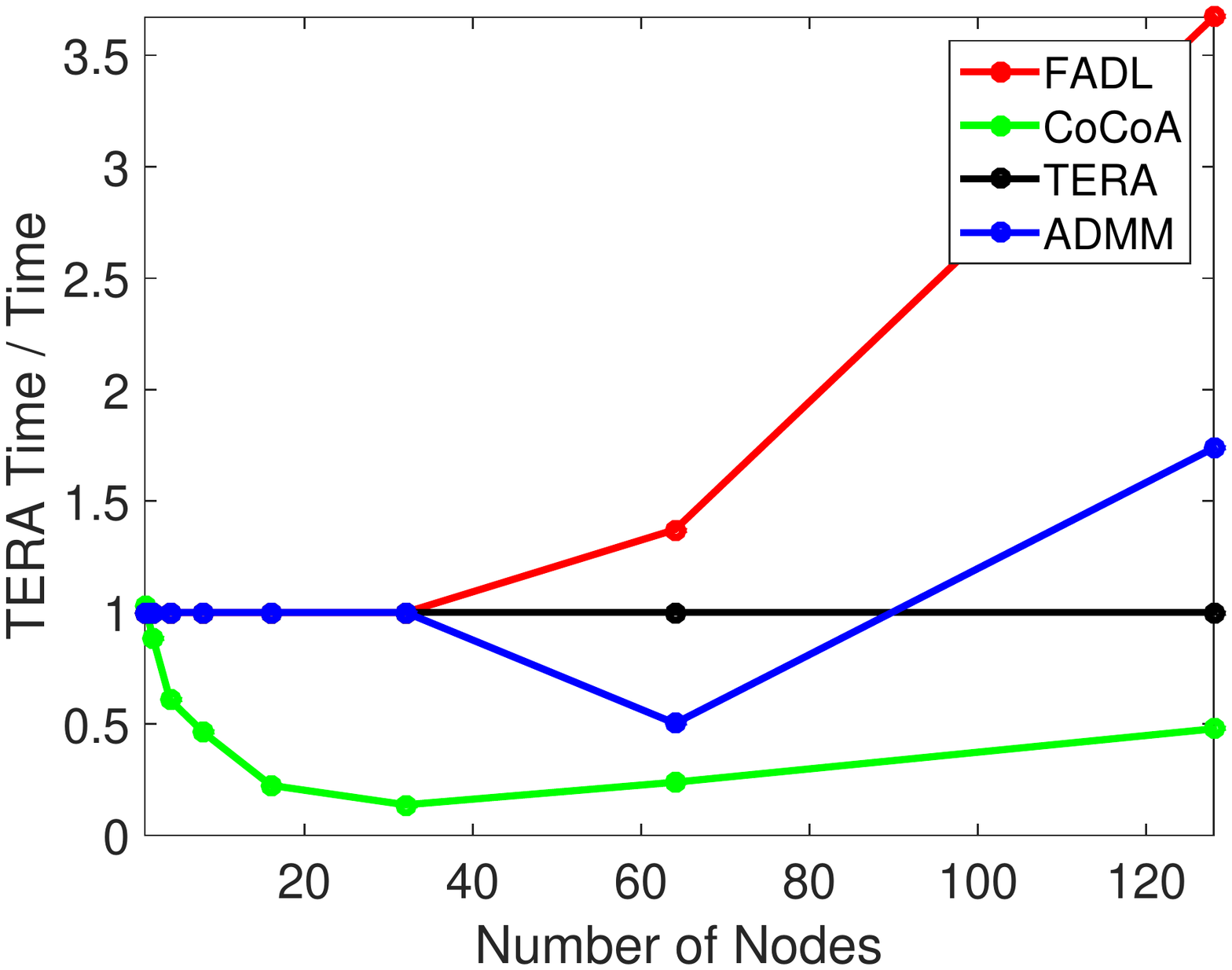}
}

\subfigure[webspam]{
\includegraphics[width=0.46\linewidth]{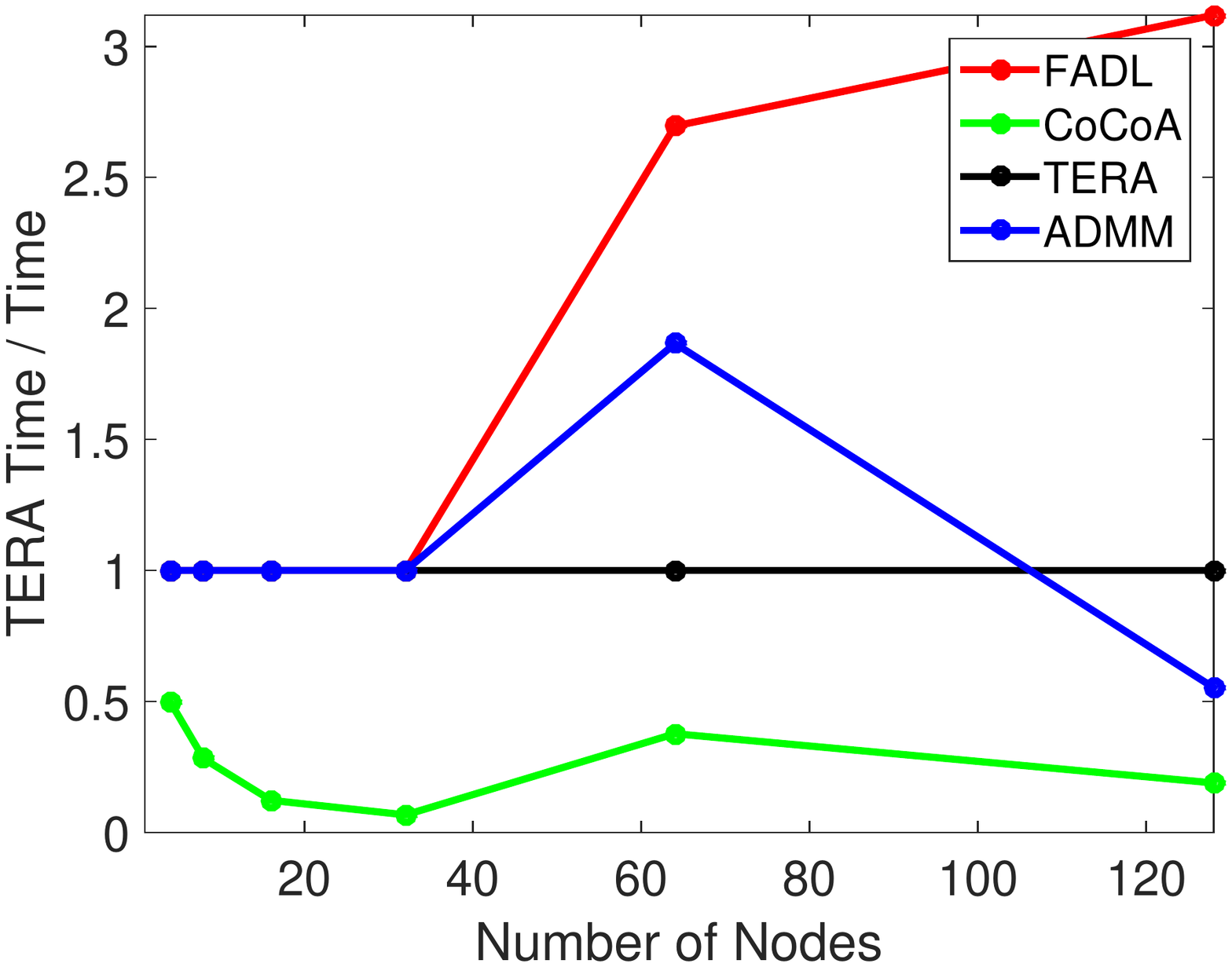}
}
\subfigure[mnist8m]{
\includegraphics[width=0.46\linewidth]{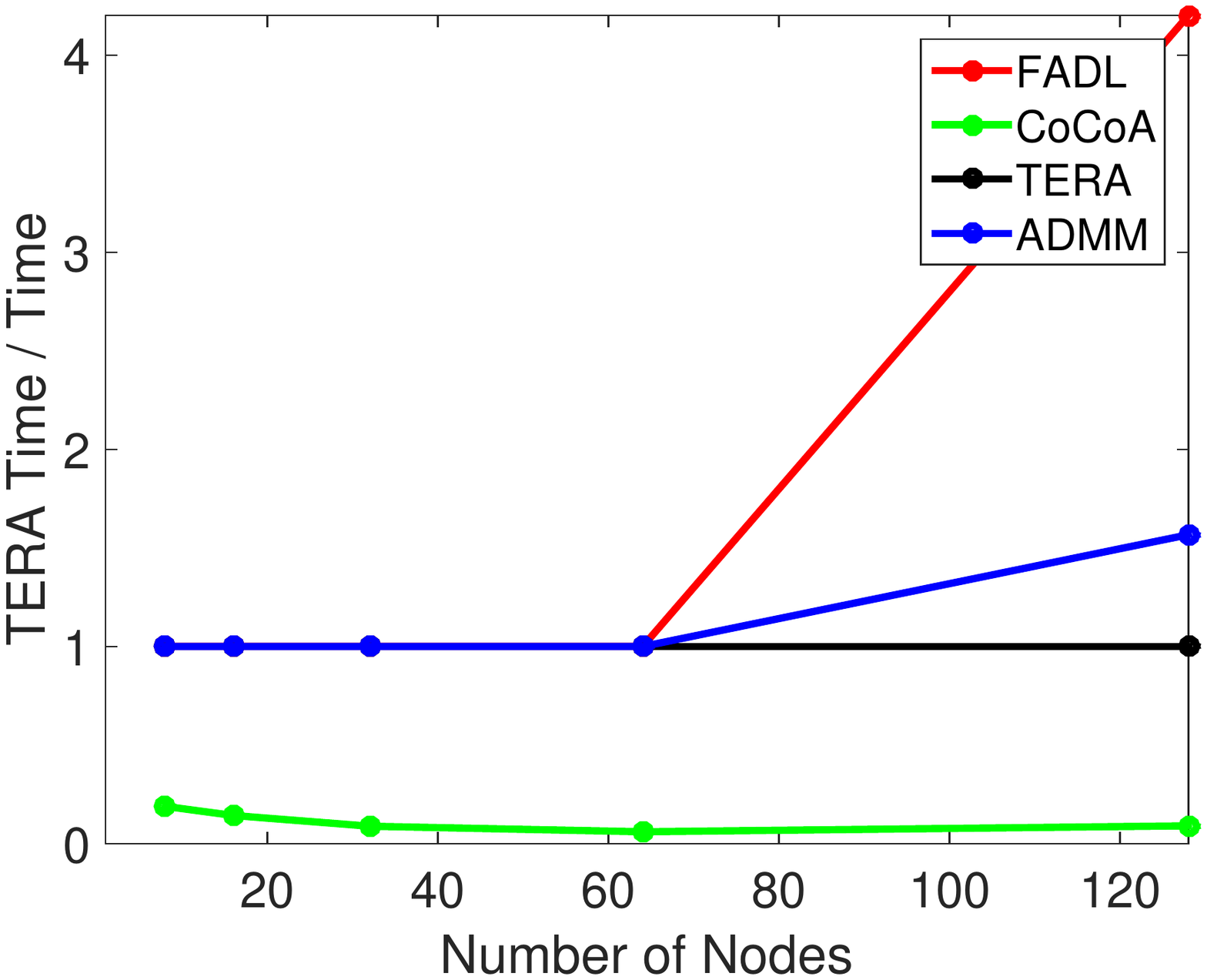}
}

\subfigure[rcv]{
\includegraphics[width=0.46\linewidth]{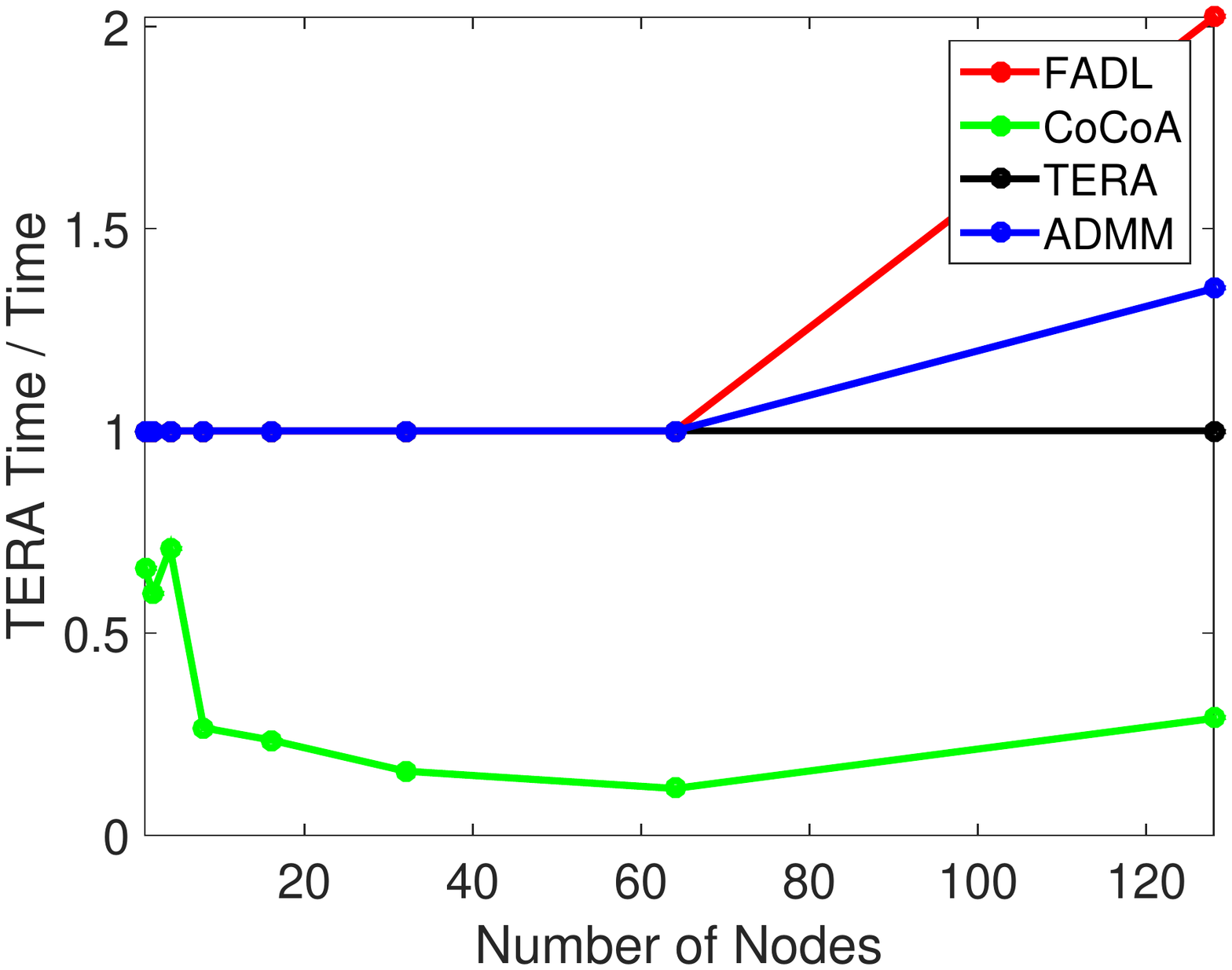}
}

\caption{Plots showing time (relative to TERA) as a function of the number of nodes. Each method was terminated when it reached within 0.1\% of the steady state AUPRC value achieved by full, perfect training of~(\ref{risk}).}
\label{fig:auprctime}
\end{figure}

Figure~\ref{fig:auprctime} is relevant for this study.
CoCoA shows impressive speed-up over TERA on {\it kdd2010} but it is much slower on all the other datasets. It is unclear why CoCoA fares so well on {\it kdd2010} but not on the other datasets. ADMM gives an overall decent performance when compared to TERA. FADL is consistently faster than TERA, with speed-ups ranging anywhere from 1-10. In communication-heavy scenarios where reducing the number of communication passes is most important, methods such as FADL and ADMM have great value (see Figure~\ref{fig:auprcpass}), with the possibility of getting even higher speed-ups over TERA. Except for {\it kdd2010} for which FADL is slower than CoCoA for small number of nodes, it is generally the fastest method.

\subsubsection{Speed-up as a function of $P$}
\label{subsubsec:speedup}

Let us revisit Figure~\ref{fig:timepass1} and look at the plots corresponding to {\it kdd2010} for FADL.\footnote{We choose FADL as an example, but the comments made in the discussion apply to other methods too.} It can be observed that the time needed for reaching a certain tolerance, say {\it Log Rel. Func. Value Diff. = -3}, is two times smaller for $P=8$ than for $P=128$. This means that using a large number of nodes is not useful, which prompts the question: {\it Is a distributed solution really necessary?} There are two answers to this question. First, as we already mentioned, when the training data is huge\footnote{The datasets, {\it kdd2010}, {\it url} and {\it webspam} are really not huge in the {\it Big data} sense. In this paper we used them only because of lack of availability of much bigger public datasets.} and {\it the data is generated and forced to reside in distributed nodes}, the right question to ask is not whether we get great speed-up, but to ask which method is the fastest. Second, for a given dataset and method, if the time taken to reach a certain approximate stopping tolerance (e.g., based on AUPRC) is plotted as a function of $P$, it usually has a minimum at a value $P>1$. Given this, it is appropriate to choose a $P$ optimally to minimize training time. A large fraction of Big data machine learning applications involve periodically repeated model training involving newly added data. For example, in Advertising, logistic regression based click probability models are retrained on a daily basis on incrementally varying datasets. In such scenarios it is worthwhile to spend time to tune $P$ in an early deployment phase to minimize time, and then use this choice of $P$ for future runs.

\subsubsection{Computation and Communication Costs}
\label{subsubsec:costs}

Table~\ref{tab:compcommcost} shows the ratio of computational cost to communication cost for the three high dimensional datasets for all the methods.\footnote{For the medium/low dimensional datasets {\it rcv} and {\it mnist8m}, communication latencies, line search cost etc. also play a key role and an analysis of computation cost versus communication cost does not provide any great insight.}
Note that the ratio is small for {\it TERA} and so communication cost dominates the time for it. On the other hand, both the costs are well balanced for FADL. Note that ratio varies in the range of $0.625-2.845$. This clearly shows that FADL trades-off computation with communication, while significantly reducing the number of communication passes (Figures~\ref{fig:commpass1} and~\ref{fig:commpass2}) and time (Figures~\ref{fig:timepass1} and~\ref{fig:timepass2}).

\subsection{Summary}
\label{subsec:summary}

It is useful to summarize the findings of the empirical study.
\begin{itemize}
\item FADL gives a great reduction in the number of communication passes, making it clearly superior to other methods in communication heavy settings.
\item In spite of higher computational costs per iteration FADL shows the overall best performance on the total time taken. This is true even for medium and low dimensional datasets.
\item FADL shows a speed-up of 1-10 over TERA, the actual speed-up depending on the dataset and the setting.
\item FADL nicely balances computation and communication costs.
\end{itemize}

%We also notice that the gap between our methods and {\it{SQM}} and {\it{HYBRID}} decreases with increasing number of nodes. This happens because our functional approximation becomes cruder as $P$ increases. The exact $P$ up to which our approach performs better varies from dataset to dataset and depends on the learning curve. For steeper learning curves, our performance should degrade slower. The exact analysis is left as a future work. Similar observations can be made on the time curves in Figure~\ref{}.

%Finally, Figure~\ref{} shows time versus AUPRC results for different methods. Our method clearly outperforms both {\it{HYBRID}} and {\it{SQM}} in getting good AUPRC values (within $0.5\%$ of the optimal).

%To conclude, our functional approximation based distributed learning algorithm is flexible and fills several gaps in the literature. We have demonstrated that our algorithms work well when (a) the number of features is very large, (b) the functional approximation is good, and (c) moderately small relative objective function error is desired. We expect to come up with better functional approximations and hybrid algorithms in the near future that does well under all conditions.

\begin{table}[ht]
\caption{Ratio of the total computation cost to the total communication cost for various methods which were terminated when AUPRC reached within 0.1\% of the AUPRC value for 128 nodes.} % title of Table
\centering % used for centering table
\begin{tabular}{|c|c|c|c|c|c} % centered columns (4 columns)
\hline %inserts double horizontal lines
& FADL & CoCoA & TERA & ADMM  \\
\hline
{\it kdd2010} & 1.6333 &   0.1416  &   0.1422  &  1.8499\\
\hline
{\it url} & 1.3650 &   0.1040  &  0.2986 &    3.4886\\
\hline	
{\it webspam} & 1.2082 &   0.1570  &  0.2423 &    1.2543\\
%\hline
%{\it mnist8m} & 0.7942 &   0.1385 &   1.5486 &    5.7881\\
%\hline	
%{\it rcv}   &  0.9693  &  0.0486  &  3.6671  &  1.6682 \\
\hline
\end{tabular}
\label{tab:compcommcost}
\end{table} 
\section{Discussion}
\label{disc}

In this section, we discuss briefly, other different distributed settings made possible by our algorithm. The aim is to show the flexibility and generality of our approach while ensuring {\it glrc}.\comment{ We will also talk briefly about the extension to non-convex setting.}

Section~\ref{distr} considered example partitioning where examples are distributed across the nodes. First, it is worth mentioning that, due to the gradient consistency condition, {\it partitioning} is not a necessary constraint; our theory allows examples to be resampled, i.e., each example is allowed to be a part of any number of nodes arbitrarily. For example, to reduce the number of outer iterations, it helps to have more examples in each node.

Second, the theory proposed in Section~\ref{distr} holds for feature partitioning also. Suppose, in each node $p$ we restrict ourselves to a subset of features, $J_p\subset \{1,\ldots,d\}$, i.e., include the constraint,
$w_p \in  \{ w: w(j)=w^r(j) \;\; \forall r\not\in J_p\}$,
where $w(j)$ denotes the weight of the $j^{th}$ feature. Note that we do not need $\{J_p\}$ to form a partition. This is useful since important features can be included in all the nodes.

\noindent {\bf Gradient sub-consistency.} Given $w^r$ and $J_p$ we say that $\fhat_p(w)$ has gradient sub-consistency with $f$ at $w^r$ on $J_p$ if $\frac{\partial f}{\partial w(j)} (w^r) = \frac{\partial \fhat}{\partial w(j)} (w^r) \;\; \forall \; j\in J_p$.

Under the above condition, we can modify the algorithm proposed in Section~\ref{distr} to come up with a feature decomposition algorithm with {\it glrc}.

Several feature decomposition based approaches~\citep{richtarik2013,patriksson1998fp} have been proposed in the literature. The one closest to our method is the work by Patrikkson on a synchronized parallel algorithm~\citep{patriksson1998fp} which extends a generic cost approximation algorithm~\citep{patriksson1998ca} that is similar to our functional approximation. The sub-problems on the partitions are solved in parallel. Although the objective function is not assumed to be convex, the cost approximation is required to satisfy a monotone property, implying that the approximation is convex. The algorithm only has asymptotic linear rate of convergence and it requires the feature partitions to be disjoint. In contrast, our method has {\it glrc} and works even if features overlap in partitions. Moreover, there does not exist any counterpart of our example partitioning based distributed algorithm discussed in Section~\ref{distr}.

Recently~\citet{mairal2013} has developed an algorithm called MISO. The main idea of MISO (which is in the spirit of the EM algorithm) is to build majorization approximations with good properties so that line search can be avoided, which is interesting. MISO is a serial method. Developing a distributed version of MISO is an interesting future direction; but, given that line search is inexpensive communication-wise, it is unclear if such a method would give great benefits.

Our approach can be easily generalized to joint example-feature partitioning as well as non-convex settings.\footnote{For non-convex settings {\it glrc} is hard to establish, but proving a simpler convergence theory is quite possible.} The exact details of all the extensions mentioned above and related experiments are left for future work.

\section{Conclusion}
\label{conc}

To conclude, we have proposed FADL, a novel functional approximation based distributed algorithm with provable global linear rate of convergence. The algorithm is general and flexible in the
sense of allowing different local approximations at the node level, different algorithms for optimizing the local approximation, early stopping and general data usage in the nodes. We also established the superior efficiency of FADL by evaluating it against key existing distributed methods. We believe that FADL has great potential for solving machine learning problems arising in Big data.

%\bibliographystyle{plain}
\bibliography{jmlrex}  % sigproc.bib is the name of the Bibliography in this case

\begin{thebibliography}{32}
\providecommand{\natexlab}[1]{#1}
\providecommand{\url}[1]{\texttt{#1}}
\expandafter\ifx\csname urlstyle\endcsname\relax
  \providecommand{\doi}[1]{doi: #1}\else
  \providecommand{\doi}{doi: \begingroup \urlstyle{rm}\Url}\fi

\bibitem[Agarwal et~al.(2011)Agarwal, Chapelle, Dudik, and
  Langford]{agarwal2011}
A.~Agarwal, O.~Chapelle, M.~Dudik, and J.~Langford.
\newblock A reliable effective terascale linear learning system.
\newblock In \emph{arXiv:1140.4198}, 2011.

\bibitem[Bertsekas and Tsitsiklis(1997)]{bertsekas1997}
D.~P. Bertsekas and J.~N. Tsitsiklis.
\newblock \emph{Parallel and distributed computation: Numerical methods}.
\newblock Athena Scientific, Cambridge, MA, 1997.

\bibitem[Bottou(2010)]{bottou2010}
L.~Bottou.
\newblock Large-scale machine learning with stochastic gradient descent.
\newblock In \emph{COMPSTAT'2010}, pages 177--187, 2010.

\bibitem[Boyd and Vandenberghe(2004)]{boyd2004}
S.~Boyd and L.~Vandenberghe.
\newblock \emph{Convex optimization}.
\newblock Cambridge University Press, Cambridge, UK, 2004.

\bibitem[Boyd et~al.(2011)Boyd, Parikh, Chu, Peleato, and Eckstein]{Boyd2011}
S.~Boyd, N.~Parikh, E.~Chu, B.~Peleato, and J.~Eckstein.
\newblock Distributed optimization and statistical learning via the alternating
  direction method of multipliers.
\newblock \emph{Foundations and Trends in Machine Learning}, pages 1--122,
  2011.

\bibitem[Byrd et~al.(2012)Byrd, Chin, Neveitt, and Nocedal]{byrd2012}
R.~H. Byrd, G.~M. Chin, W.~Neveitt, and J.~Nocedal.
\newblock On the use of stochastic {H}essian information in optimization
  methods for machine learning.
\newblock \emph{SIAM Journal of Optimization}, pages 977--995, 2012.

\bibitem[Chang et~al.(2008)Chang, Hsieh, and Lin]{chang2008}
K.W. Chang, C.J. Hsieh, and C.J. Lin.
\newblock Coordinate descent method for large-scale l2-loss linear {SVM}.
\newblock \emph{JMLR}, pages 1369--1398, 2008.

\bibitem[Chu et~al.(2006)Chu, Kim, Lin, Yu, Bradski, Ng, and Olukotun]{chu2006}
C.T. Chu, S.K. Kim, Y.A. Lin, Y.Y. Yu, G.~Bradski, A.Y. Ng, and K.~Olukotun.
\newblock Map-reduce for machine learning on multicore.
\newblock \emph{NIPS}, pages 281--288, 2006.

\bibitem[Deng and Yin(2012)]{deng2012}
W.~Deng and W.~Yin.
\newblock On the global linear convergence of the generalized alternating
  direction method of multipliers.
\newblock \emph{Rice University CAAM Technical Report}, TR12-14, 2012.

\bibitem[Hall et~al.(2010)Hall, Gilpin, and Mann]{hall2010}
K.B. Hall, S.~Gilpin, and G.~Mann.
\newblock Mapreduce/bigtable for distributed optimization.
\newblock In \emph{NIPS Workshop on Leaning on Cores, Clusters, and Clouds},
  2010.

\bibitem[Hsieh et~al.(2008)Hsieh, Chang, Lin, Keerthi, and
  Sundararajan]{hsieh2008}
C.J. Hsieh, K.W. Chang, C.J. Lin, S.S. Keerthi, and S.~Sundararajan.
\newblock A dual coordinate descent method for large-scale linear {SVM}.
\newblock In \emph{ICML}, pages 408--415, 2008.

\bibitem[Jaggi et~al.(2014)Jaggi, Smith, Tak{\'a}{\v c}, Terhorst, Krishnan,
  Hofmann, and Jordan]{jaggi2014}
M.~Jaggi, V.~Smith, M.~Tak{\'a}{\v c}, J.~Terhorst, S.~Krishnan, T.~Hofmann,
  and M.I. Jordan.
\newblock Communication-efficient distributed dual coordinate ascent.
\newblock \emph{arXiv:1409.1458}, 2014.

\bibitem[Johnson and Zhang(2013)]{johnson2013}
R.~Johnson and T.~Zhang.
\newblock Accelerating stochastic gradient descent using predictive variance
  reduction.
\newblock \emph{NIPS}, 2013.

\bibitem[Lin et~al.(2008)Lin, Weng, and Keerthi]{lin2008}
C.J. Lin, R.C. Weng, and S.S. Keerthi.
\newblock Trust region newton method for large-scale logistic regression.
\newblock \emph{JMLR}, pages 627--650, 2008.

\bibitem[Mahajan et~al.(2013{\natexlab{a}})Mahajan, Keerthi, Sundararajan, and
  Bottou]{dhruv2013}
D.~Mahajan, S.~S. Keerthi, S.~Sundararajan, and L.~Bottou.
\newblock A functional approximation based distributed learning algorithm.
\newblock \emph{arXiv:1310.8418}, 2013{\natexlab{a}}.

\bibitem[Mahajan et~al.(2013{\natexlab{b}})Mahajan, Keerthi, Sundararajan, and
  Bottou]{mahajan2013}
D.~Mahajan, S.~S. Keerthi, S.~Sundararajan, and L.~Bottou.
\newblock A parallel {SGD} method with strong convergence.
\newblock \emph{NIPS Workshop on Optimization in Machine Learning},
  2013{\natexlab{b}}.

\bibitem[Mairal(2013)]{mairal2013}
J.~Mairal.
\newblock Optimization with first order surrogate functions.
\newblock \emph{ICML}, 2013.

\bibitem[Mann et~al.(2009)Mann, McDonald, Mohri, Silberman, and
  Walker]{mann2009}
G.~Mann, R.T. McDonald, M.~Mohri, N.~Silberman, and D.~Walker.
\newblock Efficient large-scale distributed training of conditional maximum
  entropy models.
\newblock In \emph{NIPS}, pages 1231--1239, 2009.

\bibitem[McDonald et~al.(2010)McDonald, Hall, and Mann]{mcdonald2010}
R.T. McDonald, K.~Hall, and G.~Mann.
\newblock Distributed training strategies for the structured perceptron.
\newblock In \emph{HLT-NAACL}, pages 456--464, 2010.

\bibitem[Patriksson(1998{\natexlab{a}})]{patriksson1998ca}
M.~Patriksson.
\newblock Cost approximation: A unified framework of descent algorithms for
  nonlinear programs.
\newblock \emph{SIAM J. on Optimization}, 8:\penalty0 561--582,
  1998{\natexlab{a}}.

\bibitem[Patriksson(1998{\natexlab{b}})]{patriksson1998fp}
M.~Patriksson.
\newblock Decomposition methods for differentiable optimization problems over
  cartesian product sets.
\newblock \emph{Comput. Optim. Appl.}, 9:\penalty0 5--42, 1998{\natexlab{b}}.

\bibitem[Pechyony et~al.(2011)Pechyony, Shen, and Jones]{pechyony2011}
D.~Pechyony, L.~Shen, and R.~Jones.
\newblock Solving large scale linear {SVM} with distributed block minimization.
\newblock \emph{NIPS workshop on Big Learning}, 2011.

\bibitem[Richt{\'a}rik and Tak{\'a}c(2012)]{richtarik2013}
P.~Richt{\'a}rik and M.~Tak{\'a}c.
\newblock Parallel coordinate descent methods for big data optimization.
\newblock \emph{CoRR}, abs/1212.0873, 2012.

\bibitem[Sharir et~al.(2014)Sharir, Srebro, and Zhang]{sharir2014}
O.~Sharir, N.~Srebro, and T.~Zhang.
\newblock Communication efficient distributed optimization using an approximate
  newton-type method.
\newblock \emph{arXiv:1312.7853v4}, 2014.

\bibitem[Smola and Vishwanathan(2008)]{smola2008}
A.~Smola and S.V.N. Vishwanathan.
\newblock \emph{Introduction to Machine Learning}.
\newblock Cambridge University Press, Cambridge, UK, 2008.

\bibitem[Wang and Lin(2013)]{wang2013}
P.W. Wang and C.J. Lin.
\newblock Iteration complexity of feasible descent methods for convex
  optimization.
\newblock \emph{Technical Report, National Taiwan University}, 2013.

\bibitem[Wolfe(1969)]{wolfe1969}
P.~Wolfe.
\newblock Convergence conditions for ascent methods.
\newblock \emph{SIAM Review}, 11:\penalty0 226--235, 1969.

\bibitem[Wolfe(1971)]{wolfe1971}
P.~Wolfe.
\newblock Convergence conditions for ascent methods: {II}: {S}ome corrections.
\newblock \emph{SIAM Review}, 13:\penalty0 185--188, 1971.

\bibitem[Yang(2013)]{yang2013a}
T.~Yang.
\newblock Trading computation for communication: distributed stochastic dual
  coordinate ascent.
\newblock \emph{NIPS}, 2013.

\bibitem[Yang et~al.(2013)Yang, Zhu, Jin, and Lin]{yang2013b}
T.~Yang, S.~Zhu, R.~Jin, and Y.~Lin.
\newblock Analysis of distributed stochastic dual coordinate ascent.
\newblock \emph{arXiv:1312.1031}, 2013.

\bibitem[Zhang et~al.(2012)Zhang, Lee, and Shin]{zhang2012}
C.~Zhang, H.~Lee, and K.G. Shin.
\newblock Efficient distributed linear classification algorithms via the
  alternating direction method of multipliers.
\newblock \emph{CIKM}, 2012.

\bibitem[Zinkevich et~al.(2010)Zinkevich, Weimer, Smola, and Li]{zinkevich2010}
M.~Zinkevich, M.~Weimer, A.~Smola, and L.~Li.
\newblock Parallelized stochastic gradient descent.
\newblock In \emph{NIPS}, pages 2595--2603, 2010.

\end{thebibliography}
% ACM needs 'a single self-contained file'!
%\balancecolumns

\appendix

\def\C{C}
\def\Ccomp{C^{\rm Comp}}
\def\Ccomm{C^{\rm Comm}}
\def\Csqm{C_{\rm SQM}}
\def\Cfadl{C_{\rm FADL}}
\def\Ccompsqm{C^{\rm Comp}_{\rm SQM}}
\def\Ccommsqm{C^{\rm Comm}_{\rm SQM}}
\def\Ccompfadl{C^{\rm Comp}_{\rm FADL}}
\def\Ccommfadl{C^{\rm Comm}_{\rm FADL}}
\def\Tout{T^{\rm outer}}
\def\Tin{T^{\rm inner}}
\def\Toutsqm{T^{\rm outer}_{\rm SQM}}
\def\Toutour{T^{\rm outer}_{\rm FADL}}

\section*{Appendix A: Complexity analysis}
\label{proofs}
Let us use the notations of section~\ref{distr} given around (\ref{comm}). We define the overall cost of any distributed algorithm as
\begin{eqnarray}
[(c_1\frac{nz}{P} + c_2m)\Tin + c_3 \gamma m]\Tout,
\label{costanal}
\end{eqnarray}
where $\Tout$ is the number of outer iterations, $\Tin$ is the number of inner iterations at each node before communication happens and  $c_1$ and $c_2$ denote the number of passes over the data and $m$-dimensional dot products per inner iteration respectively. For communication, we assume an {\it AllReduce} binary tree as described in~\citet{agarwal2011} with pipelining. As a result, we do not have a multiplicative factor of $log_2 P$ in our cost\footnote{Actually, there is another communication term, $\gamma b\; log_2 P$, where $b$ is the size of first block of communicated doubles in the pipeline. However, typically $b<<m$ and hence we ignore it.}. $\gamma$ is the relative computation to communication speed in the given distributed system; more precisely, it is the ratio of the times associated with communicating a floating point number and performing one floating point operation; $\gamma$ is usually much larger than $1$.
$c_3$ is the number of $m$-dimensional vectors (gradients, Hessian-vector computations etc.) we need to communicate.

\begin{table}[ht]
\caption{Value of cost parameters} % title of Table
\centering % used for centering table
\begin{tabular}{c c c c c c c} % centered columns (4 columns)
\hline\hline %inserts double horizontal lines
Method & $c_1$ & $c_2$ & $c_3$ & $\Tin$ \\ %[0.5ex] % inserts table
%heading
\hline
SQM & $2$ & $\approx 5-10$ & $1$ & $1$ \\
FADL & $2$ & $\approx 5-7$ & $2$ & $\hat{k}$ \\
\hline
\end{tabular}
\label{tab:costparams}
\end{table}

The values of different parameters for SQM and FADL are given in Table~\ref{tab:costparams}. $\Toutsqm$ is the number of overall conjugate gradient iterations plus gradient computations. $\hat{k}$ is the average number of conjugate gradient iterations (for the inner minimization of $\fhat_p$ using TRON) required per outer iteration in FADL. Typically $\hat{k}$ is between $5$ and $20$. 

Since dense dot products are extremely fast $c_2 m$ is small compared to $c_1 nz/P$ for both the approaches, we ignore it from (\ref{costanal}) for simplicity. Now for FADL to have lesser cost than TERA, we can use (\ref{costanal}) to get the condition,
%we need $\Cfadl \leq \Csqm$. Plugging in the parameter values and rearranging,
\begin{equation}
2.0(\hat{k}\Toutour - \Toutsqm)\frac{nz}{P}  \leq  (\Toutsqm - 2 \Toutour) \gamma m
\end{equation}
Let us ignore $\Toutsqm$ on the left side of this inequality (in favor of SQM) and rearrange to get the looser condition,
\begin{equation}
\frac{nz}{m}  \leq  \frac{\gamma P}{\hat{k}} \frac{1}{2.0}(\frac{\Toutsqm}{\Toutour} - 2 )
\end{equation}
Assuming $\Toutsqm > 3.0\Toutour$, we arrive at the final condition in~(\ref{comm}).
%Note that in our experiments (section~\ref{distr}), the savings we get in the outer iterations is typically more than the factor $3.2$.

\section*{Appendix B: Proofs}
\label{sec:proofs}

\subsection*{Proofs of the results in section~\ref{general}}

Let us now consider the establishment of the convergence theory given in section~\ref{general}.

\noindent {\bf Proof of Lemma 1.}
Let $\rho(t)=f(w^r+td^r)$ and $\gamma(t)=\rho(t)-\rho(0)-\alpha t\rho^\prime(0)$.
Note the following connections with quantities involved in Lemma 1: $\rho(t)=f^{r+1}$, $\rho(0)=f^r$, $\rho^\prime(t)=g^{r+1}\cdot d^r$ and $\gamma(t)=f^{r+1} - f^r - \alpha g^r\cdot(w^{r+1}-w^r)$.
(\ref{ag}) corresponds to the condition $\gamma(t)\le 0$ and (\ref{wolfe}) corresponds to the condition $\rho^\prime(t)\ge \beta\rho^\prime(0)$.

$\gamma^\prime(t) = \rho^\prime(t)-\alpha \rho^\prime(0)$.
$\rho^\prime(0)<0$.
$\rho^\prime$ is strictly monotone increasing because, by assumption A2,
\begin{equation}
\rho^\prime(t)-\rho^\prime(\ttilde) \ge \sigma (t-\ttilde)\|d^r\|^2 \;\; \forall \; t, \ttilde
\label{prop11}
\end{equation}
This implies that $\gamma^\prime$ is also strictly monotone increasing and, all four, $\rho$, $\rho^\prime$, $\gamma^\prime$ and $\gamma$ tend to infinity as $t$ tends to infinity.

Let $t_\beta$ be the point at which $\rho^\prime(t)=\beta\rho^\prime(0)$. Since $\rho^\prime(0)<0$ and $\rho^\prime$ is strictly monotone increasing, $t_\beta$ is unique and $t_\beta>0$. This validates the definition in (\ref{tbeta}). Monotonicity of $\rho^\prime$ implies that (\ref{wolfe}) is satisfied iff $t\ge t_\beta$.

Note that $\gamma(0)=0$ and $\gamma^\prime(0)<0$. Also, since $\gamma^\prime$ is monotone increasing and $\gamma(t)\to\infty$ as $t\to\infty$, there exists a unique $t_\alpha>0$ such that $\gamma(t_\alpha)=0$, which validates the definition in (\ref{talpha}). It is easily checked that $\gamma(t)\le 0$ iff $t\in [0,t_\alpha]$.

The properties also imply $\gamma^\prime(t_\alpha)> 0$, which means $\rho^\prime(t_\alpha) \ge \alpha\rho^\prime(0)$. By the monotonicity of $\rho^\prime$ we get $t_\alpha>t_\beta$, proving the lemma.

\noindent {\bf Proof of Theorem 2.} Using (\ref{wolfe}) and A1,
\begin{equation}
(\beta-1)g^r\cdot d^r \le (g^{r+1}-g^r)\cdot d^r \le Lt\|d^r\|^2
\label{th21}
\end{equation}
This gives a lower bound on $t$:
\begin{equation}
t \ge \frac{(1-\beta)}{L\|d^r\|^2} (-g^r\cdot d^r)
\label{th22}
\end{equation}
Using (\ref{ag}), (\ref{th22}) and (\ref{angle}) we get
\begin{eqnarray}
f^{r+1} \le  f^r + \alpha t g^r\cdot d^r 
        \le  f^r - \frac{\alpha(1-\beta)}{L\|d^r\|^2} (-g^r\cdot d^r)^2
        \le  f^r - \frac{\alpha(1-\beta)}{L} \cos^2 \theta \|g^r\|^2
\label{th23}
\end{eqnarray}
Subtracting $f^\star$ gives
\begin{equation}
(f^{r+1} - f^\star) \le  (f^r-f^\star) - \frac{\alpha(1-\beta)}{L} \cos^2 \theta \|g^r\|^2
\label{th24}
\end{equation}
A2 together with $g(w^\star)=0$ implies $\|g^r\|^2\ge\sigma^2\|w^r-w^\star\|^2$. Also A1 implies $f^r-f^\star \le \frac{L}{2}\|w^r-w^\star\|^2$~\cite{smola2008}. Using these in (\ref{th24}) gives
\begin{eqnarray}
(f^{r+1}-f^\star) & \le & (f^r - f^\star) - 2\alpha(1-\beta)\frac{\sigma^2}{L^2} \cos^2\theta (f^r - f^\star) \nonumber \\
                  & \le & (1 - 2\alpha(1-\beta)\frac{\sigma^2}{L^2} \cos^2\theta) (f^r - f^\star)
\label{th25}
\end{eqnarray}
Let $\delta = (1 - 2\alpha(1-\beta)\frac{\sigma^2}{L^2} \cos^2\theta)$. Clearly $0 < \delta < 1$. Theorem 2 follows.

\subsection*{Proofs of the results in section~\ref{distr}}

Let us now consider the establishment of the convergence theory given in section~\ref{distr}. We begin by establishing that the exact minimizer of $\fhat_p$ makes a sufficient angle of descent at $w^r$.

\noindent {\bf Lemma 5.} Let $\what_p^\star$ be the minimizer of $\fhat_p$. Let $d_p= (\what_p^\star-w^r)$. Then
\begin{equation}
-g^r\cdot d_p \ge (\sigma/L) \|g^r\| \|d_p\|
\label{suffangle}
\end{equation}

\noindent {\bf Proof.} First note, using gradient consistency and $\grad f_p(\what_p^\star)=0$ that
\begin{equation}
\|g^r\| = \| \grad \fhat_p(w^r)-\grad \fhat_p(\what_p^\star) \| \le L\|d_p\|
\label{lem211}
\end{equation}
Now,
\begin{eqnarray}
\hspace*{-0.1in}
-g^r\cdot d_p = (\grad \fhat_p(w^r)-\grad \fhat_p(\what_p^\star))^T(w^r-\what_p^\star)
            \ge \sigma \|d_p\|^2
            =&\sigma \|g^r\| \|d_p\| \frac{\|d_p\|}{\|g^r\|} 
            \ge \frac{\sigma}{L} \|g^r\| \|d_p\|
\label{lem212}
\end{eqnarray}
where the second line comes from $\sigma$-strong convexity and the fourth line follows from (\ref{lem211}).

\noindent {\bf Proof of Lemma 3.}
Let us now turn to the question of approximate stopping and establish Lemma 3. Given $\theta$ satisfying (\ref{thetadef}) let us choose $\zeta\in (0,1)$ such that
\begin{equation}
\label{lem2211}
\frac{\pi}{2} > \theta > \cos^{-1} \frac{\sigma}{L} + \cos^{-1} \zeta
\end{equation}

%{\bf Lemma 7.} Assume $g(w^r)\not=0$. Suppose we minimize $\fhat_p$ using an optimizer that starts from $v^0=w^r$ and generates a sequence $\{v^k\}$ having linear convergence, i.e.,
%\begin{equation}
%\fhat_p(v^{k+1}) - \fhat_p^\star \le \delta (\fhat_p(v^k) - \fhat_p^\star)
%\label{22a}
%\end{equation}
%where $\fhat_p^\star = \fhat_p(\what_p^\star)$. Then, given any $\zeta$ satisfying $0<\zeta<1$, there exists $\khat$ (which depends only on $\zeta$, $\sigma$ and $L$) such that
%\begin{equation}
%\cos \phi^k \ge \zeta \;\; \forall k\ge \khat
%\label{22b}
%\end{equation}
%where $\phi^k$ is the angle between $\what_p^\star-w^r$ and $v^k-w^r$.

%fhat to fhat_p

%Add a diagram to help
\begin{figure}[H]
\begin{center}
\includegraphics[width=0.6\columnwidth]{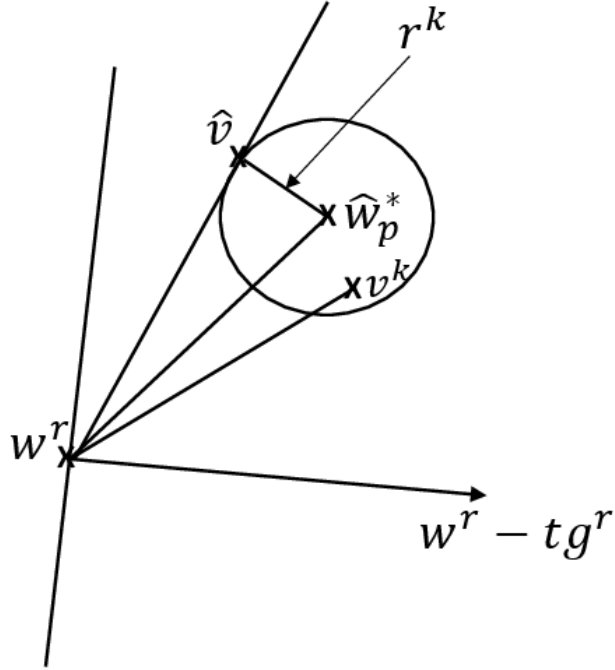}
\caption{Construction used in the proof of Lemma 3.}
\end{center}
\label{spherefig}
\end{figure}

By A3 and equations ($3.16$) and ($3.22$) in~\cite{smola2008}, we get
\begin{equation}
\frac{\sigma}{2} \|v-\what_p^\star\|^2 \le \fhat_p(v) - \fhat_p^\star \le \frac{L}{2} \|v-\what_p^\star\|^2
\label{lem221}
\end{equation}
After $k$ iterations we have
\begin{equation}
\fhat_p(v^k) - \fhat_p^\star \le \delta^k (\fhat_p(w^r) - \fhat_p^\star)
\label{lem222}
\end{equation}
We can use these to get
\begin{eqnarray}
\|v^k-\what_p^\star\|^2 \le \frac{2(\fhat_p(v^k)-\fhat_p^\star)}{\sigma} 
                        \le \frac{2\delta^k (\fhat_p(w^r)-\fhat_p^\star)}{\sigma}
                        \le \frac{\delta^kL}{\sigma} \|w^r-\what_p^\star\|^2 \defs (r^k)^2
\label{lem223}
\end{eqnarray}
For now let us assume the following:
\begin{equation}
\|v^k-\what_p^\star\|^2 \le \|w^r-\what_p^\star\|^2
\label{lem224}
\end{equation}
Using (\ref{lem223}) note that (\ref{lem224}) holds if
\begin{equation}
\frac{\delta^kL}{\sigma} \le 1
\label{lem225}
\end{equation}
Let $S^k$ be the sphere, $S^k = \{ v : \|v-\what_p^\star\|^2 \le (r^k)^2 \}$. By (\ref{lem223})
%and (\ref{lem224})
we have $v^k\in S^k$. See Figure~\ref{spherefig}. Therefore,
\begin{equation}
\phi^k \le \max_{v\in S^k} \phi(v)
\label{lem226}
\end{equation}
where $\phi^k$ is the angle between $\what_p^\star-w^r$ and $v^k-w^r$, and $\phi(v)$ is the angle between $v-w^r$ and $\what_p^\star-w^r$. Given the simple geometry, it is easy to see that $\max_{v\in S^k} \phi(v)$ is attained by a point $\vhat$ lying on the boundary of $S^k$ (i.e., $\|\vhat-\what_p^\star\|^2 = (r^k)^2$) and satisfying $(\vhat-\what_p^\star)\perp (\vhat-w^r)$. This geometry yields
\begin{eqnarray}
\cos^2 \phi(\vhat) = \frac{\|\vhat-w^r\|^2}{\|\what_p^\star-w^r\|^2}
                   = \frac{\|\what_p^\star-w^r\|^2-(r^k)^2}{\|\what_p^\star-w^r\|^2} 
                   = 1-\frac{(r^k)^2}{\|\what_p^\star-w^r\|^2} = 1-\frac{\delta^kL}{\sigma}
\label{lem227}
\end{eqnarray}
Since $\phi^k\le \phi(\vhat)$,
\begin{equation}
\cos^2 \phi^k \ge 1-\frac{\delta^kL}{\sigma}
\label{lem228}
\end{equation}
Thus, if
\begin{equation}
1-\frac{\delta^kL}{\sigma} \ge \zeta^2
\label{lem229}
\end{equation}
then
\begin{equation}
\cos \phi^k \ge \zeta \;\; \forall k\ge \khat
\label{22b}
\end{equation}
holds. By (\ref{lem2211}) this yields $\phase{-g^r,v^k-w^r} \le \theta$, the result needed in Lemma 3. Since $\zeta>0$, (\ref{lem229}) implies (\ref{lem225}), so (\ref{lem224}) holds and there is no need to separately satisfy it. Now (\ref{lem229}) holds if
\begin{equation}
k \ge \khat \defs \frac{\log (L/(\sigma(1-\zeta^2)))}{\log(1/\delta)}
\label{lem2210}
\end{equation}
which proves the lemma.

\noindent {\bf Proof of Theorem 4.} It trivially follows from a combination of Lemma 3 and Theorem 2.

\end{document}